%% file: bcdf_jmlr_arxiv.tex
\documentclass[letterpaper, oneside, 12pt]{article}

\usepackage{amsmath}
\usepackage{graphicx}%
\usepackage{amsfonts}%
\usepackage{amssymb}
\usepackage{subfigure}
\usepackage{rotating}
\usepackage{epstopdf}
\usepackage{url}
\usepackage{comment}
\usepackage{enumerate}
\usepackage{mathtools}
\usepackage{multirow}
\usepackage{natbib}
\usepackage[usenames,dvipsnames]{color}

\usepackage{setspace}
\usepackage[hmargin=1.0in,vmargin=1.0in]{geometry}

\usepackage[ruled]{algorithm}
\usepackage{algorithmicx}
\usepackage{algpseudocode}
\usepackage{tikz}

\algnewcommand{\algorithmicgoto}{\textbf{go to}}
\algnewcommand{\Goto}[1]{\algorithmicgoto~\ref{#1}}

\usepackage{paralist}
\usepackage{colortbl}
\usepackage{hyperref}
\hypersetup{colorlinks=true,citecolor=red,linkcolor=red,urlcolor=red}

\usepackage{array}
\newcolumntype{L}[1]{>{\raggedright\let\newline\\\arraybackslash\hspace{0pt}}m{#1}}
\newcolumntype{C}[1]{>{\centering\let\newline\\\arraybackslash\hspace{0pt}}m{#1}}
\newcolumntype{R}[1]{>{\raggedleft\let\newline\\\arraybackslash\hspace{0pt}}m{#1}}

\input{defs}

\title{Bayesian Conditional Density Filtering}

    \makeatletter
    \let\@fnsymbol\@arabic
    \makeatother
\title{Bayesian Tensor Regression}
\author{Shaan Qamar$^{\bf*}$\thanks{Shaan Qamar, Ph.D., Department of Statistical Science, Duke University, Durham, NC 27708-0251, (E-mail: shaan.qamar@duke.edu).} \and Rajarshi Guhaniyogi$^{\bf*}$\thanks{Rajarshi Guhaniyogi,  Assistant Professor, Department of Applied Math \& Stat, SOE2, UC Santa Cruz, 1156 High Street, Santa Cruz, CA 95064 (E-mail: rguhaniy@ucsc.edu).} \and David B. Dunson\thanks{David B. Dunson, Arts \& Sciences Distinguished Professor, Department of Statistical Science, 218 Old Chemistry Building, Box 90251, Duke University, Durham, NC 27708-0251 (E-mail: dunson@duke.edu)}}

\begin{document}
\date{July 15, 2015}
\maketitle

\begin{abstract}
We propose a Conditional Density Filtering (C-DF) algorithm for efficient online Bayesian inference.  C-DF adapts MCMC sampling to the online setting, sampling from approximations to conditional posterior distributions obtained by propagating surrogate conditional sufficient statistics (a function of data and parameter estimates) as new data arrive.  These quantities eliminate the need to store or process the entire dataset simultaneously and offer a number of desirable features.  Often, these include a reduction in memory requirements and runtime and improved mixing, along with state-of-the-art parameter inference and prediction.  These improvements are demonstrated through several illustrative examples including an application to high dimensional compressed regression.  Finally, we show that C-DF samples converge to the target posterior distribution asymptotically as sampling proceeds and more data arrives.
\end{abstract}
\renewcommand\thefootnote{}\footnote{$*$\bf These authors contributed equally}



\section{Introduction}
Modern data are increasingly high dimensional, both in the number of observations $n$ and the number of predictors measured $p$.  Statistical methods increasingly make use of low-dimensional structure assumptions to combat the curse of dimensionality (e.g., sparsity assumptions) and efficient model fitting tools must evolve quickly to keep pace with the rapidly growing dimension and complexity of data they are applied to.
 Bayesian methods provide a natural probabilistic characterization of uncertainty in the parameters and in predictions, but there is a lack of scalable inference algorithms having guarantees on accuracy.  Markov chain Monte Carlo (MCMC) methods for Bayesian computation are routinely used due to ease, generality and convergence guarantees.  When the number of observations is truly massive, however, data processing and computational bottlenecks render many MCMC methods infeasible as they demand (1) the entire dataset (or available sufficient statistics) be held in memory; and (2) likelihood evaluations for updating model parameters at every sampling iteration which can be costly.

A number of alternative strategies have been proposed in recent years for scaling MCMC to large datasets.  One possibility is to parallelize computation within each MCMC iteration using GPUs or multicore architectures to free bottlenecks in updating unknowns specific to each sample and in calculating likelihoods \citep{medlar2013swiftlink}. Another possibility is to rely on Hamiltonian Monte Carlo with stochastic gradient methods used to approximate gradients with subsamples of the data \citep{welling2011bayesian, teh2014consistency}. \cite{korattikara2013austerity} instead use sequential hypothesis testing to choose a subsample to use in approximating the acceptance ratio in Metropolis-Hastings. More broadly, data subsampling approaches attempt to mitigate the MCMC computational bottleneck by choosing a small set of points at each iteration for which the likelihood is to be evaluated, and doing so in a way that preserves validity of the draws as samples from the desired target posterior  \citep{quiroz2014speeding,maclaurin2014firefly}. Building on ideas reminiscent of sequential importance resampling and particle filtering \citep{doucet2000sequential,arulampalam2002tutorial}, these methods are promising and are an active area of research. Yet another approach assigns different data subsets to different machines, runs MCMC in parallel for each subset, and recombines the resulting samples  \citep{scott2013bayes,minsker2014robust}. However, theoretical guarantees for the latter have not been established in great generality.

Sequential Monte Carlo (SMC) \citep{chopin2002sequential,arulampalam2002tutorial,lopes2011particle,doucet2001introduction}  is a popular technique for online Bayesian inference that relies on resampling particles sequentially as new data arrive. Unfortunately, it is difficult to scale SMC to problems involving large $n$ and $p$ due to the need to employ very large numbers of particles to obtain adequate approximations and prevent particle degeneracy.  The latter is addressed through rejuvenation steps using all the data (or sufficient statistics), which becomes expensive in an online setting.  One could potentially rejuvenate particles only at earlier time points, but this may not protect against degeneracy for models involving many parameters.  More recent particle learning (PL) algorithms \citep{carvalho2010particle} reduce degeneracy for the dynamic linear model, with satisfactory density estimates for parameters -- but they too require propagating a large number of particles, which significantly adds to the per-iteration computational complexity.  MCMC can be extended to accommodate data collected over time by adapting the transition kernel $K_t$, and drawing a few samples at time $t$, so that samples converge in distribution to the joint posterior distribution $\pi_t$ in time \citep{yang2013sequential}.  However, this method requires the full data or available sufficient statistics to be stored, leading to storage and processing bottlenecks as more data arrive in time.  Indeed, models with large parameter spaces often have sufficient statistics which are also high dimensional (e.g., linear regression).  In addition, MCMC often faces poor mixing requiring longer runs with additional storage and computation.

In simple conjugate models, such as Gaussian state-space models, efficient updating equations can be obtained using methods related to the Kalman filter.  Assumed density filtering (ADF) was proposed \citep{lauritzen1992propagation,boyen1998tractable,opper1998bayesian} to extend this computational tractability to broader classes of models.  ADF approximates the posterior distribution with a simple conjugate family, leading to approximate online posterior tracking. The predominant concern with this approach is the propagation of errors with each additional approximation to the posterior in time.  Expectation-propagation (EP) \citep{minka2001expectation,minka2009virtual} improves on ADF through additional iterative refinements, but the approximation is limited to the class of assumed densities and has no convergence guarantees. Moreover in fairly standard settings, arbitrary approximation to the posterior through an assumed density severely underestimates parameter and predictive uncertainties.  Similarly, variational methods developed in the machine learning literature attempt to address various difficulties facing MCMC methods by making additional approximations to the joint posterior distribution over model parameters. These procedures often work well in simple low dimensional settings, but fail to adequately capture dependence in the joint posterior, severely underestimate uncertainty in more complex or higher dimensional settings and generally come with no accuracy guarantee.  Additionally, they often require computing expensive gradients for large datasets. \cite{hoffman2013stochastic} proposed stochastic variational inference (SVI) which uses stochastic approximation to the full gradient for a subset of parameters, thus circumventing this computational bottleneck. However, SVI requires retrieving, storing or having access to the entire dataset at once, which is often not feasible in very large or streaming data settings.  A parallel literature on online variational approximations \citep{hoffman2010online} focus primarily on improving batch inferences by feeding in data sequentially.  In addition, these methods require specifying or tuning of a `learning-rate,' and generally come with no storage reduction.  \cite{broderick2013streaming} recently proposed a streaming variational Bayes (SVB) algorithm to facilitate storage for only the recent batch of data for a data stream. However, all variational methods (batch or online) rely on a factorized form of the posterior that typically fails to capture dependence in the joint posterior and severely underestimates uncertainty. Recent attempts to design careful online variational approximations \citep{luts2013variational}, though successful in accurately estimating marginal densities, are limited to specific models and no theoretical guarantees on accuracy are established except for stylized cases.

We propose a new class of Conditional Density Filtering (C-DF) algorithms that extend MCMC sampling to streaming data.  Sampling proceeds by drawing from conditional posterior distributions, but instead of conditioning on conditional sufficient statistics (CSS) \citep{carvalho2010particle,johannes2010particle}, C-DF conditions on surrogate conditional sufficient statistics (SCSS) using sequential point estimates for parameters along with the data observed. This eliminates the need to store the data in time (process the entire dataset at once), and leads to an approximation of the conditional distributions that produce samples from the correct target posterior asymptotically.  The C-DF algorithm is demonstrated to be highly versatile and efficient across a variety of settings, with SCSS enabling online sampling of parameters often with dramatic reductions in the memory and per-iteration computational requirements.  C-DF has also successfully been applied to non-negative tensor factorization models for massive binary and count-valued tensor streaming data, with state-of-the-art performance demonstrated  against a wide-range of batch and online competitors   \citep{huzero,huscalable}. 

Section 2 introduces the C-DF algorithm in generality, along with definitions, assumptions, and a description of how to identify updating quantities of interest.  Section 3 demonstrates approximate online inference using C-DF in the context of several illustrating examples. Section 3.1 applies C-DF to linear regression and the one-way Anova model. More complex model settings are considered in Sections 3.2, namely extensions of the C-DF algorithm to a dynamic linear model and binary regression using the high dimensional  probit model.  Here, we investigate the performance of C-DF in settings with an increasing parameter space. Along with comparing inferential and predictive performance, we discuss various computational, storage and mixing advantages of our method over state-of-the-art competitors in each.  Section 4 presents a detailed implementation of the C-DF algorithm for high dimensional compressed regression.  We report state-of-the-art inferential and predictive performance across extensive simulation experiments as well as for real data studies in Section 5. The C-DF algorithm is also applied to a Poisson mixed effects model to update the parameters for a variational approximation to the posterior distribution in Appendix \ref{sec:poi_mix}.   Section 6 presents a finite-sample error bound for approximate MCMC kernels and establishes the asymptotic convergence guarantee for the proposed C-DF algorithm. Section 7 concludes with a discussion of extensions and future work. Proofs and additional figures pertaining to Section 3 appear in Appendix \ref{sec:proofs} and \ref{sec:additional_figs}, respectively.

\section{Conditional density filtering}\label{sec:cdf_alg}

Define $\bTheta = (\btheta_1, \btheta_2, \ldots, \btheta_k)$ as the collection of unknown parameters in probability model $P(Y | \bTheta)$ and $Y \in \mathcal{Y}$, with $\btheta_j \in \bPsi_j$, and $\bPsi_j$ denoting an arbitrary sample space (e.g., a subset of $\Re^p$). Data $\bD_t$ denotes the data observed at time $t$, while $\bD^{(t)} = \{\bD_s, ~s = 1, \dots, t\}$ defines the collection of data observed through time $t$. Below, $\btheta_{-j} = \bTheta \setminus \btheta_j$ and let $\btheta_{-j} = (\btheta_{-j,1}, \btheta_{-j,2})$ for each $j = 1, \dots, k$. Either of $\btheta_{-j,1}$ or $\btheta_{-j,2}$ is allowed to be a null set.

\subsection{Surrogate conditional sufficient statistics}

\begin{definition}
$\bS_j^{(t)}$ is defined to be a {\em conditional sufficient statistic} (CSS) for $\btheta_j$ at time $t$ if $\btheta_j \perp \bD^{(t)} \given \btheta_{-j,1},\bS_j^{(t)}$. Suppose $\btheta_j\given\btheta_{-j},\bD_t\stackrel{\mathcal{L}}{=}\btheta_j\given\btheta_{-j,1}, h(\bD_t,\btheta_{-j,2})$\footnote{$\theta_j | \theta_{-j}, D$ denotes the conditional distribution of parameter $\theta_j$ given all other model parameters $\theta_{-j}$ and a dataset $D$.}.
Then it is easy to show that for known functions $f,h$, $\bS_j^{(t)} = f(h(\bD_1, \btheta_{-j,2}), \ldots, h(\bD_t, \btheta_{-j,2}))$. This satisfies $\btheta_j \given \btheta_{-j}, \bD^{(t)} \stackrel{\mathcal{L}}{=} \btheta_j \given \btheta_{-j,1}, \bS_j^{(t)}$,
or equivalently $\btheta_j \perp \bD^{(t)} \given \btheta_{-j,1},\bS_j^{(t)}$.
\end{definition}
Because $\bS_j^{(t)}$ depends explicitly on $\btheta_{-j,2}$, its value changes whenever new samples are drawn for this collection of parameters.  This necessitates storing entire data $\bD^{(t)}$ or available sufficient statistics.  The following issues inhibit efficient online sampling: 
(1) sufficient statistics, when they exist, often scale poorly with the size of the parameter-space, thus creating a significant storage overhead;  
(2) updating $\bS_j^{(t)}, ~j = 1, \dots, k$, for every iteration at time $t$ causes an immense per-iteration computational bottleneck; and %
(3) updating potentially massive numbers of observation-specific latent variables can lead to significant computational overhead unless conditional independence structures allow for parallelized draws. 
To address these challenges, we propose {\em surrogate conditional sufficient statistics} (SCSS) as a means to approximate full conditional distribution $\btheta_j | \btheta_{-j}, \bD^{(t)}$ at time $t$.
\begin{definition}\label{eq:scss_definition}
Suppose $\btheta_j\given\btheta_{-j},\bD_t\stackrel{\mathcal{L}}{=}\btheta_j\given\btheta_{-j,1}, h(\bD_t,\btheta_{-j,2})$.  For known functions $g,h$, define $\bC_j^{(t)} = g(\bC_j^{(t-1)}, h(\bD_t, \widehat{\btheta}_{-j,2}^t))$ as the {\em surrogate conditional sufficient statistic} for $\btheta_j$, with $\widehat{\btheta}_{-j,2}^t$ being a consistent estimator of $\btheta_{-j,2}$ at time $t$.  Then, $\btheta_j | \btheta_{-j,1}, \bC_j^{(t)}$ is the C-DF approximation to $\btheta_j | \btheta_{-j}, \bD^{(t)}$.
\end{definition}

If the full conditional for $\btheta_j$ admits a surrogate quantity $\bC_j^{(t)}$, approximate sampling via C-DF proceeds by drawing $\mt{\btheta}_j \sim \mt{\pi}_j(\cdot | \btheta_{-j,1}, \bC_j^{(t)})$.  Crucially, $\bC_j^{(t)}$ depends only an estimate for $\btheta_{-j,2}$ which remain fixed while drawing samples from approximate full conditional distribution $\tilde\pi_j(\cdot)$.  This avoids having to update potentially expensive functionals $h(\bD^{(t)}, \btheta_{-j,2})$ (i.e., CSS) for each parameter $\btheta_j$ at every MCMC iteration.  If the approximating full conditionals are conjugate, then sampling proceeds in a Gibbs-like fashion.  Otherwise, draws from $\mt{\pi}_j(\btheta_j | \btheta_{-j,1}, \bC_j^{(t)})$ may be obtained via a Metropolis step for a proposal distribution $p_j(\cdot | -)$ for $\btheta_j$ and acceptance probability
\begin{align}
\alpha_j(\btheta, \btheta') = \frac{\mt{\pi}_j(\btheta' | \btheta_{-j,1}, \bC_j^{(t)})}{\mt{\pi}_j(\btheta | \btheta_{-j,1}, \bC_j^{(t)})} \frac{p_j(\btheta | \btheta_j')}{p_j(\btheta' | \btheta_j)}.
\end{align} 
Section \ref{sec:motivating_examples} provides examples of the former, and Section \ref{sec:DLM_CDF} of the latter.  In both cases, draws using the C-DF approximation to the conditional distributions have the correct asymptotic stationary distribution (see Section \ref{sec:theory_cdf}).  

While this article focuses on cases where the conditional distributions admit surrogate quantities or sufficient statistics, this is often not the case for at least some of the model parameters. In such settings, various distributional approximations (e.g., Laplace approximation, variational Bayes, expectation propagation etc.) are often employed for tractability.  Here, it is possible to use C-DF in conjunction with these methods to yield approximate inference in a streaming data setting, albeit without theoretical guarantees on the limiting distribution of the samples obtained.  Appendix \ref{sec:poi_mix} considers an application to Poisson regression where the conditional distributions do not admit surrogate quantities and no augmentation scheme is available.   Here, a variational approximation to the joint posterior admits SCSS, and we demonstrate using the C-DF algorithm when a full conditional $\pi_t(\btheta_j | -)$ is replaced by an approximating kernel $q_j(\btheta_j | \btheta_{-j,1}, \bC_j^{(t)})$ at time $t$.  Propagating SCSS associated with the latter allows us to obtain approximate online inference, although the limiting distribution for sampled draws are not guaranteed to have the correct asymptotic stationary distribution. 

\subsection{The C-DF algorithm}
Define a fixed grouping of parameters $\bTheta = \{\btheta_j : 1 \le j \le p\}$, into sets $\mathcal{G}_l$, $l = 1, \dots, k$, subject to $\mathcal{G}_l \cap \mathcal{G}_{l'} = \emptyset, ~l \ne l'$ and $\cup_{l = 1}^k \mathcal{G}_l = \bTheta$.  The model specification and conditional independence assumptions often suggest natural parameter partitions, though alternate partitions may be more suitable for specific tasks. 
 For streaming data, these sets may be identified to maximize computational and storage gains.  
 Examples of various partitioning schemes are presented in the context of several illustrating examples in Section \ref{sec:all_illustrating_examples}.
 
For model parameters indexed by $\mathcal{G}_l$, $\{\btheta_j : j \in \mathcal{G}_l\}$, $1 \le l \le k$, the C-DF algorithm samples sequentially from the respective approximating conditional distributions $\tilde\pi_t\big(\btheta_j | \bTheta_{\mathcal{G}_l}^{(j)}, \bC_j^{(t)}\big)$, where $\bTheta_{\mathcal{G}_l}^{(j)} = \bTheta_{\mathcal{G}_l} \setminus \btheta_j$. Efficient updating equations for $\bC_j^{(t)}$ based on $\bC_j^{(t-1)}$, the incoming data shard $\bD_t$, and estimates for $\bTheta_{(l)} = \{\btheta_j, ~j \not\in \mathcal{G}_l\}$ from the previous time-point result in scalable parameter inference with sampling-based approximations to the posterior distribution; see definition \ref{eq:scss_definition}.  The  approximate samples are in turn used to obtain estimates for $\btheta \in \bTheta_{\mathcal{G}_l}$\footnote{In cases where closed-form expression for the conditional mean is  available, this may be used instead of Monte Carlo estimates.} and the algorithm continues by iterating over the parameter partitions. An outline of the C-DF algorithm is given in Algorithm \ref{alg1}.


\begin{algorithm}[h]
\caption{A sketch of the C-DF algorithm for approximate online MCMC} \label{alg1}
\begin{algorithmic}[1]
\Require (1) Data shard $\bD_t$ at time $t$; (2) parameter partition index sets $\mathcal{G}_l$, $l = 1, \dots, k$, with $\sum_{l=1}^k I(j \in \mathcal{G}_l) =1$, $j = 1, \dots, p$; and (3) parameter SCSS $\bC_j^{(t-1)}, j = 1, \dots, p$.
\Ensure Approximate posterior draws $\mt{\btheta}_j^{(1)}, \dots, \mt{\btheta}_j^{(S)}$ and SCSS $\bC_j^{(t)}, \: j = 1, \dots, p$.
\Function{CDF.SAMPLE}{$\bD_t, \{\bC_j^{(t-1)}\}, \{\mathcal{G}_l\}$}
\For{$l = 1 : k$}\\
\hspace{35pt}\mbox{\tt // step 1: update SCSS for $\bTheta_{\mathcal{G}_l}$}
	\For{$j\in\mathcal{G}_l$}
	\\
	\hspace{55pt}\mbox{\tt // $\bTheta_{(l)} = \{\btheta_j, ~j \not\in \mathcal{G}_l\}$}
	\State set $\bC_j^{(t)} \leftarrow g\big(\bC_j^{(t-1)}, h(\bD_t, \widehat{\bTheta}^{(t-1)}_{(l)}\big)$
	\EndFor
	\\
	\\
\hspace{35pt}\mbox{\tt // step 2: approximate C-DF sampling}
	\For{$s = 1 : S$}
	\For{$j \in \mathcal{G}_l$}
	\\
	\hspace{70pt}\mbox{\tt // $\bTheta_{\mathcal{G}_l} = \{ \btheta_j, \: j \in \mathcal{G}_l\}$ and  $\bTheta_{\mathcal{G}_l}^{(j)} = \bTheta_{\mathcal{G}_l} \setminus \btheta_j$}
	\State sample $\mt{\btheta}_j^{(s)} \sim \mt{\pi}_j(\cdot | \bTheta_{\mathcal{G}_l}^{(j)}, \bC_j^{(t)})$
	\EndFor
	\EndFor
	\\
	\\
	\hspace{35pt}\mbox{\tt // step 3: update parameter estimates}
	\For{$j \in \mathcal{G}_l$}
	\State set $\widehat{\btheta}^{(t)}_j \leftarrow \mathrm{mean}\big(\tilde\btheta_j^{(1)}, \dots, \tilde\btheta_j^{(S)}\big)$
	\State store $\mt{\btheta}_j^{(1)}, \dots, \mt{\btheta}_j^{(S)}$ as approximate posterior samples at time $t$
	\EndFor
	\EndFor
\EndFunction
\end{algorithmic}
\end{algorithm}

\section{Illustrating examples using C-DF} \label{sec:all_illustrating_examples}

The following notation is used for examples considered in this section as well as those presented in Section \ref{sec:complex_examples}. Data $\bD_t$ denotes the data observed at time $t$, while $\bD^{(t)} = \{\bD_s, ~s = 1, \dots, t\}$ defines the collection of data observed through time $t$.  Where appropriate, $\bD_t = (\bX^t, \by^t)$ with $\bX^t=(\bx_1^t, \dots, \bx_n^t)'$ and $\by^t=(y_1^t, \dots,y_n^t)'$.  Shards of a fixed size arrive sequentially over a $T=500$ time horizon, and 500 MCMC iterations are drawn to approximate the corresponding posterior distribution $\pi_t$ at every time point $t = 1, \dots, T$.  Where applicable, the following quantities are reported to measure estimation and inferential performance for competing methods: (i) Mean squared estimation error on parameters of interest; (ii) mean squared prediction error (MSPE); and (iii) length and coverage of 95\% predictive intervals.  Results are averaged over 10 independent replications with associated standard errors appearing as subscripts in Tables.  All reported runtimes are based on a non-optimized R implementation run on an x$86\times 64$ Intel(R) Core(TM) i5 machine.


Finally, plots of kernel density estimates for marginal posterior densities on representative model parameters are shown at various time points. At time $t$, let
\begin{align}\label{eq:accuracy_pl}
1 - \frac{1}{2} \int \big|\pi_t(\btheta_j) - \tilde{\pi}_t(\btheta_j)\big| \: d\btheta_j
\end{align}
be a measure of `accuracy' between approximating C-DF density $\mt{\pi}_t$ and the full conditional distribution $\pi_t$ obtained using batch MCMC (S-MCMC).  As defined, accuracy ranges between 0 and 1 (larger values are better). Accuracy is plotted as a function of time for examples of this section, with figures appearing in Appendix \ref{sec:additional_figs}.


\subsection{Motivating examples}\label{sec:motivating_examples}

\subsubsection{Linear regression}\label{sec:lm}
For the Gaussian error model, a response $y \in \Re$ given an associated $p$-dimensional predictor $x \in \Re^p$ is modeled in the linear regression setting as
\begin{align} \label{eqn:linreg}
y = x' \bbeta + \epsilon, \quad \epsilon \sim \mathrm{N}(0,\sigma^2).
\end{align}

A standard Bayesian analysis proceeds by assigning conjugate priors $\bbeta \sim \mathrm{N}(\bzero, \bI_p)$ and $\sigma^2 \sim \mathrm{IG}(a,b)$, with associated full conditionals given as
\begin{align*}
\footnotesize
\sigma^2|\bbeta,\bD^{(t)} &\sim \mathrm{IG}(a' , b'),  \quad & a'= a + nt / 2, \quad b' = b + \frac{1}{2} \big(S_t^{YY} - 2\bbeta'\bS_t^{XY} + \bbeta'\bS_t^{XX}\bbeta\big) \\
\bbeta | \sigma^2, \bD^{(t)} &\sim \mathrm{N}(\mu_t, \bSigma_t), \quad & \bSigma_t = \big(\bS_t^{XX} / \sigma^2 + \bI_p\big)^{-1}, \quad \bmu_t=\bSigma_t \bS_t^{XY} / \sigma^2.
\end{align*}
These are parametrized in terms of sufficient statistics $\bS^{XX}_t=\bS^{XX}_{t-1}+\bX^{t'}\bX^t$, $\bS^{XY}_t=\bS^{YY}_{t-1}+\bX^{t'}\by^t$ and $S^{YY}_t=S^{YY}_{t-1}+\by^{t'}\by^t$, thus enabling efficient inference using S-MCMC.

A C-DF algorithm from approximate online inference in this setting begins by defining a partition over the model parameters; here, we choose $\bTheta_{\mathcal{G}_1}=\{\bbeta\}$ and $\bTheta_{\mathcal{G}_2}=\{\sigma^2\}$.  Sampling then proceeds as:
\begin{enumerate}[(1)]
\item Observe data $\bD_t$ at time $t$. If $t=1$ initialize all parameters at some default values (e.g., $\bbeta = 0$, $\sigma^2 = 1$ assuming a centered and scaled response); otherwise set $\hat{\sigma}^2_{t} \leftarrow \hat{\sigma}^2_{t-1}$;
\item Define $\bC_1^t = \{\bC_{1,1}^t, \bC_{1,2}^t\}$ as SCSS for $\bbeta$. Update $\bC_1^t$ as $\bC_{1,1}^{t}=\bC_{1,1}^{t-1}+ \bX^{t'}\bX^t / \hat{\sigma}^2_{t}$ and $\bC_{1,2}^{t}=\bC_{1,2}^{t-1}+\bX^{t'}\by^t / \hat{\sigma}^2_{t}$;
\item Draw $S$ samples from the approximate Gibbs full conditional $\bbeta | \hat{\sigma}^2, \bC_1^{(t)} \sim \mathrm{N}(\hat{\bmu}_t, \hat{\bSigma}_t)$ ($\hat{\bSigma}_t=(\bC_{1,1}^t+\bI_p)^{-1},\hat{\bmu}_t=\hat{\bSigma}_t \bC_{1,2}^t$), and set $\hat{\bbeta}_t \leftarrow \mathrm{mean}(\bbeta^{(1:S)})$ (or use the analytical expression for the posterior mean);
\item Define $\bC_2^{t} = \{\bC_{2,1}^t, \bC_{2,2}^t\}$ as SCSS for $\sigma^2$. Update $\bC_2^t$ as $\bC_{2,1}^{t}=\bC_{2,1}^{t-1} + \hat{\bbeta}_t' \bX^{t'}\bX^t \hat{\bbeta}_t$ and $\bC_{2,2}^{t}=\bC_{2,2}^{t-1}+ \hat{\bbeta}_t'\bX^{t'}\by^t$;
\item Draw $S$ samples from the approximate Gibbs full conditional $\sigma^2| \hat{\bbeta}_t, \bC_2^{(t)} \sim \mathrm{IG}\big(a', b + (S_t^{YY} - 2\bC_{2,2}^t + \bC_{2,1}^t)/2\big)$, and set $\hat{\sigma}_t^2 \leftarrow \mathrm{mean}(\sigma^{2(1:S)})$ (or use the analytical expression for the posterior mean).
\end{enumerate}

\begin{figure}[H]
\centering
\begin{tabular}{C{0.33\columnwidth} | C{0.66\columnwidth}}
Linear regression & One-way Anova \\
\hline
\includegraphics[width=\linewidth]{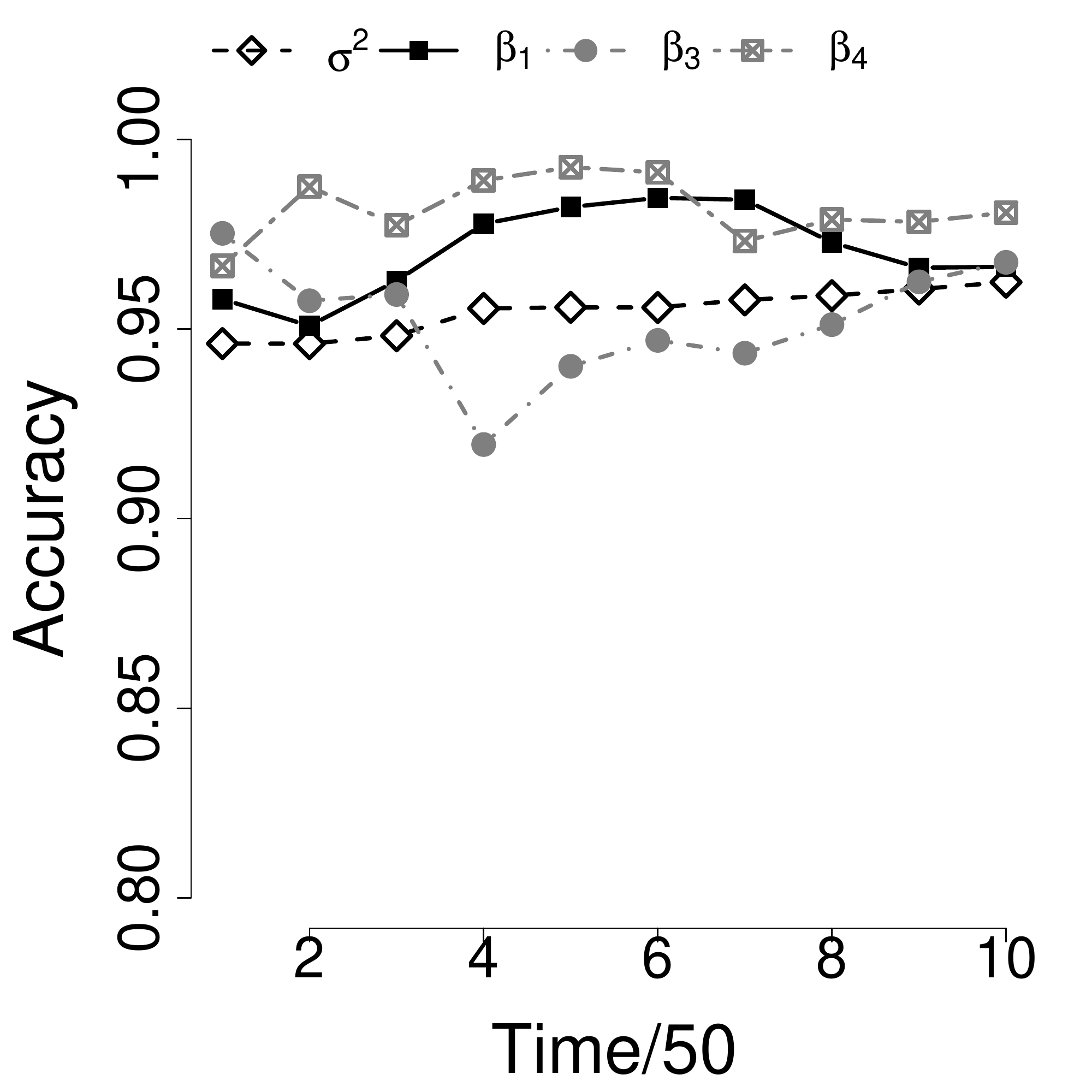} & 
\includegraphics[width=0.49\linewidth]{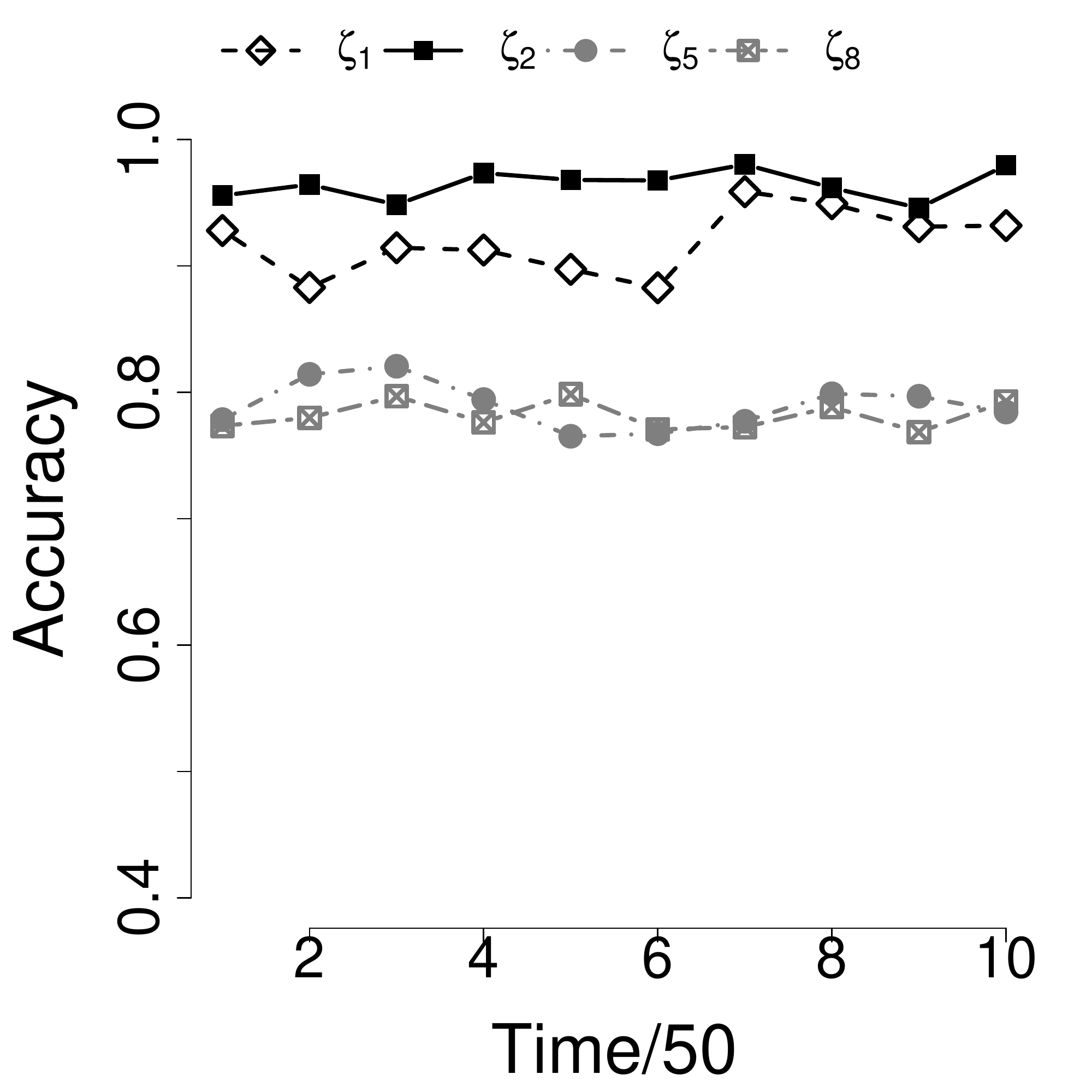} 
\includegraphics[width=0.49\linewidth]{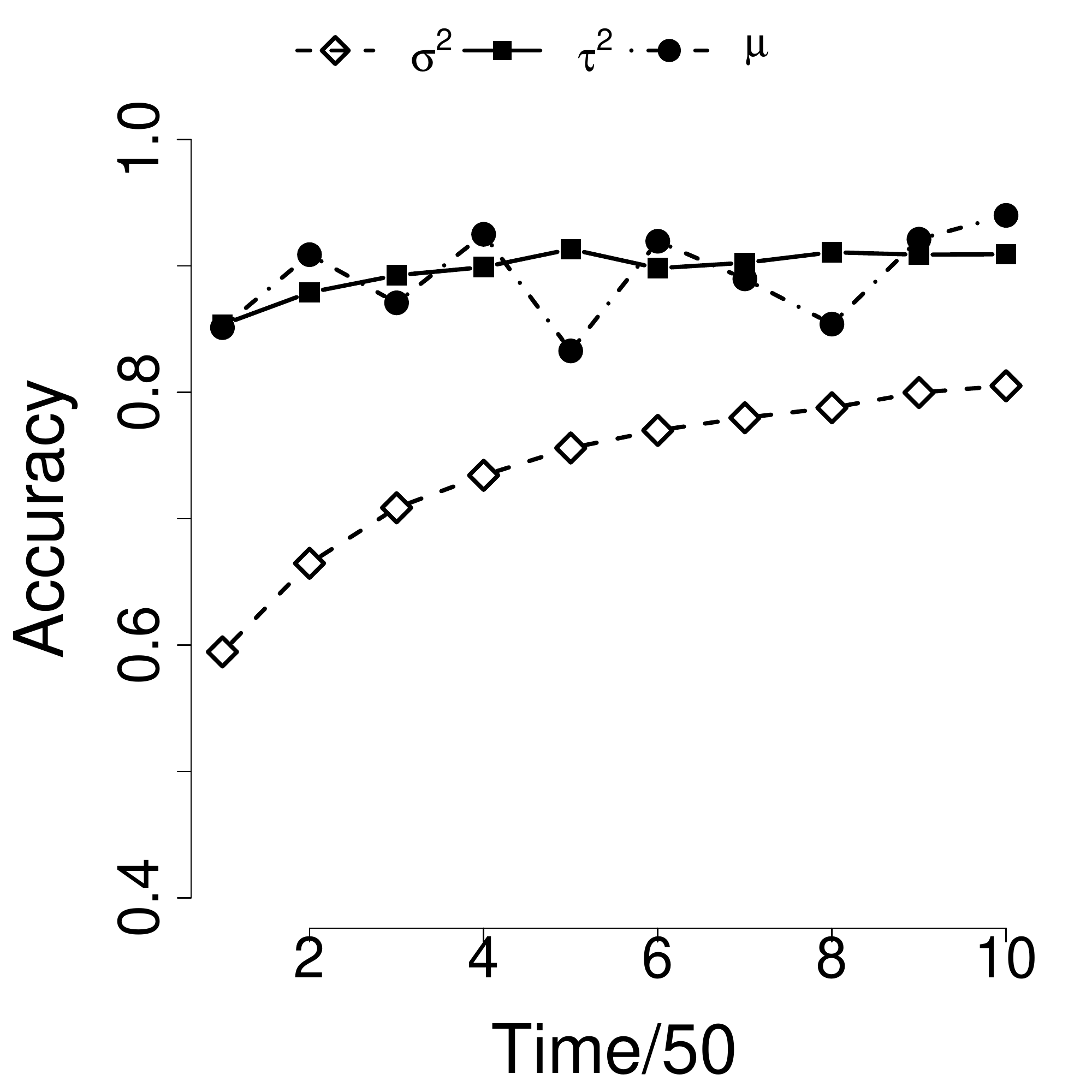}
\end{tabular}
\caption{Parameter accuracy plotted over time as defined in
  \eqref{eq:accuracy_pl} for the motivating examples of Section
  \ref{sec:motivating_examples}. Image 1: accuracy for representative
  regression coefficients $\beta_j$ and $\sigma^2$ in the linear
  regression example. Images 2 and 3: accuracy for
  representative group means $\zeta_j$, along with hierarchical
  parameters $\mu,\:\tau^2$ and $\sigma^2$ for the one-way Anova
  model.}
\label{fig:accuracy_plots}
\end{figure}

Data shards of size $n_t = 10$ are generated using predictors drawn from $\mathrm{U}(0,1)$, with true parameters $\bbeta_0 = (1.00, 0.50, 0.25, -1.00,0.75)$ and $\sigma_0=5$.  Density estimates for model parameters are displayed at $t = 200, 500$, with accuracy comparisons in Figure \ref{fig:accuracy_plots} validating that approximate C-DF draws converge to the true stationary distribution in time.  Excellent parameter MSE and coverage using the C-DF algorithm are reported in Table \ref{tab:lm_performance_stats}.
\begin{table}[!h]
\centering
\caption{Inferential performance for C-DF and S-MCMC for parameters of interest. Coverage and length are based on 95\% credible intervals and is averaged over all the $\beta_j$'s ($j=1,\dots,5$) and all time points and over 10 replications.  We report the time taken to produce 500 MCMC samples with the arrival of each data shard.  MSE along with associated standard errors are reported at different time points.}
\begin{tabular}{ l | c | c | c | c | c | c }
& Avg. coverage $\bbeta$ & Length & Time (sec) & \multicolumn{3}{c}{$\mathrm{MSE}= \sum_{j=1}^{p} (\hat{\beta}_t - \beta_{0})^2 / p$ }\\
\hline
& & & & $t=200$ & $t=400$ & $t=500$   \\ \cline{5-7}
\multirow{2}{*}{}
C-DF  & 1.0 & 0.60$_{0.01}$ & 95$_{4.12}$ & $0.27_{0.001}$ & $0.15_{0.001}$ &  $0.06_{0.001}$\\
S-MCMC & 1.0 & 0.60$_{0.01}$ & 119.4$_{4.64}$ & $0.12_{0.001}$ & $0.08_{0.001}$ & $0.04_{0.001}$\\
\end{tabular}
\label{tab:lm_performance_stats}
\end{table}

\begin{figure}[!h]
\centering
\includegraphics[width=0.32\columnwidth]{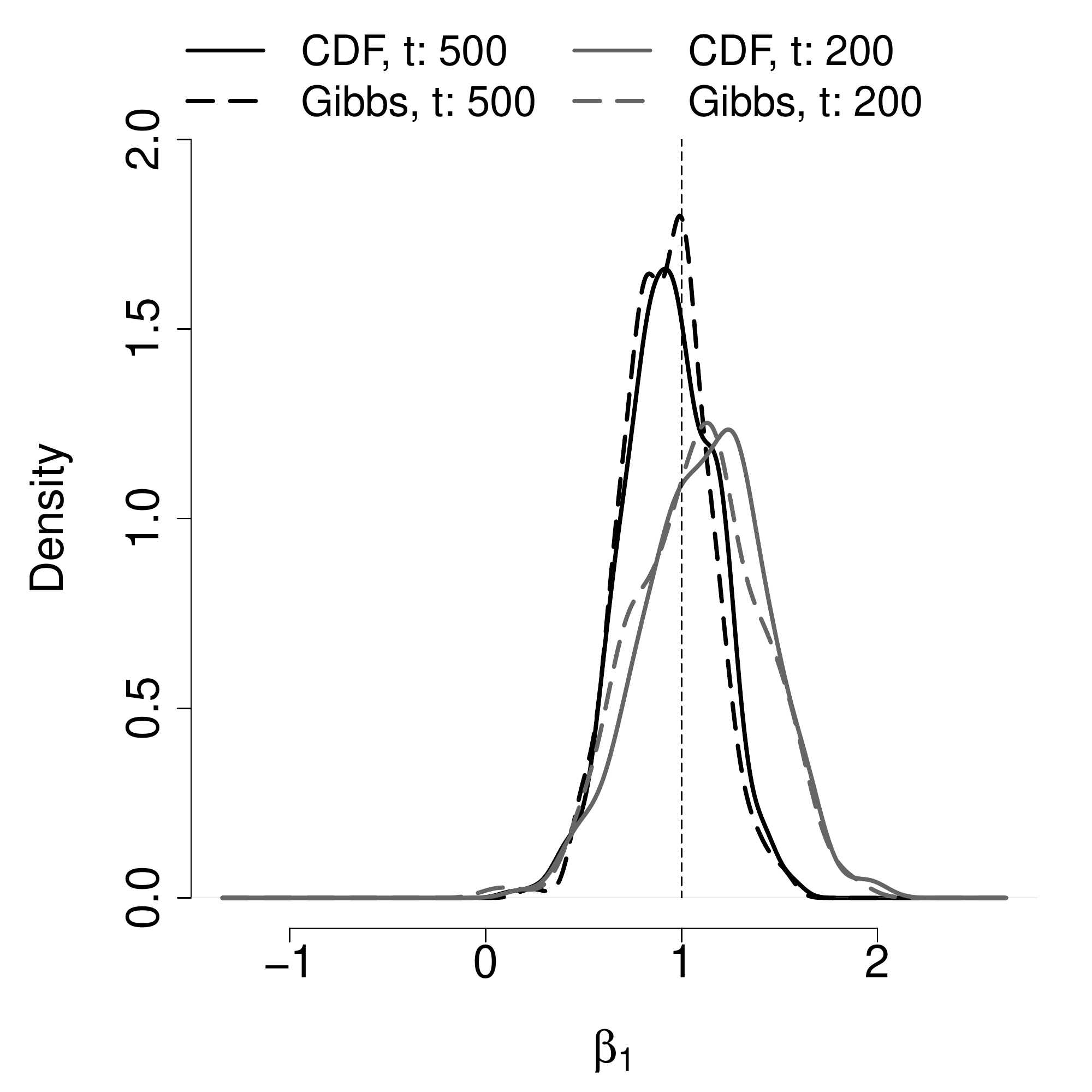}
\includegraphics[width=0.32\columnwidth]{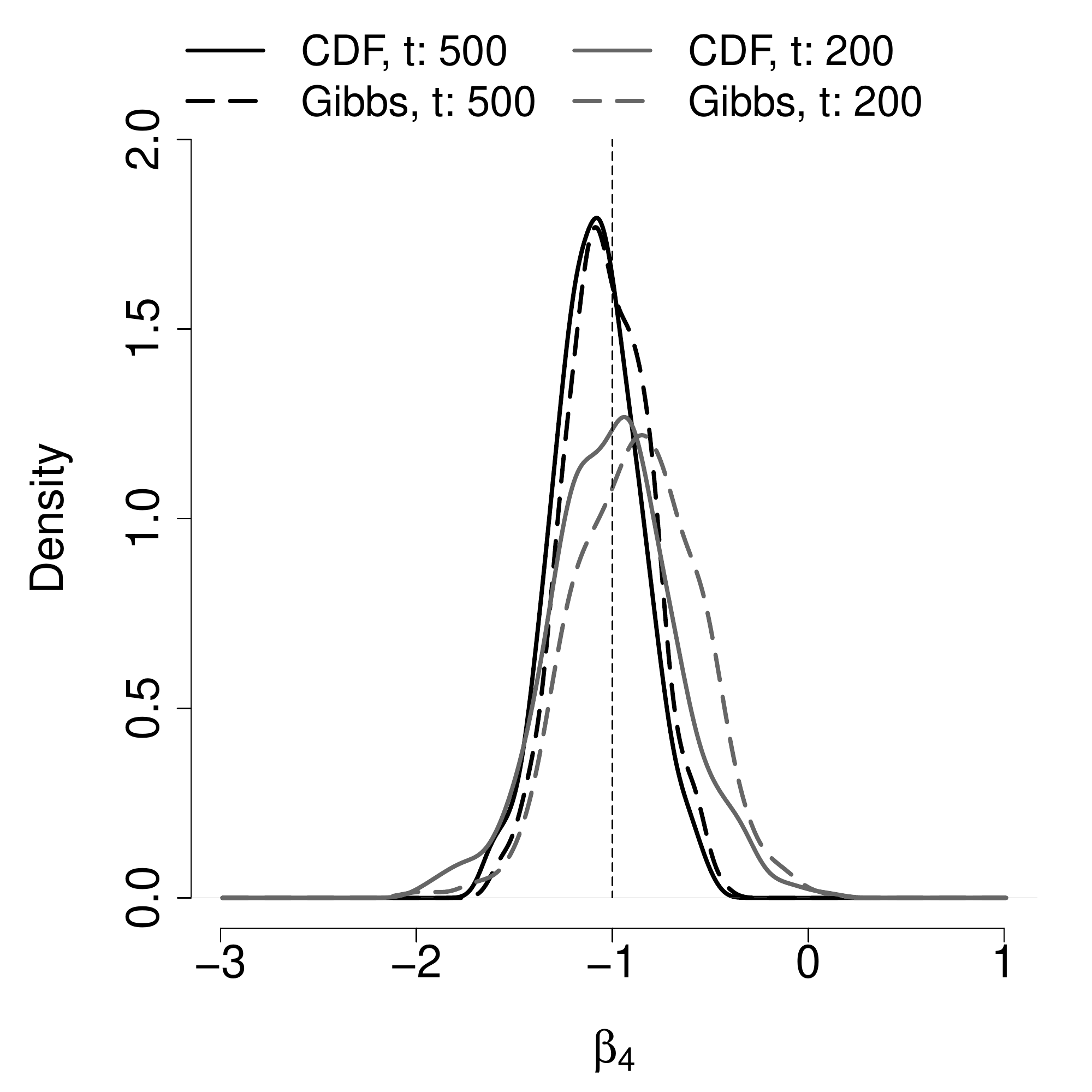}
\includegraphics[width=0.32\columnwidth]{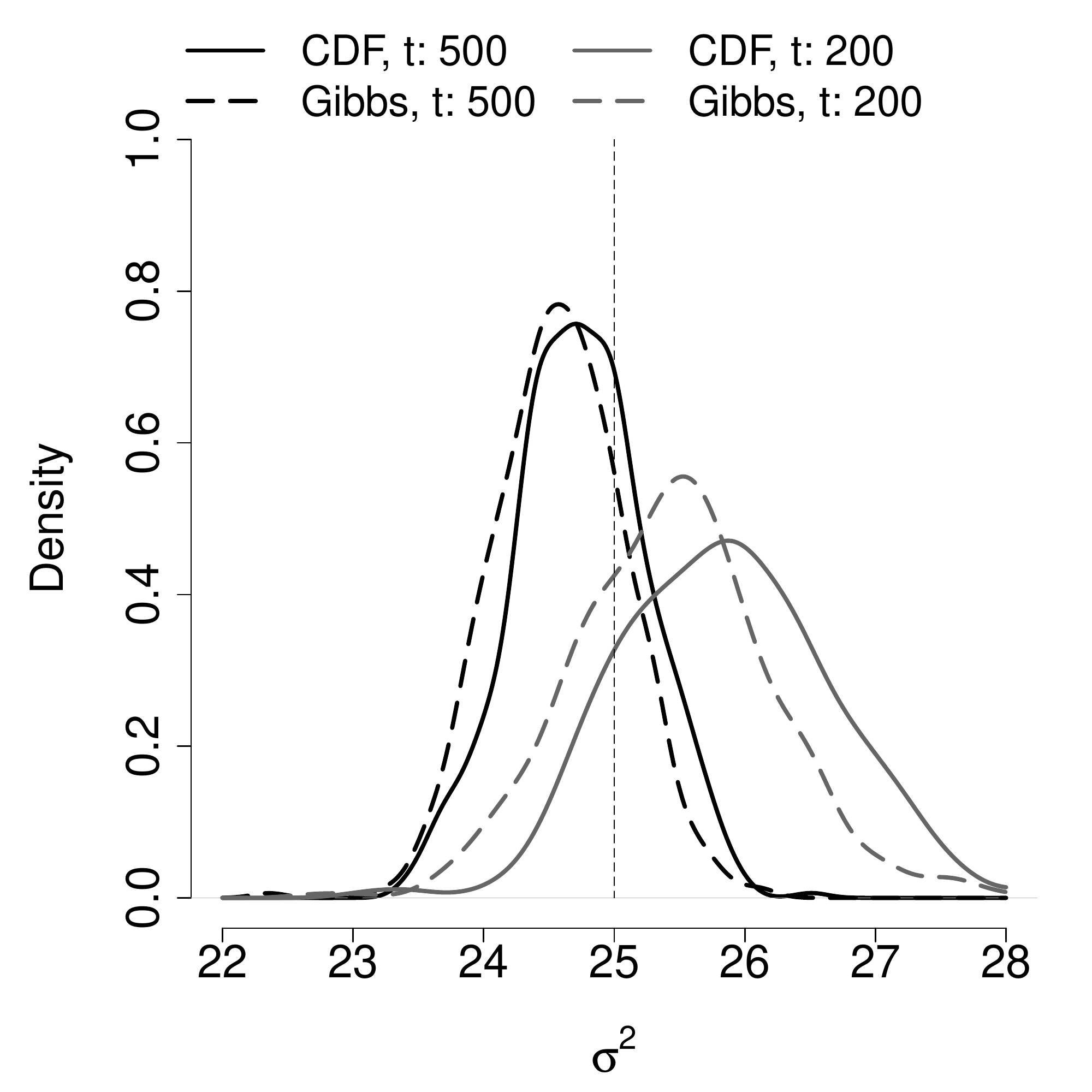} 
\caption{Kernel density estimates for posterior draws using  the C-DF algorithm and S-MCMC at $t = 200, 500$. Shown from left to right are plots of model parameters $\beta_1$, $\beta_4$, and $\sigma^2$, respectively.}
\label{fig:lm_density}
\end{figure}


\subsection{One-way Anova model} \label{sec:anova}
Consider the one-way Anova model with $k$ fixed `treatment' groups
\begin{align} \label{eqn:1anova}
\begin{aligned}
& y_{ij} = \zeta_i+\epsilon_{ij}, \quad \epsilon_{ij} \sim \mathrm{N}(0,\sigma^2) \\
& \zeta_i \sim \mathrm{N}(\mu,\tau^2), \quad i = 1, \dots, k,
\end{aligned}
\end{align}
with default priors $\pi( \mu) \,\propto\, 1, ~\tau^2 \sim \mathrm{IG}(a,b)$, and $\sigma^2 \sim \mathrm{IG}(\alpha, \beta)$.  As was the case in the linear regression model, here sufficient statistics may be propagated to yield an efficient S-MCMC sampler.  With the arrival of a new data shard at time $t$, group-specific sufficient statistics are updated as $S_i^t=S_i^{t-1} +  \sum_{j=1}^n y_{ij}^t$ and $S_i^{2(t)} = S_i^{2(t-1)} + ||\by_i^t||^2,~ i = 1, \dots, k$.  At time $t$, inference for S-MCMC proceeds by drawing from the following full conditionals distributions:
\begin{equation} \footnotesize \begin{aligned} \label{eq:gibbs_conditionals}
&\zeta_i | \sigma, \mu, \tau, \by \sim \mathrm{N}\Big(\frac{\tau^2 S_i^t+\sigma^2\mu}{nt\tau^2+\sigma^2}, \frac{\tau^2\sigma^2}{nt\tau^2+\sigma^2}\Big),
&\sigma^2 | \bzeta, \by &\sim \mathrm{IG}\Big(\alpha + \frac{nkt}{2},\beta + \frac{\sum_{i=1}^k (S_i^{2(t)} - 2\zeta_i S_i^t + nt \zeta_i^2)}{2}\Big) \\
&\mu | \bzeta, \tau \sim \mathrm{N}\Big(\frac{\sum_{i=1}^{k}\zeta_i}{k}, \frac{\tau^2}{k}\Big),
&\tau^2 | \bzeta, \mu &\sim \mathrm{IG}\Big(a+\frac{k}{2},b+\frac{\sum_{i=1}^k(\zeta_i-\mu)^2}{2}\Big).
\end{aligned} \end{equation}

For the C-DF algorithm, a natural partition suggested by the hierarchical structure of model \eqref{eqn:1anova} is $\bTheta_{\mathcal{G}_1} = \{\bzeta, \sigma^2\}$ and $\bTheta_{\mathcal{G}_2} = \{\mu, \tau^2\}$.  For this parameter partition, modified full conditionals are defined in terms of surrogate quantities as well as the previously defined group-specific sufficient statistics. Approximate inference for C-DF then proceeds as
\begin{enumerate}[(1)]
\item Observe data $\by_1^t, \dots, \by_k^t$ at time $t$.  If $t = 1$, set $\zeta_i = 0$, $\sigma = \mathrm{sd}(\vec(\by_1, \dots, \by_k))$, $\mu = \mathrm{mean}(\vec(\by_1, \dots, \by_k))$, $\tau = 1$. Otherwise, set $\hat{\mu}_t \leftarrow \hat{\mu}_{t-1}$, $\hat{\tau}_t \leftarrow \hat{\tau}_{t-1}$;
\item Update surrogate statistic $\bC_{1}^t$ component-wise as $ C_{1i}^t \leftarrow C_{1i}^{t-1}+\hat{\tau}_t^2 S_i^t, ~i = 1, \dots, k$;
\item For $s = 1, \dots, S$: draw from (a) modified full conditional $\zeta_i | \sigma, \hat{\mu}_t, \hat{\tau}_t, \bC_1^t$, $i = 1, \dots, k$; and (b) $\sigma^2 | \bzeta, \by$ as given in \eqref{eq:gibbs_conditionals}. C-DF full conditional $\zeta_i | \sigma, \hat{\mu}_t, \hat{\tau}_t, \bC_1^t$ is given by
\begin{align}
\zeta_i | \sigma, \hat{\mu}_t, \hat{\tau}_t, \bC_1^t \sim \mathrm{N}\bigg(\frac{C_{1i}^t + \sigma^2\hat{\mu}_t}{nt \hat{\tau}_t^2+\sigma^2}, \frac{\hat{\tau}_t^2 \sigma^2}{nt \hat{\tau}_t^2+\sigma^2}\bigg), ~i = 1, \dots, k;
\end{align}
\item Set $\hat{\zeta}_i \leftarrow \text{mean}(\zeta_i^{(1:S)})$, $i = 1,\dots, k$, and $\hat{\sigma}^2 \leftarrow \sigma^{2(1:S)}$. Update surrogate statistic $\bC_2^t=(C_{2,1}^t,C_{2,2}^t)$: $C_{2,1}^t \leftarrow ||\hat{\bzeta}||^2$ and $C_{2,2}^t \leftarrow \sum_{i = 1}^k \hat{\zeta}_i$;
\item For $s = 1, \dots, S$: draw from modified full conditional distributions 
(a) $\mu |\tau, \bC_2^t\sim \mathrm{N}\big(C_{2,2}^t / k, \tau^2 / k \big)$ and (b) $\tau^2 | \mu, \bC_2^t\sim \mathrm{IG}\big(a + k / 2, b + (C_{2,1}^t - 2\mu C_{2,2}^t + k\mu^2) / 2 \big)$;
\item Finally, set $\hat{\mu}_t \leftarrow \text{mean}(\mu^{(1:S)})$ and $\hat{\tau}_t \leftarrow \text{mean}(\tau^{(1:S)})$.
\end{enumerate}

\begin{figure}[!h]
\centering
\includegraphics[width=0.32\columnwidth]{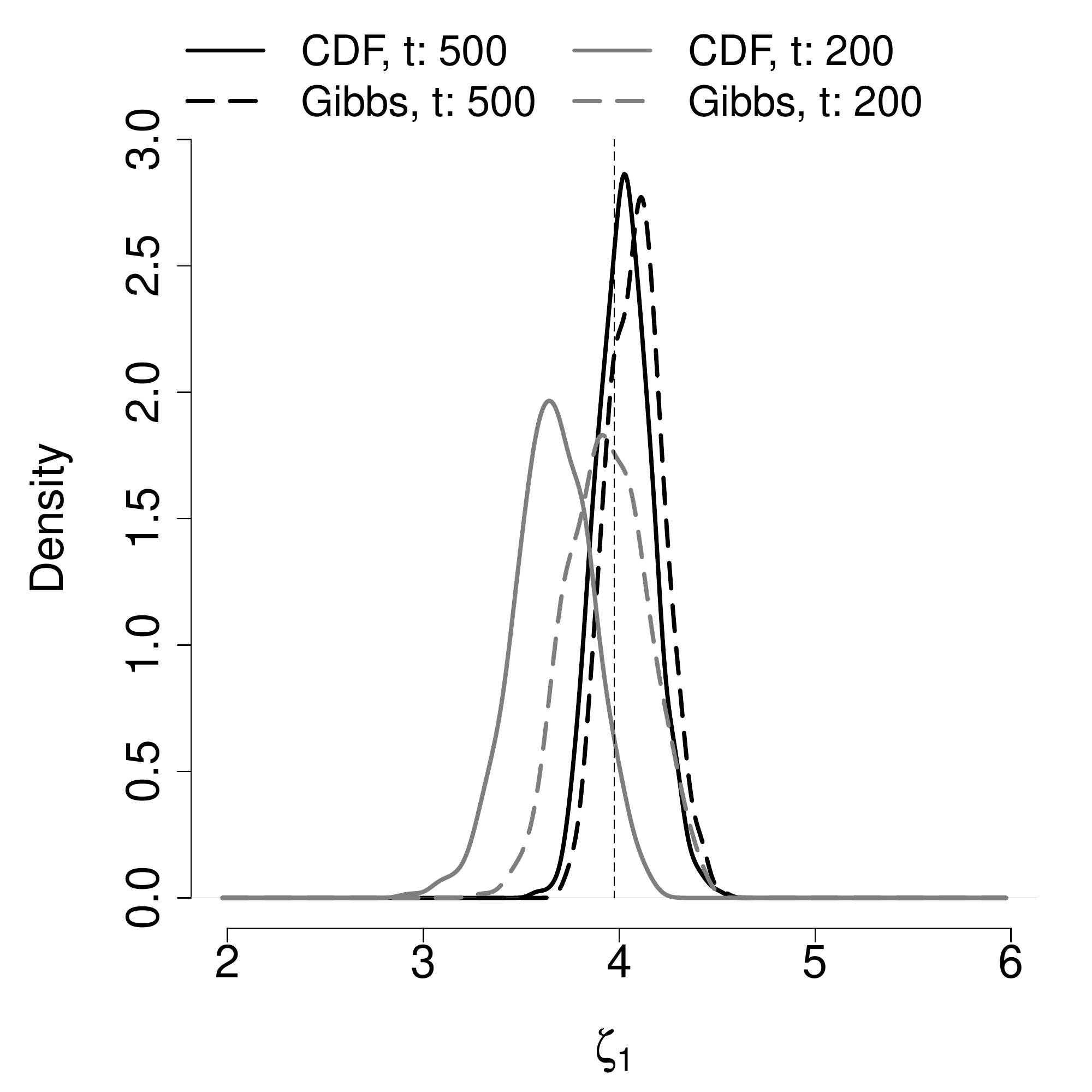}
\includegraphics[width=0.32\columnwidth]{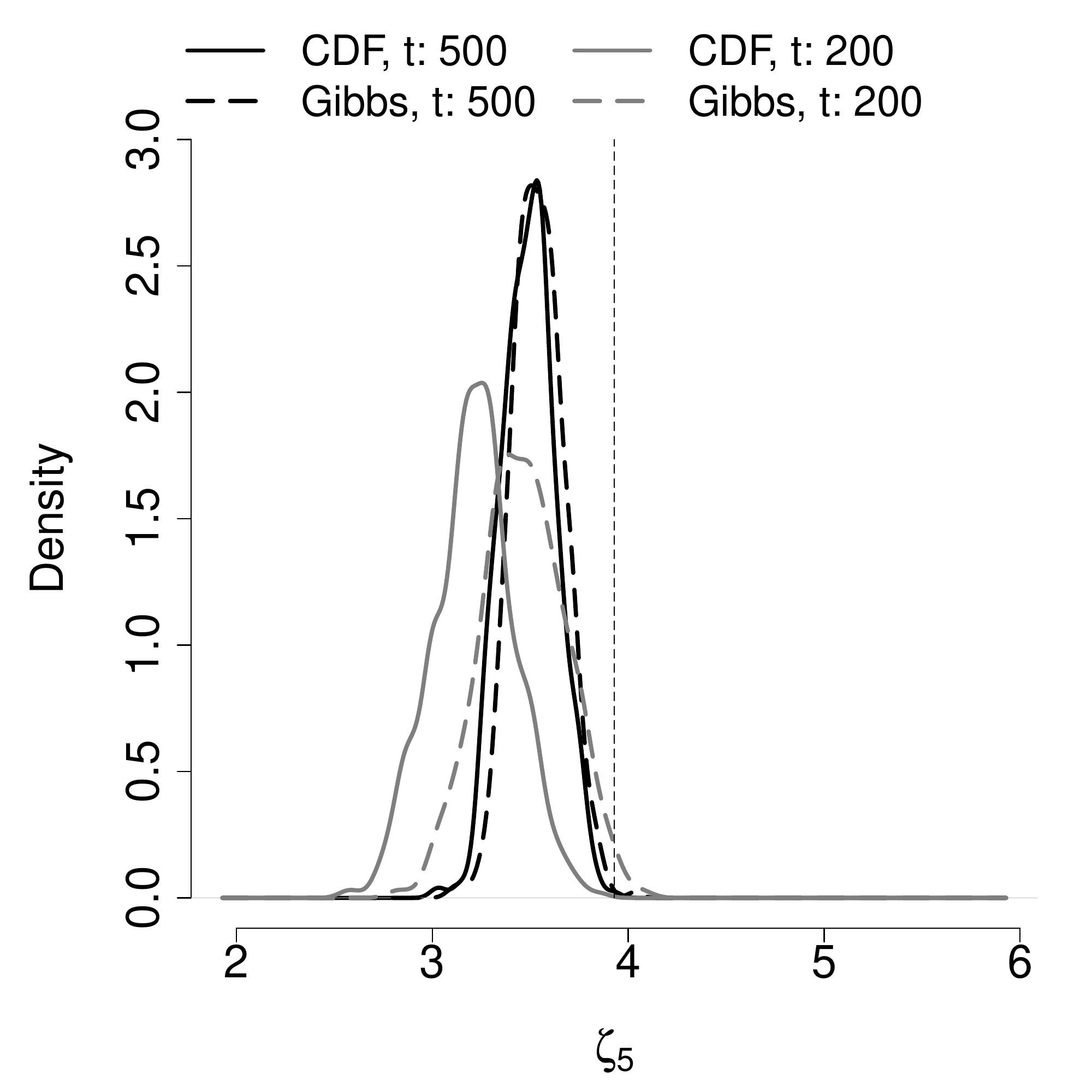}
\includegraphics[width=0.32\columnwidth]{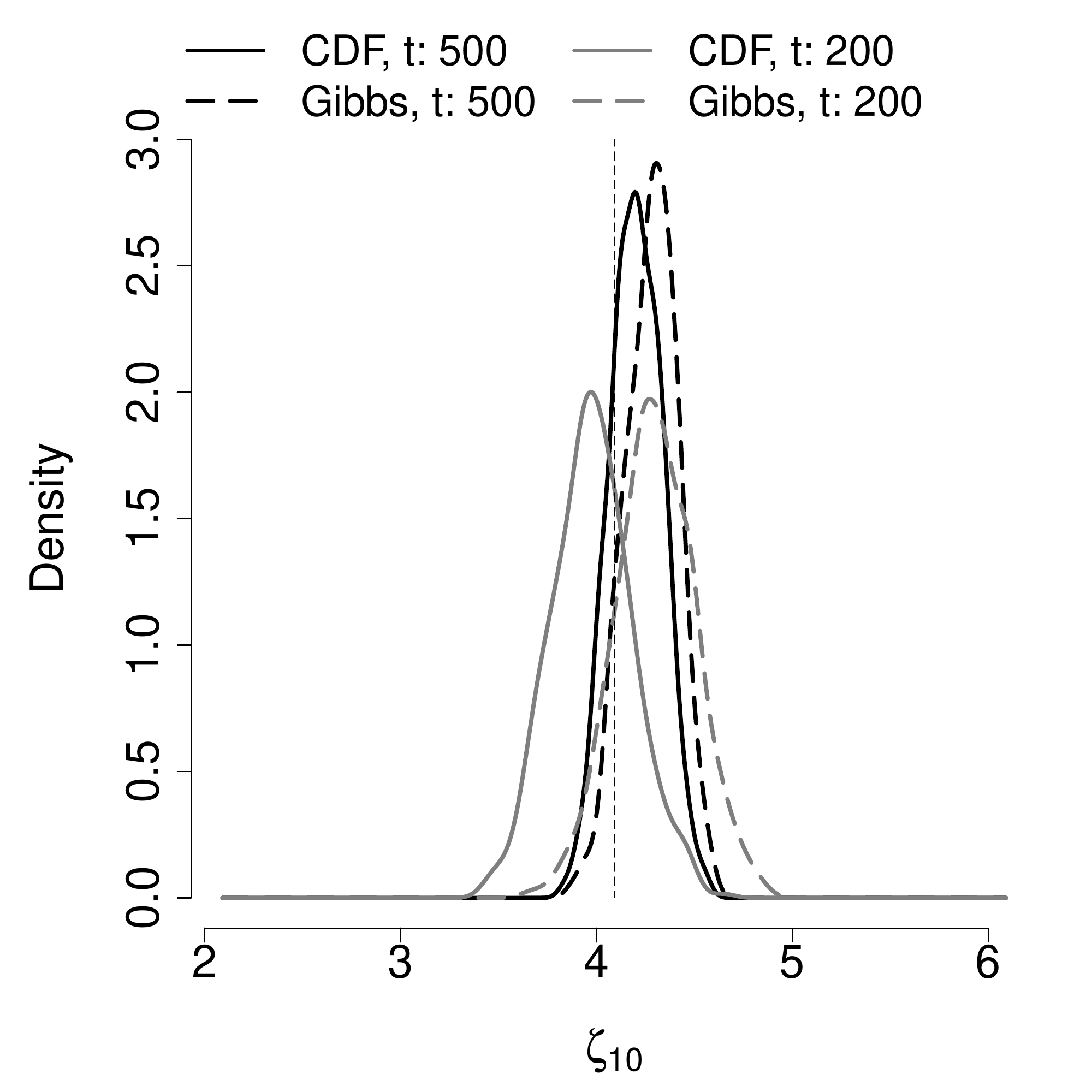} \\
\includegraphics[width=0.33\columnwidth]{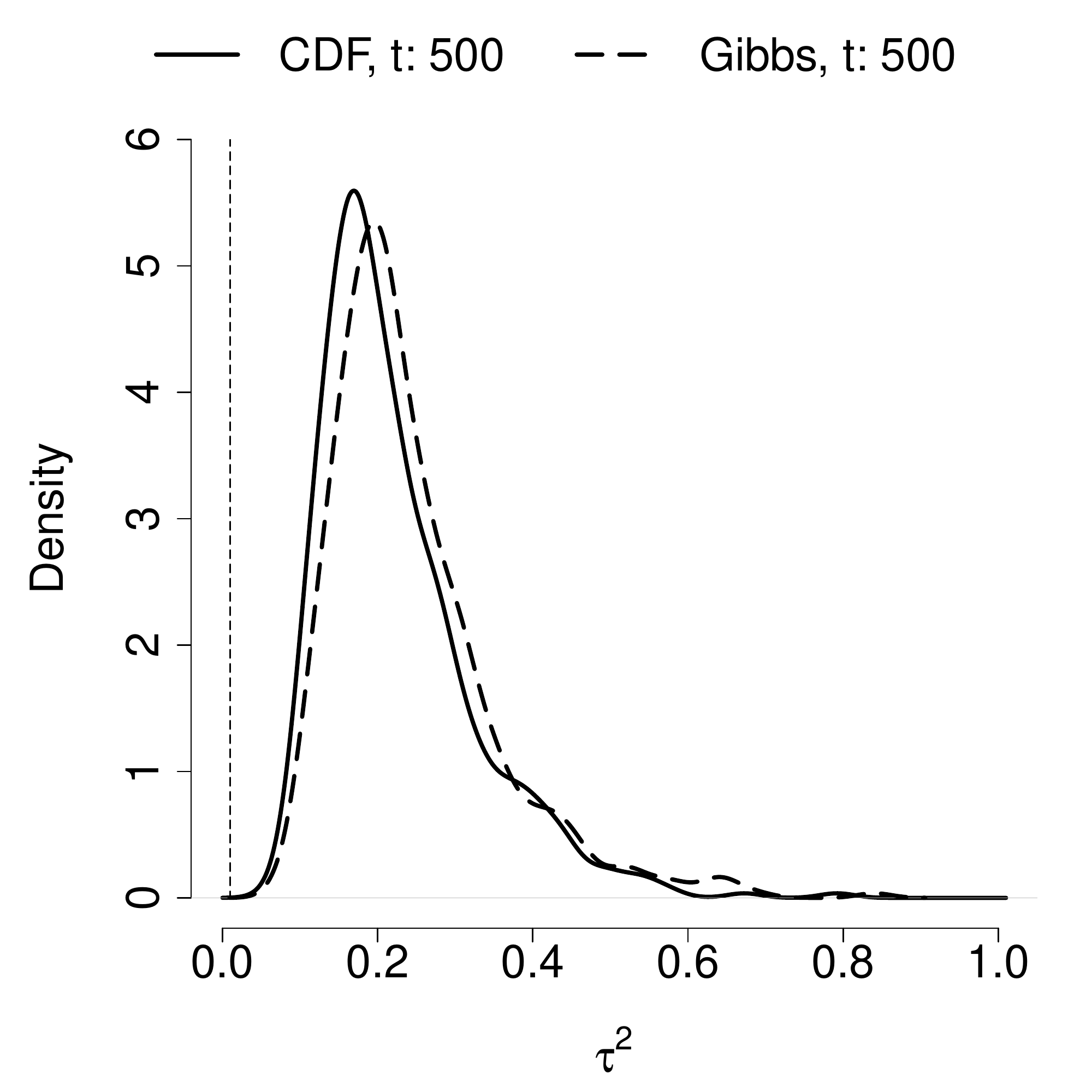}
\includegraphics[width=0.33\columnwidth]{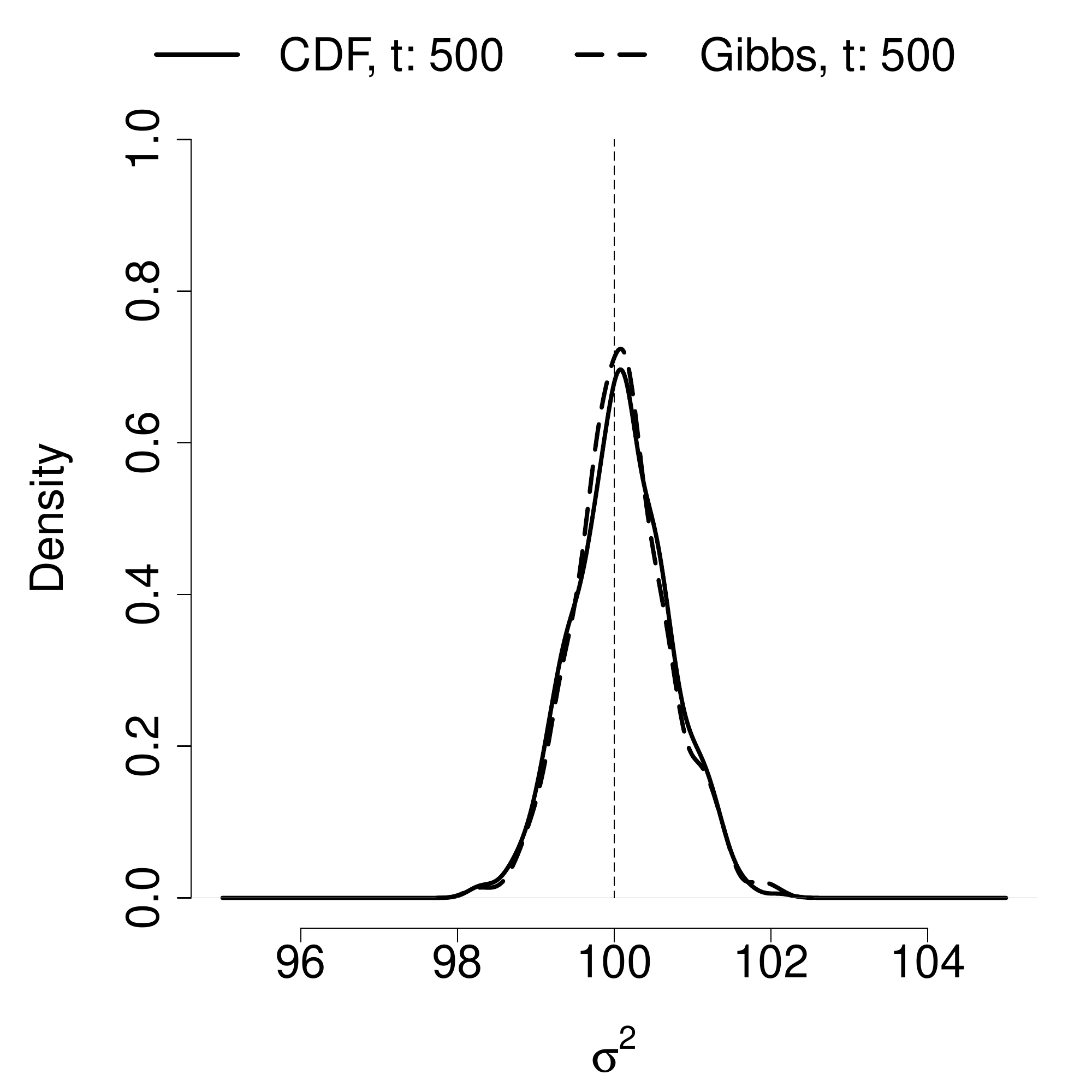}
\caption{Row \#1 (left to right): Kernel density estimates for posterior draws of $\zeta_1, \: \zeta_5, \: \zeta_{10}$ using S-MCMC and the C-DF algorithm at $t = 200, 500$; Row \#2 (left to right): Kernel density estimates for model parameters $ \tau^2, \text{ and } \sigma^2$ at $t = 500$.}
\label{fig:anova_densities}
\end{figure}

Data shards of size $n_t = 10$ are generated according to model \eqref{eqn:1anova} with parameters $\zeta_i \overset{\text{iid}}{\sim} \mathrm{N}(4, \tau^2 = 0.01)$ and $\sigma=10$. Figure \ref{fig:anova_densities} displays kernel density estimates for posterior draws using S-MCMC and the C-DF algorithm at $t = 200, 500$.  SMC \citep[SMC-CH;][]{chopin2002sequential} and ADF are added as additional competitors for this example.  To initialize ADF, the joint posterior over $(\zeta_1, \dots, \zeta_k, \log(\sigma^2)) := \btheta$ is approximated in time assuming a multivariate-normal density $\mt{\pi}_t(\btheta) \sim \mathrm{N}(\bmu_t, \bSigma_t)$.  To begin, we integrate over hyper-parameters $\mu, \tau^2$ to obtain marginal prior $\pi(\bzeta, \log(\sigma^2))$.  For $t > 1$, the approximate posterior at time $(t - 1)$ becomes the prior at $t$, and parameters $\bmu_t, \bSigma_t$ are updated using Newton-Raphson steps in the sense of \cite{mccormick2012dynamic}; in particular, $\bSigma_t = (-\nabla^2 \ell(\bmu_{t-1}))^{-1}$ and $\bmu_t = \bmu_{t-1} + \bSigma_{t-1} \nabla \ell(\bmu_{t-1})$, with  $\ell(\btheta) = \log\{ p(\by | \btheta) \pi(\btheta)\}$.

ADF is extremely sensitive to good calibration for $\sigma^2$, without which estimates for $\bzeta$ are far from the truth  even at $t = 500$.  Optimal performance is obtained by using the first data shard and performing the ADF approximation at $t = 1$ until convergence.  Thereafter,  parameters estimates are propagated as described above. Though resulting parameters  point estimates are accurate, ADF severely underestimates uncertainty as seen by the length and coverage of resulting 95\% credible intervals reported in Table \ref{tab:anova_performance_stats}.  This occurs in-part because ADF makes a global approximation on the joint posterior (restricting the propagation of uncertainty in a very specific way), whereas C-DF makes {\em local} approximations to set of full conditional distribution. The C-DF approximation results in steadily increasing accuracy along with good parameter inference as shown in Table \ref{tab:anova_performance_stats}.
\begin{table}[!h]
\centering
\caption{Inferential performance for C-DF, S-MCMC, SMC-CH and ADF for parameter $\bzeta$.  Coverage is based on 95\% credible intervals averaged over all time points, all $\bzeta$ and over 10  replications.  We report the time taken to produce 500 MCMC samples with the arrival of each data shard.  MSE along with associated standard errors are reported at different time points.}
\begin{tabular}{ l  | c |  c | c | c | c | c  }
& Avg. coverage $\bzeta$ & Length & Time (sec) & \multicolumn{3}{c}{$\mathrm{MSE}= \sum_{l=1}^k (\hat{\zeta}_t - \zeta_{0})^2 / k$}\\ \hline
& & & & $t=200$ & $t=400$ & $t=500$\\ \cline{5-7}
C-DF & $0.87_{0.09}$ & 0.53$_{0.005}$ &  $85.0_{5.25}$ & $0.77_{0.22}$ & $0.29_{0.11}$ &  $0.26_{0.15}$\\
S-MCMC  & $0.92_{0.10}$ & 0.52$_{0.002}$ & $119.4_{8.43}$ & $0.41_{0.05}$ & $0.22_{0.14}$ &  $0.20_{0.16}$\\
SMC-CH & $0.70_{0.26}$ & $0.21_{0.06}$ & $243.06_{4.80}$ & $0.93_{0.30}$ & $0.18_{0.09}$ & $0.10_{0.04}$\\
ADF  & $0.36_{0.23}$ & $0.08_{0.02}$ & $0.88_{0.01}$ & $0.42_{0.18}$ & $0.28_{0.12}$ & $0.27_{0.11}$ \\
\end{tabular}
\label{tab:anova_performance_stats}
\end{table}

\subsection{Advanced data models}\label{sec:complex_examples}
The C-DF algorithm introduced in Section \ref{sec:cdf_alg} modifies the set of full conditional distributions to depend on propagated surrogate quantities which yields an approximate kernel.  This enables efficient online MCMC, with guarantees on correctness of C-DF samples as data accrue over time (see Section \ref{sec:theory_cdf}).   We present extensions of the C-DF algorithm to models in settings where (a) the model involves an increasing number of parameters growing with the sample size, or (b) some of the conditional distributions are not in closed form.

Section \ref{sec:DLM_CDF} presents an application to a dynamic linear model for a continuous response centered on a first-order auto-regressive process. The `forward-filtering and backward-sampling' Kalman filter updates for the latent process enables online posterior computation, albeit with an increasing computational cost as time goes on. Section \ref{sec:probit} considers an application to binary regression where the conditional posterior distribution over the regression coefficients does not assume a closed form.  For logistic regression, a variational approximation to the posterior introduces additional variational parameters for each observation to obtain a lower-bound for the likelihood \citep{jaakkola1997variational}.  Additional non-conjugate message passing approximations have been considered in \cite{braun2010variational} and \cite{tan2014stochastic}.  One may also resort to ADF using a Laplace approximation to the posterior over regression coefficients and propagating associated mean and covariance parameters in time.  However, the latter are known to severely underestimate parameter uncertainty.  Fortunately, data augmentation via the probit model enables conditionally conjugate sampling. In both illustrations, we discuss how to use the C-DF algorithm to overcome the computational and storage bottlenecks in an increasing parameter setting.

\subsubsection{Dynamic Linear Model}\label{sec:DLM_CDF}
We consider the first-order dynamic linear model \citep{west1997bayesian}, namely
\begin{align*}
y_{t+1} \sim \mathrm{N}(\theta_{t+1},\sigma^2), \quad
\theta_{t+1} \sim \mathrm{N}(\phi\,\theta_t,\tau^2)
\end{align*}
where noisy observations $y_t, \: t \ge 1$ are modeled as arising from an underlying stationary AR(1) process with lag parameter $\phi, \: |\phi| < 1$.  Default priors $\sigma^2 \sim \mathrm{IG}(a_0,b_0), \: \tau^2\sim \mathrm{IG}(c_0,d_0), \: \phi \sim \mathrm{U}(-1,1)$ are chosen to complete the hierarchical model, and assume $\theta_0 \sim \mathrm{N}(0, h_0)$, the stationary distribution for the latent process. The full conditional distributions are given by
\begin{align*}
&[\theta_{t+1}| -] 
\sim \mathrm{N}\left(\frac{\sigma^{-2}y_{t+1} + \tau^{-2}\phi\theta_t}{\sigma^{-2} + \tau^{-2}}, \frac{1}{\sigma^{-2}+\tau^{-2}}\right), \\
&[\theta_s | -] 
\sim \mathrm{N}\left(\frac{\sigma^{-2}y_{s} + \tau^{-2}\phi(\theta_{s-1} + \theta_{s+1})}{\sigma^{-2} + (\phi^2+1)\tau^{-2}}, \frac{1}{\sigma^{-2}+(\phi^2+1)\tau^{-2}}\right), ~ 1 \le s \le t,\\
&[\sigma^2| -] 
\sim \mathrm{IG}(a_t,b_t), \quad a_{t+1} = a_t + 1/2, \quad b_{t+1}=b_t + (y_{t+1}-\theta_{t+1})^2 / 2, \\
&[\tau^2 | -] 
 \sim \mathrm{IG}(c_t,d_t), \quad c_{t+1} = c_t + 1/2, \quad d_{t+1}=d_t + (\theta_{t+1}-\phi\theta_t)^2/2, \\
&[\phi | -] 
\,\propto\, \pi(\phi) \: \mathrm{N}\left(\frac{\sum_{s=1}^{t}\theta_s\theta_{s-1}}{\sum_{s=1}^{t}\theta_{s-1}^2},\frac{\tau^2}{\sum_{s=1}^{t-1}\theta_s^2}\right) \, I(|\phi| < 1).
\end{align*}

Forecasting future trajectories of the response is a common goal, hence good characterization of the distribution over $\theta_t, \phi, \tau^2$ is of interest.   Hence, retrospective sampling of $\{\theta_s, s < t\}$ is only meaningful in as much as it propagates uncertainty to the current time point.  Forward-filtering and backward-sampling using Kalman filtering updates for the latent process enables online posterior computation, but this scales poorly as the time horizon grows.

To extend the use of the C-DF algorithm in this growing parameter setting, we propose sampling of the latent process over a moving time-window of size $b$.   This eliminates the propagation of errors that might otherwise result from poor estimation at earlier time points.  As the sampling window shifts forward, trailing latent parameters are fixed at their most recent point estimates.  Parameter partitions for the C-DF algorithm must therefore also evolve dynamically, and are defined at time $t$ as $\bTheta_{\mathcal{G}_1}=\{\theta_t, \dots, \theta_{t-b+1},\tau^2,\sigma^2,\phi\}$, $\bTheta_{\mathcal{G}_2}=\{\theta_{t-b}, \dots, \theta_1\}$, $t>b$.  Unlike previous examples, conditional distribution $\pi(\phi | -)$ is not available in closed-form.  Nevertheless, propagated surrogate quantities (SCSS) enable approximate MCMC draws to be sampled from C-DF full conditional $\mt{\pi}(\phi | -)$ via Metropolis-Hastings.  Here, steps for approximate MCMC using the C-DF algorithm in the context of a Metropolis-within-Gibbs sampler are:
\begin{enumerate}[(1)]
\item At time $t$ observe $y_t$;
\item If $t\leq b$ : Draw $S$ samples for $(\theta_1,\dots,\theta_t,\tau^2,\sigma^2,\phi)$ from the Gibbs full conditionals. In addition, $C_1^t=C_2^t=C_3^t=C_4^t=0, ~ t\leq b$.
\item If $t>b$ : Repeat $S$ times the following
\begin{enumerate}[(a)]
\item for $t-b < s < t$, draw sequentially from $[\theta_s | \theta_{s-1},\theta_{s+1}, - ]$ and finally from $[\theta_{t} | \theta_{t-1}, -]$, noting that $\theta_{t-b} = \hat{\theta}_{t-b}$;
\item draw $\tau^2 \sim \mathrm{IG}\big(a_t, b_t \big)$, and $b_t = C_4^{(t-1)}-2\phi C_3^{(t-1)}+\phi^2 C_2^{(t-1)} + \frac{1}{2}\big((\theta_{t-b+1}-\phi\hat{\theta}_{t-b})^2 + \sum_{j=2}^{b}(\theta_{t+j-b}-\phi\theta_{t+j-b-1})^2\big)$;
\item draw $\sigma^2 \sim \mathrm{IG}\big(a_t, C_1^{(t-1)} + \frac{1}{2}\sum_{j=1}^{b}(y_{t+j-b}-\theta_{t+j-b})^2\big)$;
\item sample $\phi \sim \mt{\pi}(\cdot | C_3^{(t-1)}, C_4^{(t-1)}, \theta_{(t-b+1):t})$ via a Metropolis-Hastings step.
\end{enumerate}
\item Set $\hat{\theta}_{t-b} \leftarrow \mathrm{mean}(\theta_{t-b})$, and then update surrogate quantities $C_1^t \leftarrow C_1^{(t-1)} + \frac{1}{2}(y_{t-b}-\hat{\theta}_{t-b})^2$, $C_2^t \leftarrow C_2^{(t-1)} + \frac{\hat{\theta}_{t-b-1}}{2}$, $C_3^t \leftarrow C_3^{(t-1)} +\hat{\theta}_{t-b-1}\hat{\theta}_{t-b}$, $C_4^t=C_4^{(t-1)}+ \frac{\hat{\theta}_{t-b}^2}{2}$.
\end{enumerate}

\begin{figure}[!h]
\centering
\includegraphics[width=0.32\columnwidth]{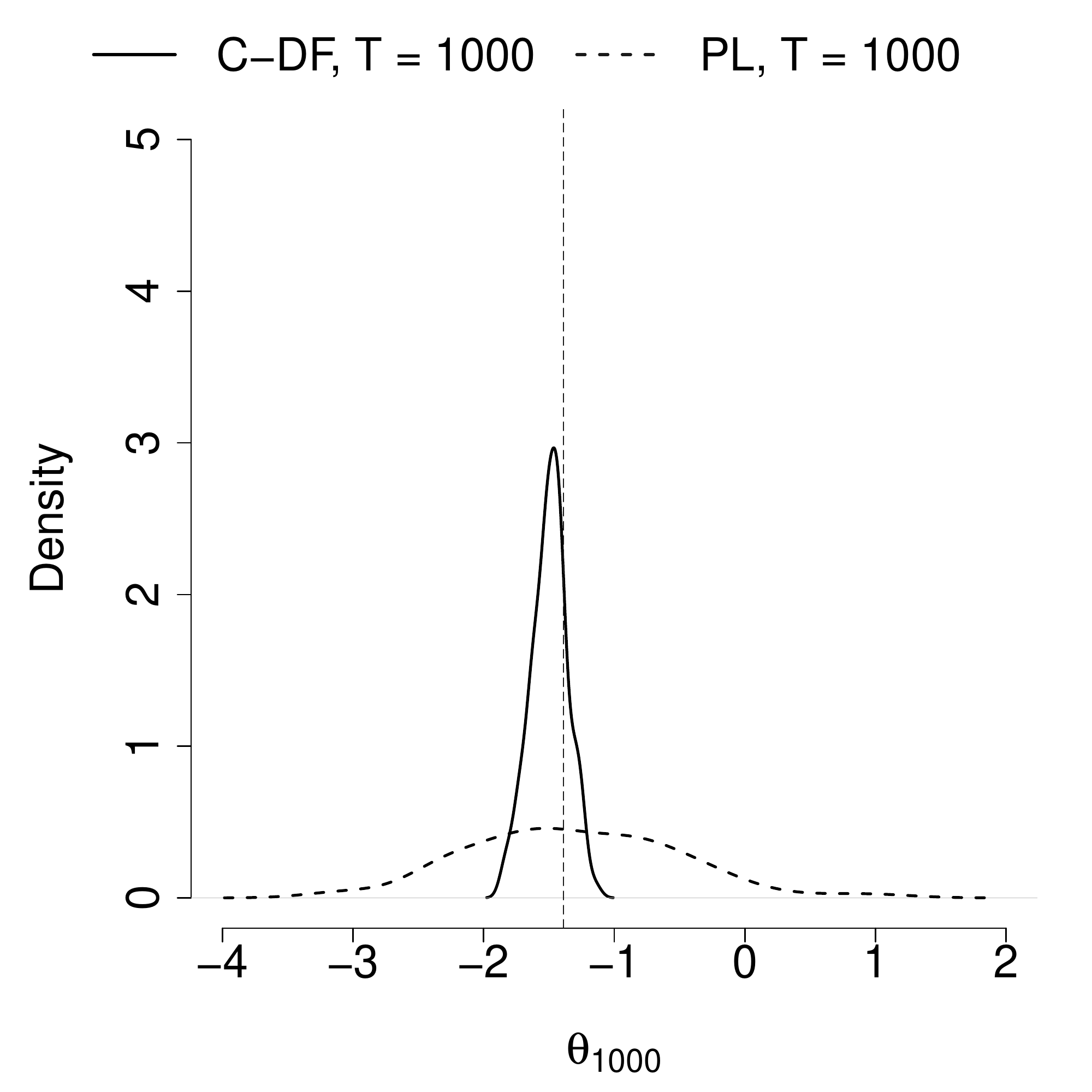}
\includegraphics[width=0.32\columnwidth]{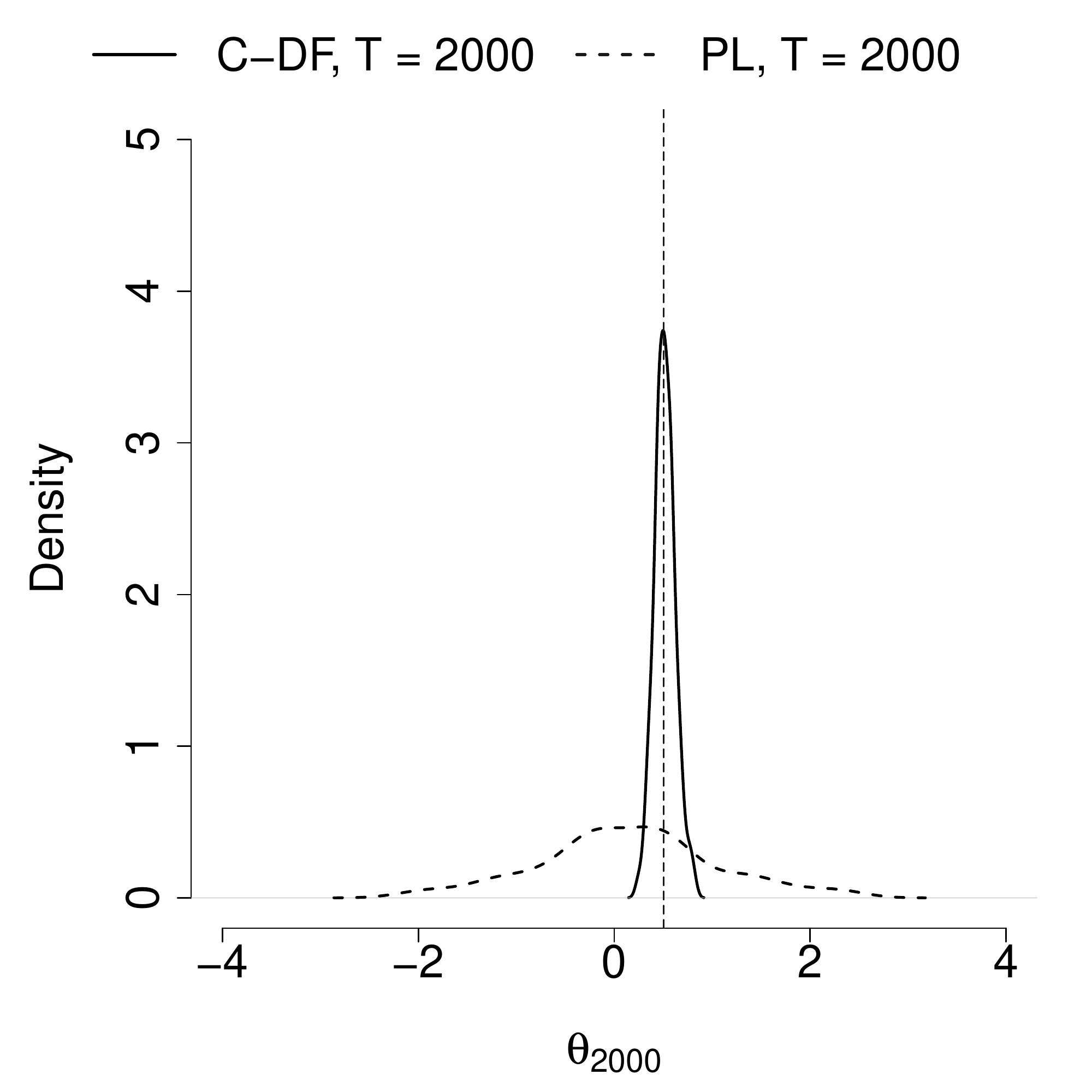}
\includegraphics[width=0.32\columnwidth]{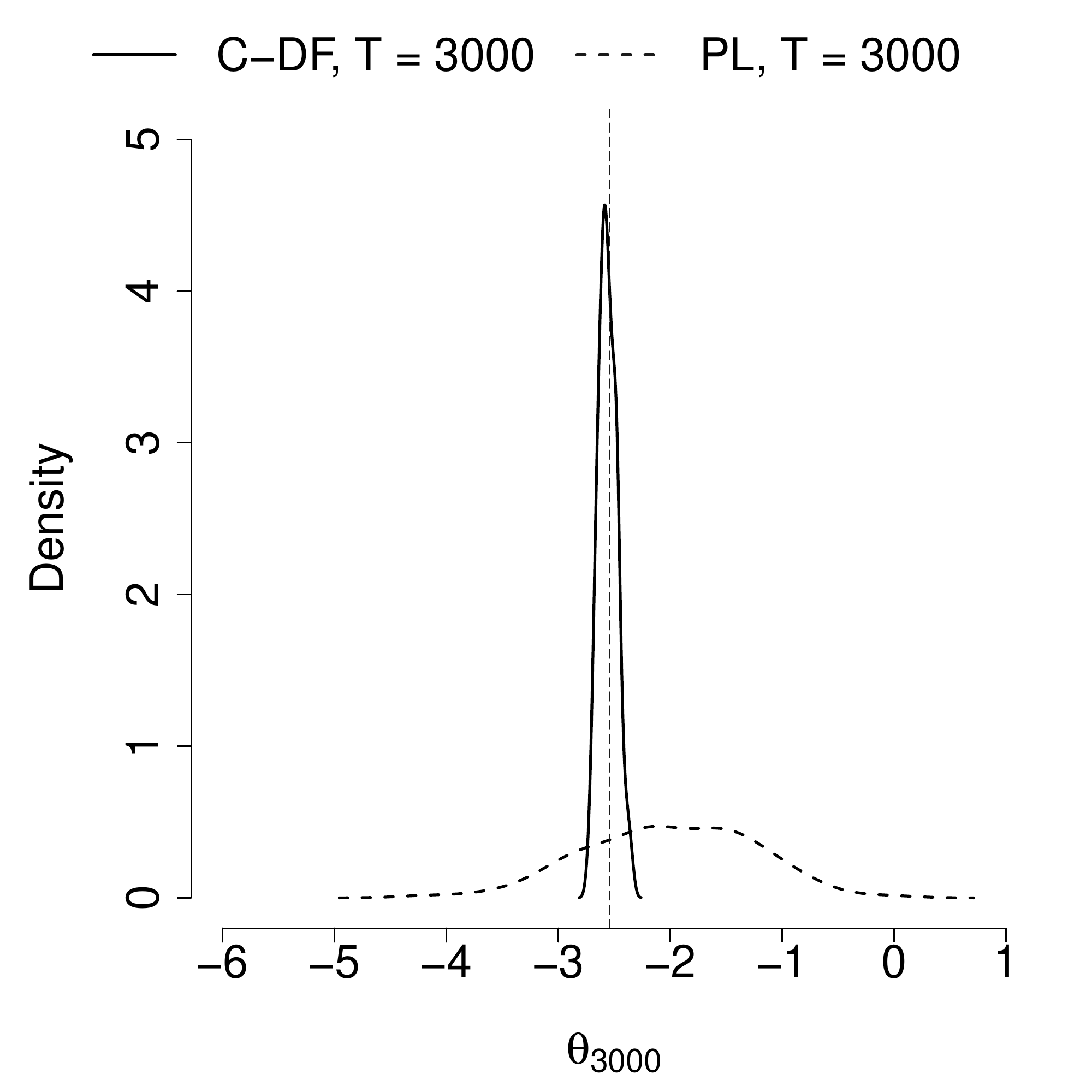} \\
\includegraphics[width=0.32\columnwidth]{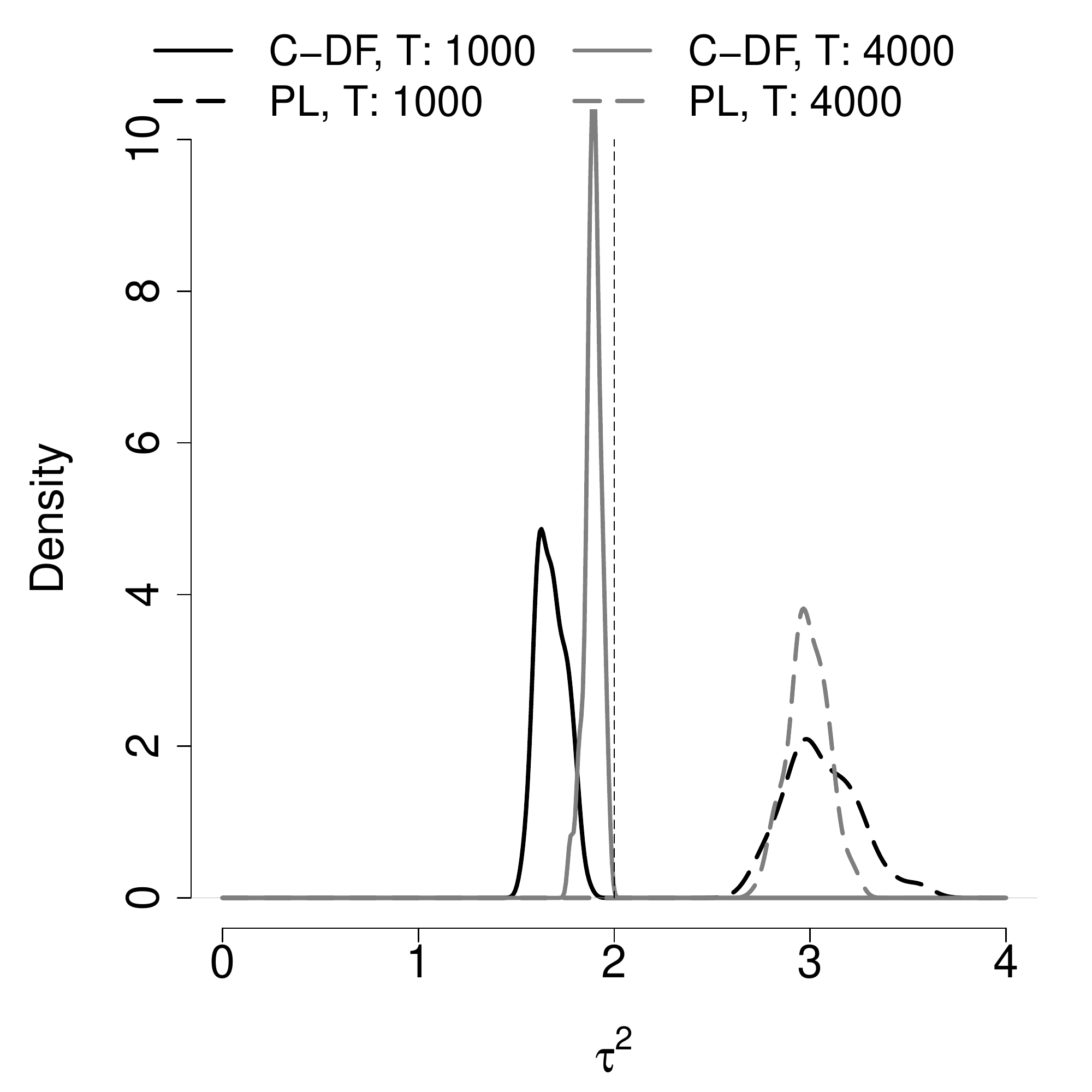}
\includegraphics[width=0.32\columnwidth]{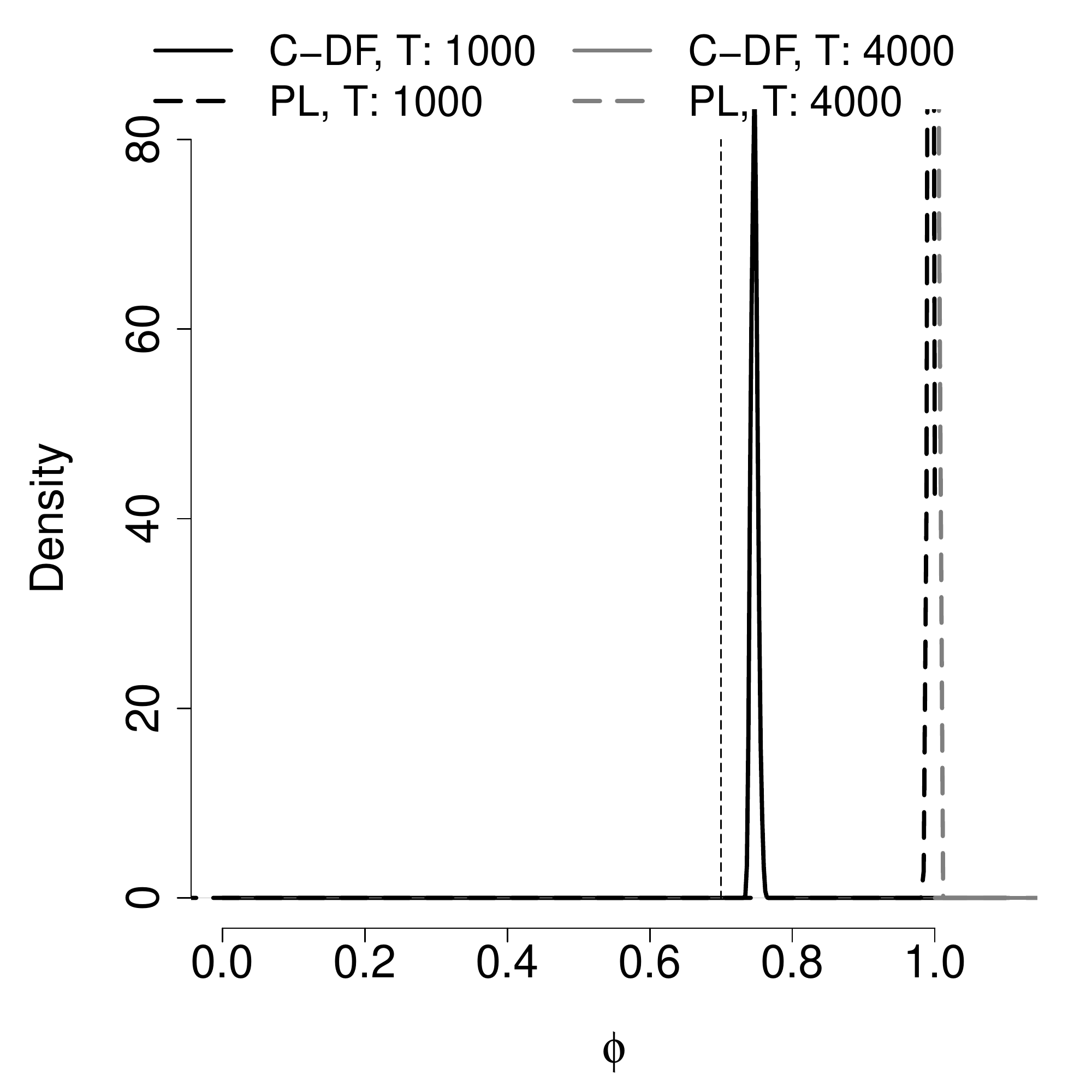}
\caption{Row \#1 (left to right):  Kernel density estimates for posterior draws of $\theta_t$ using PL and the C-DF algorithm at $t = 1000, 2000, 3000$; Row \#2 (left to right) plots of model parameters $\tau^2$ and $\phi$, respectively.}
\label{fig:dlm_chutiya}
\end{figure}

We vary signal-to-noise ratio $\tau^2 / \sigma^2$ for the data generating process in simulation experiments.  A number of high-signal cases were examined, and results are reported  for a representative case with $\tau = \sqrt{2}, \sigma = 0.1$.
A sufficiently large window size $b$ is necessary to prevent  C-DF from suffering due to poor estimation at early time-points, causing propagation of error in the defined surrogate quantities.  For the  competitor, 100 particles were propagated in time, and hence we fix $b = 100$ as well.
Kernel density estimates for $\theta_t$ using Particle learning \citep[PL;][]{lopes2011particle} and the C-DF algorithm are shown in Figure \ref{fig:dlm_chutiya}.  In the case of C-DF,  estimates for all model parameters are found to be concentrated near their true values, whereas for PL, $\theta_t$ at different times are centered correctly, albeit with much higher posterior variance (presumably due to poor estimation of noise parameters $\tau^2, \sigma^2$).
In addition to being 35\% faster than PL, average coverage for the latent AR(1) process using the C-DF algorithm is near 80\% with credible intervals roughly 10 times narrower than PL (see Table \ref{tab:performance_stats}).  C-DF is substantially more efficient in terms of memory and storage utilization, requiring only 6\% of what PL uses for an identical simulation.  Finally, C-DF produces accurate estimates for latent parameters $\tau^2, \phi$ (in contrast to PL), although their posterior spread appears somewhat over shrunk.  This may be remedied by using larger window size $b$ (at the expense of increased runtime) depending on the task at hand.  

\begin{table}[h]
\centering
\caption{Inferential performance for C-DF and Particle Learning (PL).  Coverage and length are based on 95\% credible intervals for $\theta_t$ averaged over all time points and 10  replications.  For truth $\theta_{t0}$ at time $t$, we report $\mathrm{MSE} = \tfrac{1}{Tn} \sum_{t=1}^{Tn} (\hat{\theta}_t - \theta_{t0})^2$.  We report the time taken to run C-DF with $50$ Gibbs samples at each time for $\tau^2,\btheta,\sigma^2$ and $500$ MH samples for $\phi$.}
\begin{tabular}{ l | c | c | c | c }
& Avg. coverage $\btheta$ & Length & Time (sec) & MSE \\
\hline
C-DF  & $0.78_{0.10}$ & $0.33_{0.11}$ & $1138.60_{0.10}$ & $0.011_{0.001}$ \\
PL & $1_{0.00}$ & $3.36_{0.46}$ & $1750.58_{0.10}$ & $0.096_{0.027}$
\end{tabular}
\label{tab:performance_stats}
\end{table}

\begin{table}[h]
\centering
\caption{Computational and storage requirements for the Dynamic Linear Model using C-DF and PL.  $C_{i,j}^{t}$, is the $i$-th CSS corresponding to the $j$-th particle in PL, $i=1:4, \: j=1:N$, $N = 100$ is the number of particles propagated by PL, and $G = 500$ is the number of Metropolis samples used by both PL and C-DF.  Memory used to store and propagate SCSS and CSS for C-DF and PL is reported. Sampling and update complexities are in terms of big-O.}
\begin{tabular}{ l | c | c | c | c | c }
& Stats & Data & Sampling & Updating & Memory (bytes) \\ \hline
C-DF & $C_i^{t}$ & $\{y_i\}_{i > nt-b}$ & $S(N + G)$ flops & $N$ flops & 128 \\
PL & $C_{i,j}^{t}$ & $\{y_i\}_{i \ge 1}$ &  $NG$ flops & $N$ flops & 3330
\end{tabular}
\label{tab:dlm_chutiyapa}
\end{table}

\subsubsection{An application to binary response data}
\label{sec:probit}

We consider an application of the C-DF algorithm for the probit model $\mathrm{Pr}(y_i = 1|{\bf x}_i) = \Phi( \bx_i' \bbeta)$, with standard normal distribution function $\Phi(\cdot)$.   The \cite{albert1993bayesian} latent variable augmentation applies Gibbs sampling, with $y_i = \sgn(z_i)$, and $z_i | \bbeta \sim \mathrm{N}(\bx_i' \bbeta, 1)$.   Assuming a conditionally conjugate prior $\bbeta \sim \mathrm{N}( \bzero, \bSigma_\bbeta)$, the Gibbs sampler alternates between the following steps:
\begin{inparaenum}[(1)]
\item conditional on latent variables $\bz = (z_1, z_2, \dots)'$, sample $\bbeta$ from its $p$-variate Gaussian full conditional; and
\item conditional on $\bbeta$ and the observed binary response $\by$, impute latent variables from their truncated normal full conditionals.
\end{inparaenum}
However, imputing latent variables $\{ z_i : i \ge 1 \}$ presents an increasing computational bottleneck as the sample size increases, as does recalculating the conditional sufficient statistics (CSS) for $\bbeta$ given these latent variables at each iteration.
C-DF alleviates this problem by imputing latent scores for the last $b$ training observations and propagating surrogate quantities. ``Budget'' $b$ is allowed to grow with the predictor dimension, and as a default we set $b = p \log p $.  For data shards of size $n$, define $\mathcal{I}_t = \{i > nt - b\}$ as an index over the final $b$ observations at time $t$.  Parameter partitions in this setting are dynamic  as in Section \ref{sec:DLM_CDF}, with $\bTheta_{\mathcal{G}_1}=\{\bbeta; z_i, \: i \in \mathcal{I}_t\}$ and $\bTheta_{\mathcal{G}_2} = \{z_i : i \leq nt-b\}$. The C-DF algorithm proceeds as follows:
\begin{enumerate}[(1)]
\item Observe data $\bX^t, \by^t$ at time $t$;
\item If $t = 1$, set $\bbeta = 0$, and draw $z_i \sim \mathrm{N}(0, 1), ~i = 1, \dots, n$. If $t \le b / n$, $\bC_t = \bzero$ and draw $S$ Gibbs samples for $(z_1, \dots, z_{nt}, \bbeta)$ from the full conditionals.
\item If $t > b / n$, set $\bbeta \leftarrow \hat{\bbeta}_{t-1}$, update sufficient statistic $\bS_t^{XX} \leftarrow \bS_{t-1}^{XX} + \bX^{t'} \bX^t$, and compute $\bSigma_t^{XX} =  (\bS_t^{XX} + \bSigma_\bbeta^{-1})^{-1}$;
\item For $s = 1, \dots, S$: \begin{inparaenum}[(a)]
\item draw $z_i \sim \mathrm{TN}(\bx_i' \bbeta),~ i \in \mathcal{I}_t$, and define $\bz_{\mathcal{I}_t}$ and $\bX_{\mathcal{I}_t}$ respectively as the collection of latent draws and data for moving window $b$;
\item draw $\bbeta | \bz_{\mathcal{I}_t}^{(s)}, \bC_{t-1} \sim \mathrm{N}( \bSigma_t^{XX} \bC_t, \bSigma_t^{XX})$, where $\bC_t(\bz_{\mathcal{I}_t}^{(s)}) = \bC_{t-1} + \bX_{\mathcal{I}_t}' \bz_{\mathcal{I}_t}^{(s)}$;
\end{inparaenum}
\item Set $\hat{\bbeta}_t \leftarrow \mathrm{mean}(\bbeta^{(s)})$.  For $u_i(\bbeta) = y_i \phi(-\bx_i' \bbeta) / \Phi(y_i \bx_i' \bbeta)$, the trailing $n$ latent scores within moving window $b$ are fixed at their expected value, namely $\hat{z}_i \leftarrow \bx_i' \hat{\bbeta}_t + u_i(\hat{\bbeta}_t), ~i \in \mathcal{I}_t^\mathrm{out} = \{(nt+b-1) : (n(t+1)+b-1)\}$;
\item Denote $\bX_{\mathcal{I}_t^\mathrm{out}}$ and $\bz_{\mathcal{I}_t^\mathrm{out}}$ as predictor data and latent scores for the ``outgoing'' set of data indexed by $\mathcal{I}_t^\mathrm{out}$.  Update surrogate CSS $\bC_t \leftarrow \bC_{t-1} + \bX_{\mathcal{I}_t^\mathrm{out}}' \hat{\bz}_{\mathcal{I}_t^\mathrm{out}}$, and set $\bTheta_{\mathcal{G}_1} \leftarrow  \bTheta_{\mathcal{G}_1} \cup \hat{\bz}_{\mathcal{I}_t^\mathrm{out}}$.
\end{enumerate}

\begin{figure}[!h]
\centering
\includegraphics[width=0.32\columnwidth]{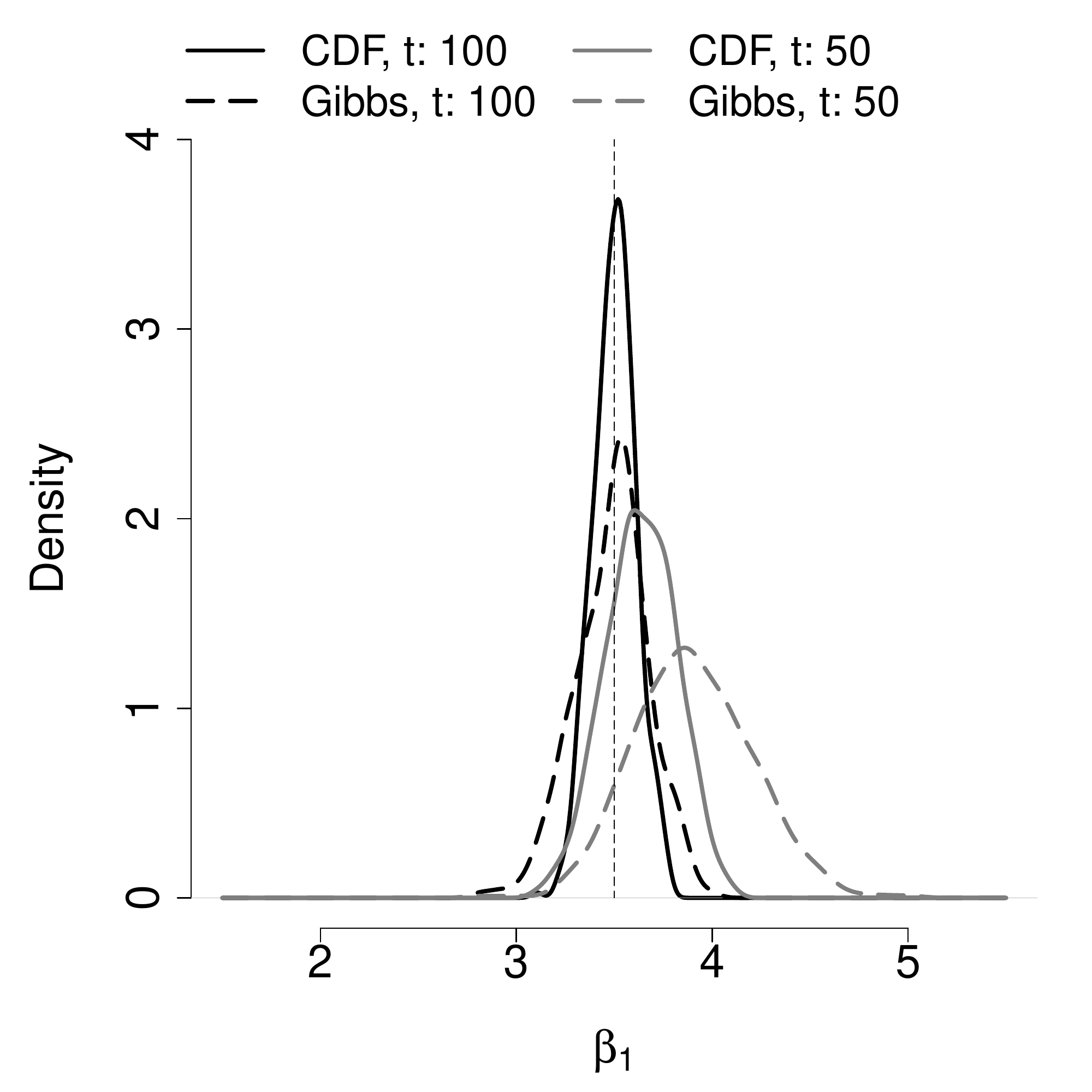}
\includegraphics[width=0.32\columnwidth]{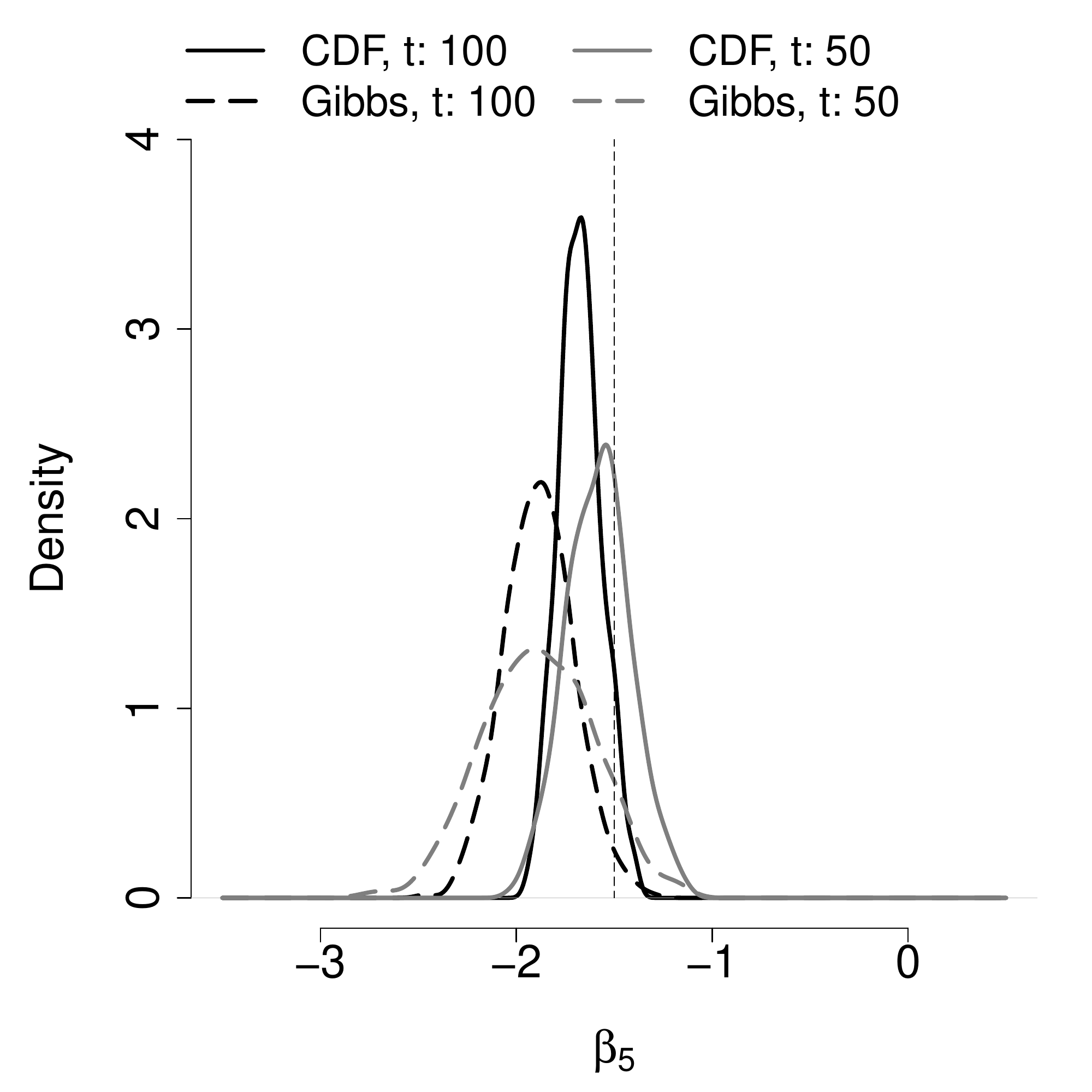}
\includegraphics[width=0.32\columnwidth]{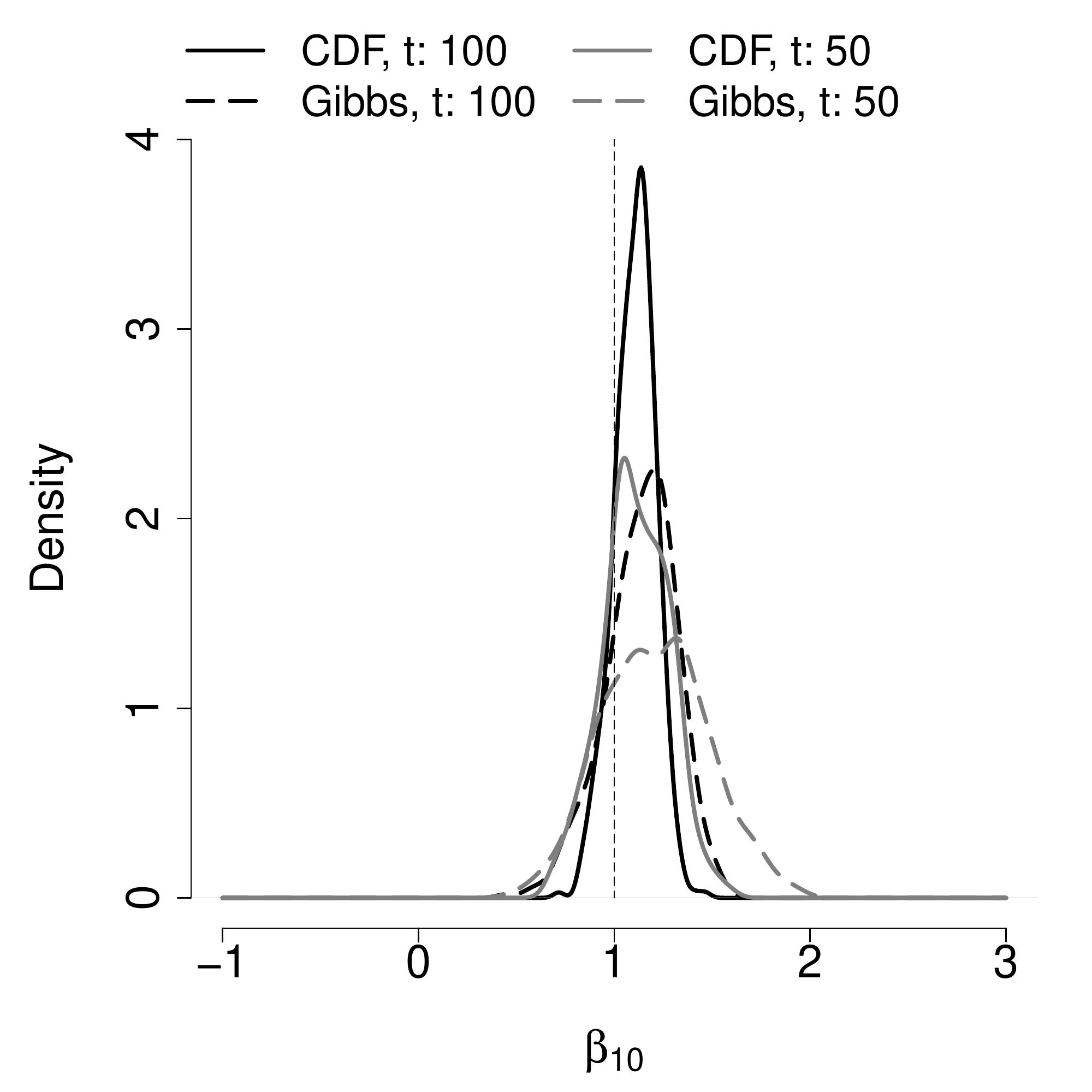} \\
\includegraphics[width=0.32\columnwidth]{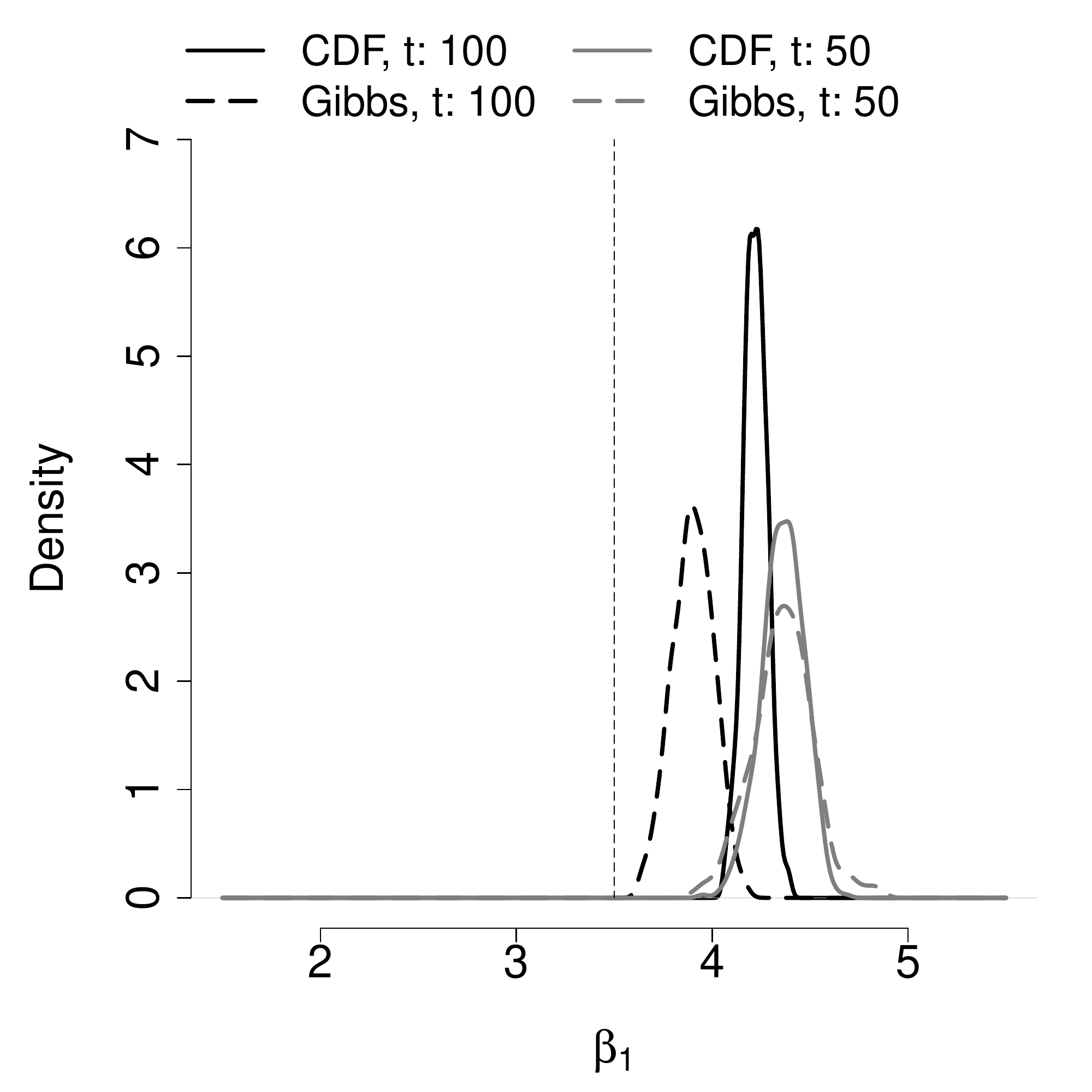}
\includegraphics[width=0.32\columnwidth]{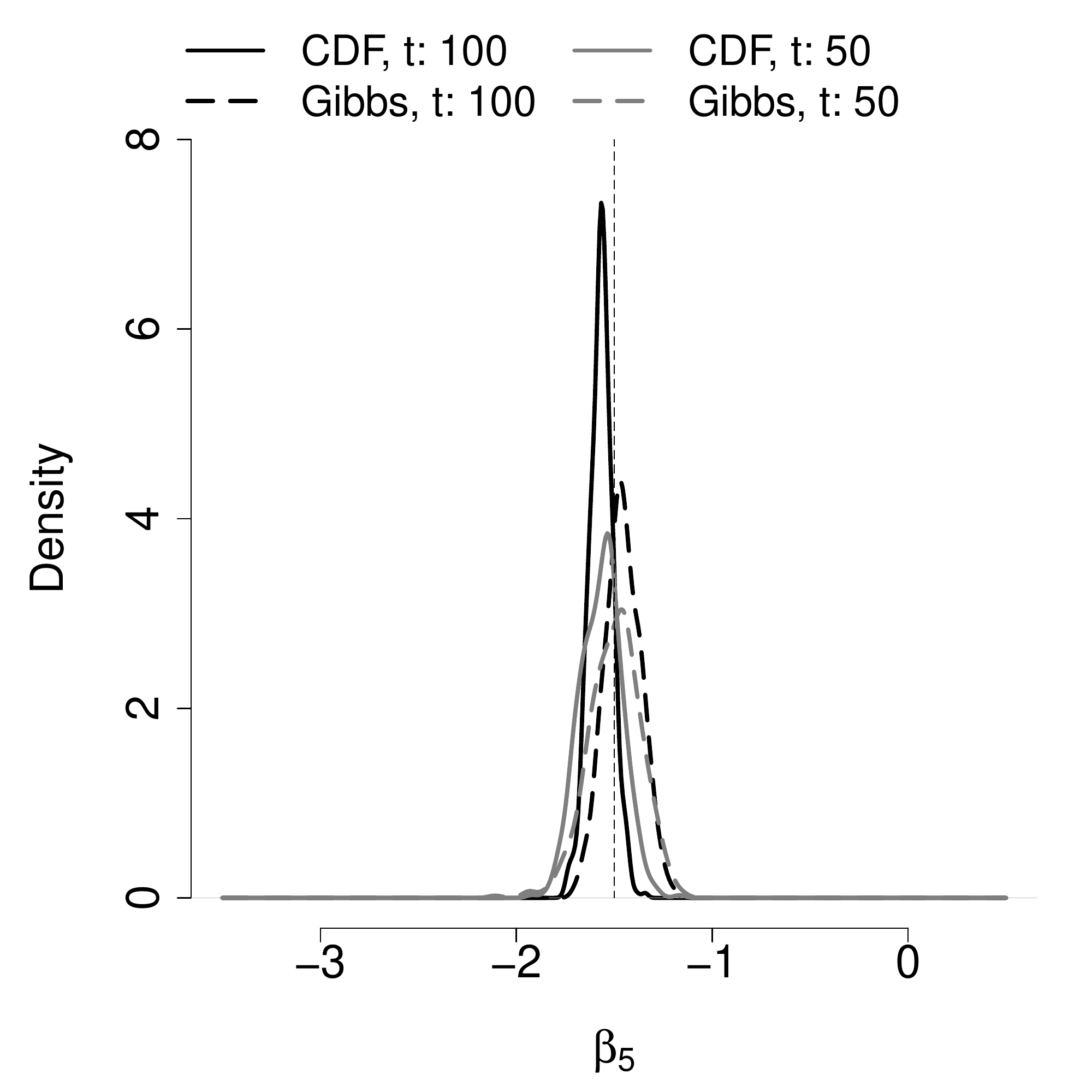}
\includegraphics[width=0.32\columnwidth]{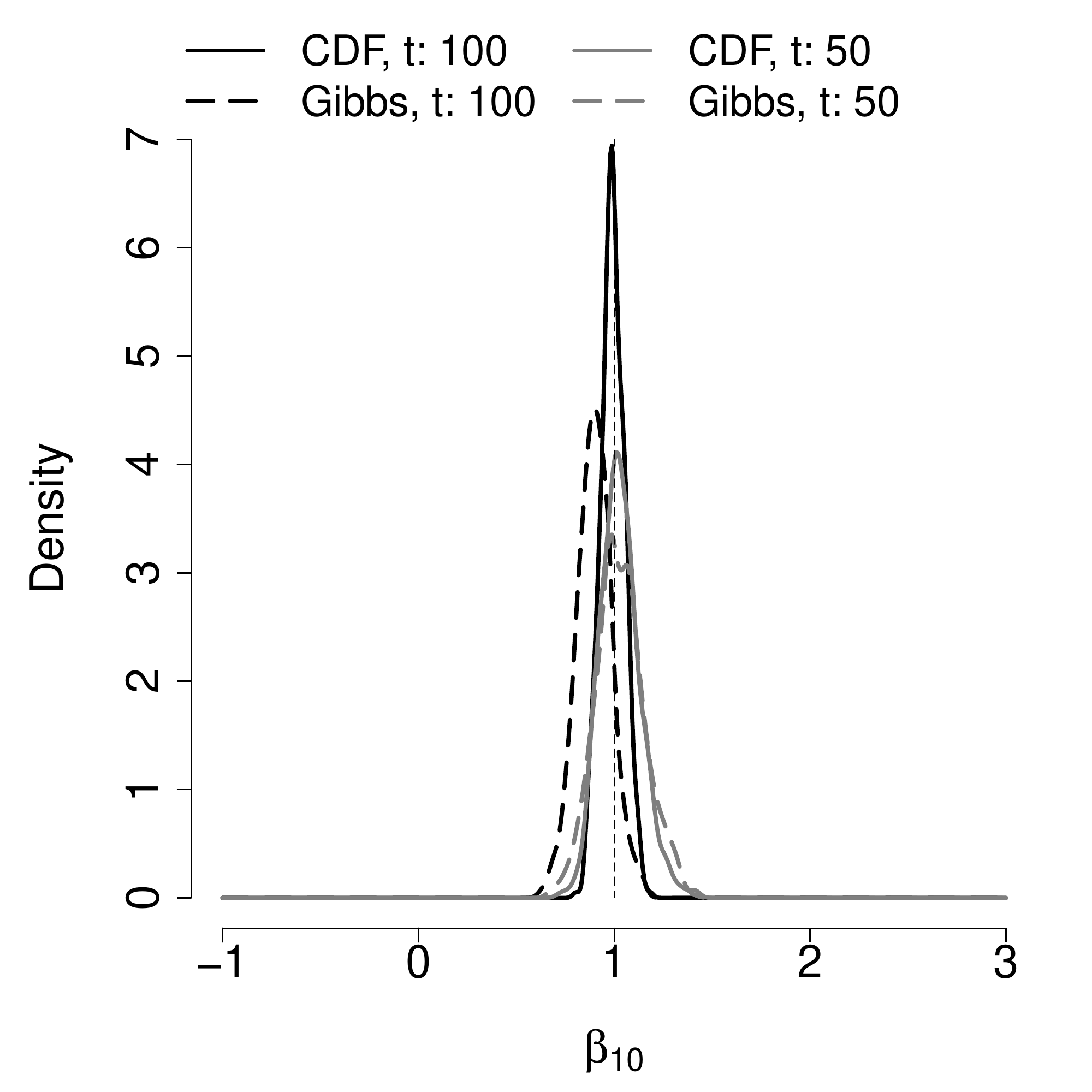}
\caption{Kernel density estimates for posterior draws using  the C-DF algorithm and S-MCMC at $t = 50, 100$.  Shown from left to right are plots of model parameters $\beta_1$, $\beta_5$, and $\beta_{10}$ (and top to bottom are case 1 and case 2), respectively.}
\label{fig:probit_densities}
\end{figure}

We report simulation results for the following examples: \begin{enumerate}[({Case} 1)]
\item $(p, b, n, t) = (100, 500, 25, 100)$: $\beta_{j,0} \sim \mathrm{U}(-3/4,3/4)$ for $ j \in [11, 100]$;
\item $(p, b, n, t) = (500, 3500, 100, 100)$: $\beta_{j,0} \sim \mathrm{U}(-1/3,1/3)$ for $j \in [11, 200]$.
\end{enumerate}
The first 10 regression coefficients are (3.5, -3.5, -2.0,  2.0, -1.5, 1.5, -1.5, 1.5, -1.0,  1.0), with $\beta_{j,0} = 0$ for $j > 200$ in case 2.  Data are generated as $x_{ij} \sim \mathrm{N}(0, \sigma = 0.25)$, and $\mathrm{P}(y_i  = 1) = \Phi(\bx_i' \bbeta)$.  Table \ref{tab:probit_performance_stats} summarizes inferential performance for the regression parameters in each of the simulated cases.  In case 2, although coverage for predictors with large coefficients (i.e., $\beta_1$ and $\beta_2$) is less than the nominal value, the average coverage across all predictors produced by C-DF is 70\% despite the high dimension with a significant number of ``noise'' predictors.  In addition, C-DF has very good mean-square estimation of parameter coefficient in both cases.  As a competitor to C-DF, S-MCMC (batch Gibbs), SMC \citep{chopin2002sequential} and S-VB \citep{broderick2013streaming} are implemented. Results for SMC with $2000$ propagated particles are shown in Table \ref{tab:probit_performance_stats}. For clarity, Figure \ref{fig:probit_densities} displays a number of marginal kernel density estimates at $t= 50, 100$ for C-DF and S-MCMC.
\begin{table}[h]
\small\centering
\caption{Inferential performance for C-DF, S-MCMC, SMC-CH, and S-VB in simulation studies $(p, b, n, t) = (100, 500, 25, 100)$ and $(p, b, n, t) = (500, 3500, 100, 100)$.  Coverage and length are based on 95\% credible intervals averaged over all predictors and over 10 replications.  MSE is reported over all predictors, while $\mathrm{MSE}_{10} = \tfrac{1}{10}\sum_{j=1}^{10} (\hat{\beta}_j - \beta_{j0})^2$.  We report the total time taken to produce 500 MCMC samples after the arrival of each data shard.}
\begin{tabular}{ l  l | c | c | c | c | c }
& & Avg. coverage $\bbeta$ & Length & Time (sec) & MSE$_{10}$ & MSE\\ \hline
\multirow{2}{*}{Case 1 ($p = 100$)}
& C-DF  & 0.77$_{0.08}$ & 0.42 & 28.9$_{2.5}$ & 0.025$_{0.01}$ & 0.04$_{0.01}$ \\
& S-MCMC & 0.96$_{0.01}$ & 0.65 & 152.5$_{10.3}$ &  0.018$_{0.01}$ & 0.02$_{0.01}$ \\
& SMC-CH & 0.28$_{0.05}$ & 0.34 & 19.9$_{1.8}$  & 1.020$_{0.38}$ & 0.40$_{0.13}$\\
& S-VB & 0.68$_{0.05}$ & 0.39 & 11.0$_{0.3}$ & 0.342$_{0.06}$ & 0.06$_{0.01}$\\
\hline
\multirow{2}{*}{Case 2 ($p = 500$)}
& C-DF  & 0.70$_{0.03}$ & 0.23 & 461.5$_{29.6}$ & 0.20$_{0.05}$ & 0.016$_{0.003}$ \\
& S-MCMC & 0.92$_{0.02}$ & 0.33 & 2,196.3$_{170.5}$ & 0.050$_{0.01}$ & 0.010$_{0.001}$\\
& SMC-CH & 0.10$_{0.02}$ & 0.12 & 650.8$_{9_6}$ & 1.750$_{0.23}$ & 0.730$_{0.08}$\\
& S-VB & 0.55$_{0.02}$ & 0.20 & 116.0$_{6.7}$ &  0.026$_{0.01}$ & 0.017$_{0.00}$\\
\end{tabular}
\label{tab:probit_performance_stats}
\end{table}

S-MCMC has the worst-case storage requirement and scales linearly in the number of training samples.  At each Gibbs iteration, draws for the latent scores is $O(nt)$ for S-MCMC compared to $O(b)$ for C-DF.  For both methods, sampling from the full conditional for $\bbeta$ is $O(p^3)$, updating sufficient statistics is $O(np^2)$, and updating surrogate quantities is $O(np)$.
Computational and storage requirements for both methods are summarized in Table \ref{tab:comp_complex}.
\begin{table}[h]
\small\centering
\caption{Computational and storage requirements for the latent variable probit model using C-DF and S-MCMC.  Budget $b$ represents the number of latent scores updated by C-DF when processing data shard at time $t$.  Runtime is quickly dominated by the sampling complexity which scales linearly in time for the augmented Gibbs sampler (S-MCMC).  Memory is reported for case 2, $(p, b, n, t) = (500, 3500, 100, 100)$. Sampling and update complexities are in terms of big-O.}
\begin{tabular}{ l | c | c | c | c | c }
& Stats & Data & Sampling & Updating & Memory (M-bytes) \\ \hline
C-DF & $\bS^{XX}, \bC$ & $\{\bx_i, y_i\}_{i > nt-b}$ & $S(p^2 + bp)$ flops & $p^3 + np^2$ flops & 20.0 \\
S-MCMC & $\bS^{XX}$ & $\{\bx_i, y_i\}_{i \ge 1}$ & $S(p^2 + ntp)$ flops & $p^3 + np^2$ flops & 46.0
\end{tabular}
\label{tab:comp_complex}
\end{table}

\section{C-DF for online compressed regression}\label{sec:bcr}
We consider an application to compressed linear regression where $\by_1,\dots, \by_t \in \Re^n$ are a sequence of $n$-dimensional response vectors with  associated predictors $\bX_1,\dots,\bX_t \in \Re^{n \times p}$ observed over time.  Data are modeled according to
\begin{equation}\label{eq:Bayessquash}
\by_t | \bPhi, \bbeta, \sigma^2 \sim \mathrm{N}(\bX_t\bPhi'\bbeta, \sigma^2\bI_n)
\end{equation}
for an $m \times p$ projection matrix $\bPhi$ with $m \ll p$.  A Bayesian analysis proceeds by sampling from posterior $[\bbeta, \bPhi, \sigma^2 | \bD^{(t)}]$ using the following default prior specification: $\bbeta \sim \mathrm{N}(\bzero,\sigma^2\bSigma_{\bbeta})$; $\sigma^2 \sim \mathrm{IG}(a,b)$; $\bPhi \sim \mathrm{MN}(\bPhi_0, \bK, \bones_m)$ centered on a row ortho-normalized random projection matrix, $\bPhi_0$, with row-specific scaling $\kappa_i \sim \mathrm{IG}(1/2,1/2)$, $1 \le i \le m$, and $\bK = \diag (\kappa_1, \dots, \kappa_m)$.  Note that $\bPhi'\bbeta=\bPhi'\bL\bL'\bbeta$ for any orthogonal matrix $\bL$, hence $\bPhi$ and $\bbeta$ are only identified up to an orthogonal transformation. Nevertheless, regression coefficients $\bgamma = \bPhi'\bbeta$ are identifiable, and valid inference is obtained using posterior draws of the associated model parameters.

\subsection{Online inference and competitors}  \label{sec:online_sup_learn}
Conditional on $\bPhi$, the posterior distribution over $\bbeta,\sigma^2$ factorizes as
\begin{equation} \label{eq:marg_betsig}
 \begin{alignedat}{2}
&\bbeta\given \bPhi, \bD^{(t)} \sim T_{n}(\bmu_t,\bSigma_t) &\quad&
\sigma^2\given \bPhi,\bD^{(t)} \sim \mathrm{IG}(a_{1,t}/2,b_{1,t}/2) \\
&\bSigma_t =b_{1,t}/n \, \bW^{-1} &\quad& a_{1,t} =nt  \\
&\bmu_t =(b_{1,t}/n)^{-1} \bSigma_t \,\bPhi\bF_t^{Xy'} &\qquad&b_{1,t}=F_t^{yy}-\bF_t^{Xy}\bPhi' \bW^{-1} \bPhi\bF_t^{Xy'},
\end{alignedat}
 \end{equation}
where  $T_{\nu}(\cdot)$ is the multivariate t-distribution with $\nu$ degrees of freedom and hyper-parameters defined in terms of sufficient statistics $F_{t}^{yy} = F_{t-1}^{yy} + \by_t'\by_t$, $\bF_{t}^{Xy} = \bF_{t-1}^{Xy} + \by_t'\bX_t$, $\bF_{t}^{XX}= \bF_{t-1}^{XX} + \bX_t'\bX_t$ and $\bW = \bPhi\bF_t^{XX}\bPhi'+\bSigma_{\bbeta}^{-1}$.  Sampling for $\bPhi= \left[ \bPhi_1, \bPhi_2, \ldots, \bPhi_p\right]$ proceeds by drawing successively from the set of column-specific full conditionals, namely
\begin{equation}
\begin{aligned}\label{eq:phi_update}
& \bPhi_j | \{\bPhi\}_{-j}, \bK, \bbeta, \bD^{(t)}  \sim
  \mathrm{N}_m(\bmu_{\bPhi_j} \bSigma_{\bPhi_j}), \quad &\kappa_i &| \bPhi \sim
  \mathrm{IG}(c_i / 2, d_i / 2) \\
& \bSigma_{\bPhi_j} = \Big( {\textstyle \sum_{s =1}^t} \: \bbeta\bX_{js}' \bX_{js} \bbeta' / \sigma^2 + \bK^{-1} \Big)^{-1} \quad & c_i &= c + p \\
& \bmu_{\bPhi_j} = \bSigma_{\bPhi_j} \Big( {\textstyle \sum_{s =1}^t} \: \bbeta\bX_{js}' \bz_{js} / \sigma^2_s + \bK^{-1} \bPhi_{0j} \Big) \quad & d_i &= d + (\bPhi^{(i,\cdot)} - \bPhi^{(i,\cdot)}_0)'(\bPhi^{(i,\cdot)} - \bPhi^{(i,\cdot)}_0).
\end{aligned}
\end{equation}
For column $j$, $\bz_{jt} = \by_t - \sum_{l \ne j} \bX_{lt} \bPhi_l' \bbeta$, with $\bX_{lt}$ denoting the $l$-th column of $\bX_t$, and $\bPhi^{(i,\cdot)}$ represents the $i$-th row of $\bPhi$.  The S-MCMC sampling scheme for model \eqref{eq:Bayessquash} propagates sufficient statistics $F_t^{yy}, \bF_t^{Xy} \textrm{ and } \bF_t^{XX}$ after observing $\{\bX_t, \by_t\}$ at time $t$ and draws sequentially from full conditional distributions \eqref{eq:marg_betsig} and \eqref{eq:phi_update}.  Due to high autocorrelation between $\bbeta$ and $\bPhi$, however, the online Gibbs sampler faces poor mixing in the joint parameter space.

The C-DF algorithm applied to this setting partitions model parameters into $\bTheta_{\mathcal{G}_1} = \{\bbeta, \sigma^2\}$ and $\bTheta_{\mathcal{G}_2} = \{\bPhi, \bK\}$, and sampling proceeds as:
\begin{enumerate}[(1)]
\item Observe data $\bD_t=\{\bX_t,\by_t\}$ at time $t$.  If $t=1$, set $\bbeta = \bzero$, $\bPhi = \bPhi_0$, $\sigma = \bar{s}(\text{vec}(\by_1, \dots, \by_k))$, and here we assume $\by_t$ is zero-mean.  Otherwise, set $\widehat{\bPhi}_t \leftarrow \widehat{\bPhi}_{t-1}$, $\hat{\kappa}_{i,t} \leftarrow \hat{\kappa}_{i, t-1}$;
\item Update surrogate quantities $\bC^{(t)}_{1,1} \leftarrow \bC^{(t-1)}_{1,1} + \widehat{\bPhi}_t \bX_t' \bX_t \widehat{\bPhi}_t'$ and $\bC^{(t)}_{1,2} \leftarrow \bC^{(t-1)}_{1,2} +\bPhi\bX_t'\by_t$.  Using the notation in \eqref{eq:marg_betsig}, redefine $\bW = \bC^{(t)}_{1,1} + \bSigma_{\bbeta}^{-1}$ and set $a_{1,t}=nt$, $b_{1,t}=F_t^{yy} - \bC^{(t)}_{1,2} \bW^{-1} \bC^{(t)'}_{1,2}$;  
\item Draw $S$ samples compositionally from $\sigma^2 | \widehat{\bPhi}_t, \bC_{1, \cdot}^{(t)}$ and $\bbeta | \sigma^2, \widehat{\bPhi}_t, \bC_{1, \cdot}^{(t)}$.  Here, a single $m \times m$ matrix inversion is required, instead of once at every sample using S-MCMC;
\item Set $\hat{\sigma}_t^2 \leftarrow b_{1,t}/(a_{1,t}+1)$ and $\widehat{\bbeta}_t \leftarrow \bmu_t$, where $\bSigma_t =b_{1,t}/n \, \bW^{-1}$, and $\bmu_t =(b_{1,t}/n)^{-1} \bSigma_t \, \bC_{1,2}^{(t)}$
(using closed-form expressions for the posterior MAP);
\item For $j = 1, \dots, p:$ update surrogate quantities $\bC_{21,j}^{(t)} \leftarrow \bC_{21,j}^{(t-1)} + \widehat{\bbeta}_t \widehat{\bbeta}_t' (\bX_{jt}' \bX_{jt})$ and $\bC_{22, j}^{(t)} \leftarrow \bC_{22,j}^{(t-1)} + \widehat{\bbeta}_t \bX_{jt}' \by_t$;
\item Draw $S$ samples from $\bPhi, \bK |
  \widehat{\bbeta}_t, \hat{\sigma}_t^2, \bC_{2\cdot}^{(t)}$ by modifying full conditionals \eqref{eq:phi_update}, i.e., $[\bK|\bPhi, -]$ and $[\bPhi_j|\{\bPhi\}_{-j}, \bK, -]$, $1 \le j \le p$, in terms of the defined surrogate quantities;
\begin{enumerate}[(a)]
\item compute $\bH_{jt}(\bPhi_{-j}) = \bC_{22,j}^{(t)} - \sum_{l \ne j} \bC_{21, l}^{(t)} \bPhi_l$. Draw from $[\bPhi_j | \{\bPhi\}_{-j}, \bK, - ]$ with $\bSigma_{\bPhi_j}= \big( \bC_{21,j}^{(t)} / \hat{\sigma}_t^2 + \bK^{-1}\big)^{-1}$ and  $\bmu_{\bPhi_j}=\bSigma_{\bPhi_j}\big(\bH_{jt} / \hat{\sigma}_t^2 +  \bK^{-1} \bPhi_{0j}\big)$;
\item draw $[\bK | \bPhi, -]$ by sampling independently from $[\kappa_i | \bPhi, -]$, $i = 1, \dots, m$.
\end{enumerate}
\item Set $\widehat{\bPhi}_t$ and $\{\hat{\kappa}_{i,t}\}$ as the sample mean over these $S$ draws.
\end{enumerate}
By propagating surrogate quantities instead of the much larger sufficient statistics, the C-DF algorithm significantly reduces storage requirements, provides state-of-the-art inference and improves mixing efficiency as compared to S-MCMC.  The latter is measured in terms of an effective sample size, i.e., the number of MCMC iterations in order to achieve a desired predictive accuracy (see Table \ref{tab:ess} in Section \ref{sec:bcr_sim_results}).

Bayesian shrinkage methods such as Bayesian Lasso \citep{park2008bayesian} and GDP \citep{armagan2013generalized} were attempted but rendered infeasible by the need to invert a $p \times p$ matrix at each MCMC iteration of every time-point.
SMC \citep{chopin2002sequential} suffers from severe particle degeneracy for learning the high-dimensional parameter $\bPhi$ and requires a large number of particle with massive computation time (as in the binary regression example), and while sufficient statistics for particle rejuvenation are available, inference fairs no better than S-MCMC.  In addition, the need to store and propagate a large number of high-dimensional particles is completely impractical.  Instead, we derive a variational Bayes (VB) approximation to the joint posterior using a GDP\footnote{The GDP prior has been shown to induce attractive shrinkage properties, with a carefully constructed hierarchical prior that allow  Cauchy-like tails leading to better robustness (less bias due to over shrinkage) in estimating true signals, while having Laplace-like support near zero, leading to better concentration around sparse coefficient vectors.} shrinkage prior on the coefficients of a standard linear regression model, $y_i = \mathrm{N}(\bx_i' \bbeta, \sigma^2)$ and $\beta_j\given\sigma\sim \mathrm{GDP}(\zeta = \sigma\eta / \alpha,\alpha)$.  The latter is equivalent to hierarchical prior $\beta_j\given\sigma,\tau_j \sim \mathrm{N}(0,\sigma^2\tau_j),$ with $\tau_j\sim \mathrm{Exp}(\lambda_j^2/2)$ and $\lambda_j\sim \mathrm{Ga}(\alpha,\eta)$. 
For $\bTheta=(\bbeta,\btau,\blambda,\sigma^2)'$, $\btau=(\tau_1,\dots,\tau_p)'$, and $\blambda=(\lambda_1,\dots,\lambda_p)'$, we approximate $\pi(\bTheta|\bD)$ by a variational posterior with product form  $q(\bTheta) = \prod_j q_j(\btheta_j)$.
Optimal densities $q_j(\btheta_j)\,\propto\, \exp\big[E_{-q(\btheta_j)}\left\{\log \pi(\bTheta,\bD)\right\}\big]$ are obtained by minimizing the Kullback-Leibler distance between $\pi(\btheta|\bD)$ and $q(\bTheta)$, where $E_{-q(\btheta_j)}$ denotes the expectation over $\prod_{i\neq j}q_i(\btheta_i)$. The latter are well known results established in the variational Bayes literature \citep{braun2010variational,luts2013variational}. 
We implement the VB competitor in batch-mode that makes use of data seen until the present time-point. In terms of predictive inference, the latter is arguably at least as good as S-VB \citep{broderick2013streaming} which restricts usage of the full data. 


\subsection{Simulation experiments}
\label{sec:simulation}  \label{sec:bcr_sim_results}
The C-DF algorithm provides robust parameter inference and predictive performance across numerous simulation experiments which consider varying degrees of sparsity as well as predictor dimension and correlation (see Tables \ref{tab:mspe} and \ref{tab:coverage}).
Shards of $n = 100$ observations arrive sequentially over a $T = 500$ time horizon, and predictor data are generated as $\bx_i \sim \mathrm{N}(\bzero,\bR), ~\bR_{jk} = \rho^{| j - k |}$, for $j,k = 1,2,\ldots, p$ and correlation $\rho \in (0,1)$.  The response is generated as $y_i \sim \mathrm{N}(\bx_i' \bbeta, \sigma^2 = 4)$, where true coefficient vector $\bbeta$ used in each case is specified in Table \ref{tab:sim_runs}.
\begin{table}[h]
\centering
\caption{Simulation experiments for supervised compressed regression. ``High'' signal corresponds to $\beta_j \sim \mathrm{U}(-3,3)$, while ``low'' signal corresponds to $\beta_j = 0.10$ for every nonzero feature.  Sparse cases are denoted with (*).}
\begin{tabular}{c|cccc}
Case & $\rho$ & $p$ & \#$\beta_j \ne 0$ & Signal \\
	\hline
1* & 0.1 & 500 & 10 & high\\
2* & 0.1 & 1000 & 10 & high\\
3* & 0.4 & 500 & 10 & high\\
4* & 0.4 & 1000 & 10 & high\\
5 & 0.1 & 500 & 500 & high\\
6 & 0.1 & 500 & 500 & low
\end{tabular}
\label{tab:sim_runs}
\end{table}

Table \ref{tab:mspe} reports predictive MSE for each simulation experiment averaged over 50 simulated datasets.
In all cases, the results appear robust to the choice of $m$,  the dimension of the subspace which $\bPhi$ maps into.  Choosing  $m$ large can add significantly to the computational overhead of the algorithm, so we suggest setting $m = \max\{10, \log(p)\}$. VB-GDP yields the lowest MSPE for sparse truths, though C-DF and S-MCMC provide competitive performance.  Increasing the number of (correlation between) predictors causes MSPE to increase for all methods.  C-DF  performs well in all cases, while VB-GDP suffers for dense truths, especially in the low-signal setting.  In addition, C-DF results in excellent parameter estimation and variable selection, including cases with high predictor  correlation (see Table \ref{tab2}).  Table \ref{tab:coverage} reports coverage probabilities for 95\% predictive intervals for the competing methods. While C-DF and S-MCMC show proper coverage, VB suffers due to the restrictive assumption of independence between parameters a-posteriori.
\begin{table}[h]
\centering
\caption{MSPE comparisons for each simulated experiment in Table \ref{tab:sim_runs}.  Subscripts denote bootstrapped standard errors calculated using independent replications.}
\begin{tabular}{l | c | c | c | c | c | c}
& Case 1* & Case 2* & Case 3* & Case 4* & Case 5 & Case 6 \\
\hline
VB & $3.43_{0.005}$  & $4.20_{0.006}$  &  $3.49_{0.005}$ & $4.23_{0.006}$  &  $3.52_{0.007}$  & $8.79_{0.010}$ \\
C-DF & $3.49_{0.010}$ & $4.38_{0.010}$ &  $3.62_{0.007}$ & $4.40_{0.020}$ & $3.81_{0.006}$ & $3.64_{0.020}$\\
S-MCMC & $3.56_{0.020}$ & $4.40_{0.020}$ & $3.58_{0.020}$ & $4.43_{0.020}$ & $3.68_{0.020}$ & $3.70_{0.020}$\\
\end{tabular}
\label{tab:mspe}
\end{table}

\begin{table}[h]
\small\centering
\caption{Performance comparison in terms of relative parameter MSE at $t=100,200$.  Relative MSE for $\bgamma = \bPhi'\bbeta$ is computed as $ ||\widehat{\bPhi}_t'\widehat{\bbeta}_t - \bgamma_0||^2 / ||\bgamma_0||^2$ for C-DF and S-MCMC, $\bgamma_0$ is $\bgamma$ at the truth. For VB, we compute $||\bmu_\bbeta^{\rm VB} - \bgamma_0||^2 / ||\bgamma_0||^2$, where $\bmu_\bbeta^{\rm VB}$ is the approximate posterior mean for $\bbeta$. Subscripts denote bootstrapped standard errors calculated using independent replications.}
\begin{tabular}{l | c | c | c | c | c | c | c }
& Time & Case 1* & Case 2* & Case 3* & Case 4* & Case 5 & Case 6 \\
\hline
\multirow{2}{*}{VB}
& $100$ &  $0.009_{0.001}$ & $0.019_{0.001}$ & $0.018_{0.002}$ & $0.024_{0.002}$ & $0.002_{0.001}$ & $0.70_{0.06}$\\
& $200$ & $0.004_{0.001}$ & $0.008_{0.001}$ & $0.005_{0.001}$ &
$0.008_{0.001}$ & $0.001_{0.001}$ & $0.73_{0.06}$ \\
\hline
\multirow{2}{*}{C-DF}
& $100$ &  $0.010_{0.001}$ & $0.033_{0.003}$ & $0.029_{0.004}$ & $0.074_{0.001}$ & $0.013_{0.001}$ & $0.042_{0.003}$ \\ 
& $200$ & $0.004_{0.001}$ & $0.011_{0.002}$ & $0.011_{0.002}$ & $0.027_{0.001}$ & $0.003_{0.001}$ & $0.020_{0.00}$ \\
\hline
\multirow{2}{*}{S-MCMC}
& $100$ &  $0.014_{0.003}$ & $0.032_{0.004}$ & $0.024_{0.003}$ & $0.034_{0.009}$ & $0.004_{0.001}$ & $0.063_{0.006}$\\ 
& $200$ & $0.006_{0.001}$ & $0.015_{0.002}$ & $0.010_{0.002}$ & $0.017_{0.004}$ & $0.002_{0.000}$ & $0.036_{0.002}$
\end{tabular}
\label{tab2}
\end{table}

\begin{table}[h]
\centering
\caption{Predictive coverage for each simulation experiment in Table \ref{tab:sim_runs}. Empirical 95\% confidence intervals of the coverage probabilities over independent replications are also reported.}
\begin{tabular}{l | c | c | c | c | c | c }
& Case 1* & Case 2* & Case 3* & Case 4* & Case 5 & Case 6 \\
\hline
\multirow{2}{*}{VB}
& 0.82 & 0.83 & 0.82 & 0.83 & 0.81 & 0.83 \\
& (0.81,0.84)     &  (0.76,0.87)    & (0.81,0.85)     & (0.75,0.87) & (0.76,0.86) & (0.80,0.85) \\
\multirow{2}{*}{C-DF}
& $0.98 $ & $0.98 $ &  $0.98 $ & $0.98 $ & $0.97 $ & $0.99 $\\
& (0.98,0.99) & (0.97,0.99) & (0.98,0.99) & (0.97,0.98) & (0.97,1) & (0.99,1)\\
\multirow{2}{*}{S-MCMC}
& 0.96 & 0.93 & 0.95 & 0.93 & 0.96 & 0.95\\
& (0.95,0.96) & (0.91,0.96) & (0.95,0.96) & (0.90,0.96) & (0.96,0.96) & (0.93,0.97)
\end{tabular}
\label{tab:coverage}
\end{table}


Propagating surrogate quantities using the C-DF algorithm for model \eqref{eq:Bayessquash} results in a dramatic efficiency gain over S-MCMC.  A measure of this efficiency is the ``effective sample-size,'' namely, the number of samples required for the predictive MSE to drop below a chosen threshold.  Table \ref{tab:ess} reports effective sample sizes for each simulated experiments in Table \ref{tab:sim_runs}.
\begin{table}[!h]
\small\centering
\begin{tabular}{ l | c | c | c | c | c | c | c | c }
& \multicolumn{6}{c |}{Effective sample size} & \multicolumn{2}{c}{Memory (M-bytes)}\\
\hline
& Case 1* & Case 2* & Case 3* & Case 4* & Case 5 & Case 6 & $p=500$ & $p=1000$\\
\cline{2-9}
C-DF & $2040 $ & $2610 $ &  $1710 $ & $3630 $ & $5250 $ & $210 $ & 0.68 & 1.34\\
S-MCMC & 2940 & 3990 & 2820 & 4410 & 4440 & 1050 & 2.00 & 8.10
\end{tabular}
\caption{Effective sample sizes (number of MCMC samples) required until $\mathrm{MSPE} \le 5$  are shown above for each simulation experiment, with results averaged over 10 independent replications.  The storage required in terms of propagated quantities is also reported for each competitor.}
\label{tab:ess}
\end{table}

Storage of the sufficient statistics for model \eqref{eq:Bayessquash} and updating these quantities is $O(np^2)+O(np)$ at each time point.  In comparison, updating surrogate quantities for C-DF is  $O(mpn)+O(m^2n)$. When $p \gg m$, then C-DF algorithm offers a dramatic reduction in terms of the storage requirement for online inference, reducing the quadratic dependence on $p$ to linear-order.  Runtimes  using a non-optimized R implementation were comparable for S-MCMC and C-DF across the simulation studies.  


\section{Real data illustration}
We use C-DFs to analyze the UC Irvine Adult dataset and verify its performance relative to batch MCMC (S-MCMC). The latter is more computationally expensive than C-DF as shown in Table \ref{tab:comp_complex} in Section \ref{sec:probit}, but serves as a state-of-the-art performance benchmark. 
Each entry records 6 continuous and 8 categorical attributes of a census form for households which are used to predict whether UC households have an income greater than \$50,000. For our analysis, we use six continuous features and only one among the eight categorical features, namely the native country of the respondents. There are participants from 42 different native countries in the study which results in 41 dummy variables after binary coding. Deleting rows with missing entries, a total of $N=30,000$ observations are divided into $T=100$ shards of size $n=300$. Additional information for the data is available at \href{http://archive.ics.uci.edu/ml/datasets/Adult}{http://archive.ics.uci.edu/ml/datasets/Adult}.

\begin{figure}[h]
\centering
\includegraphics[width=0.32\columnwidth]{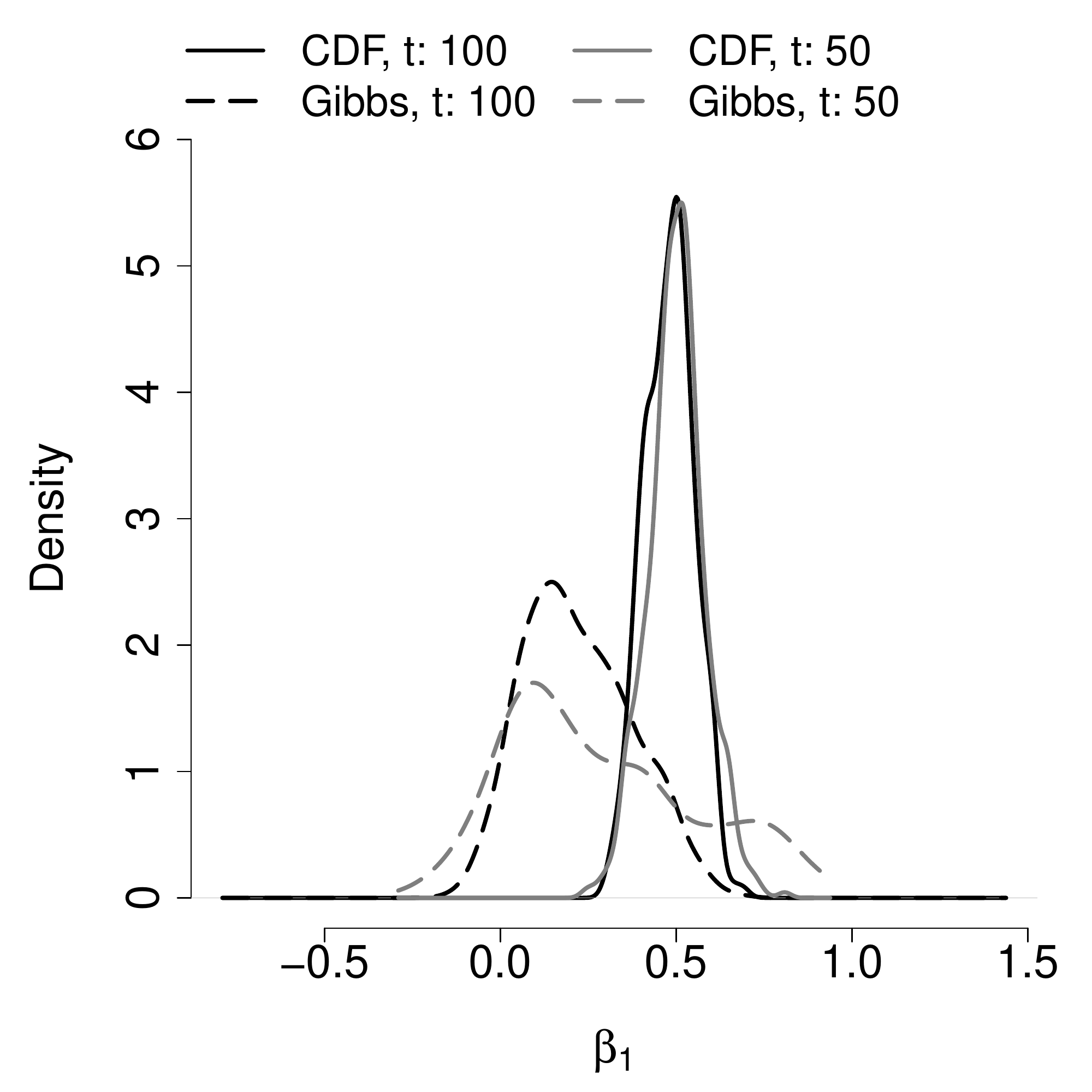}
\includegraphics[width=0.32\columnwidth]{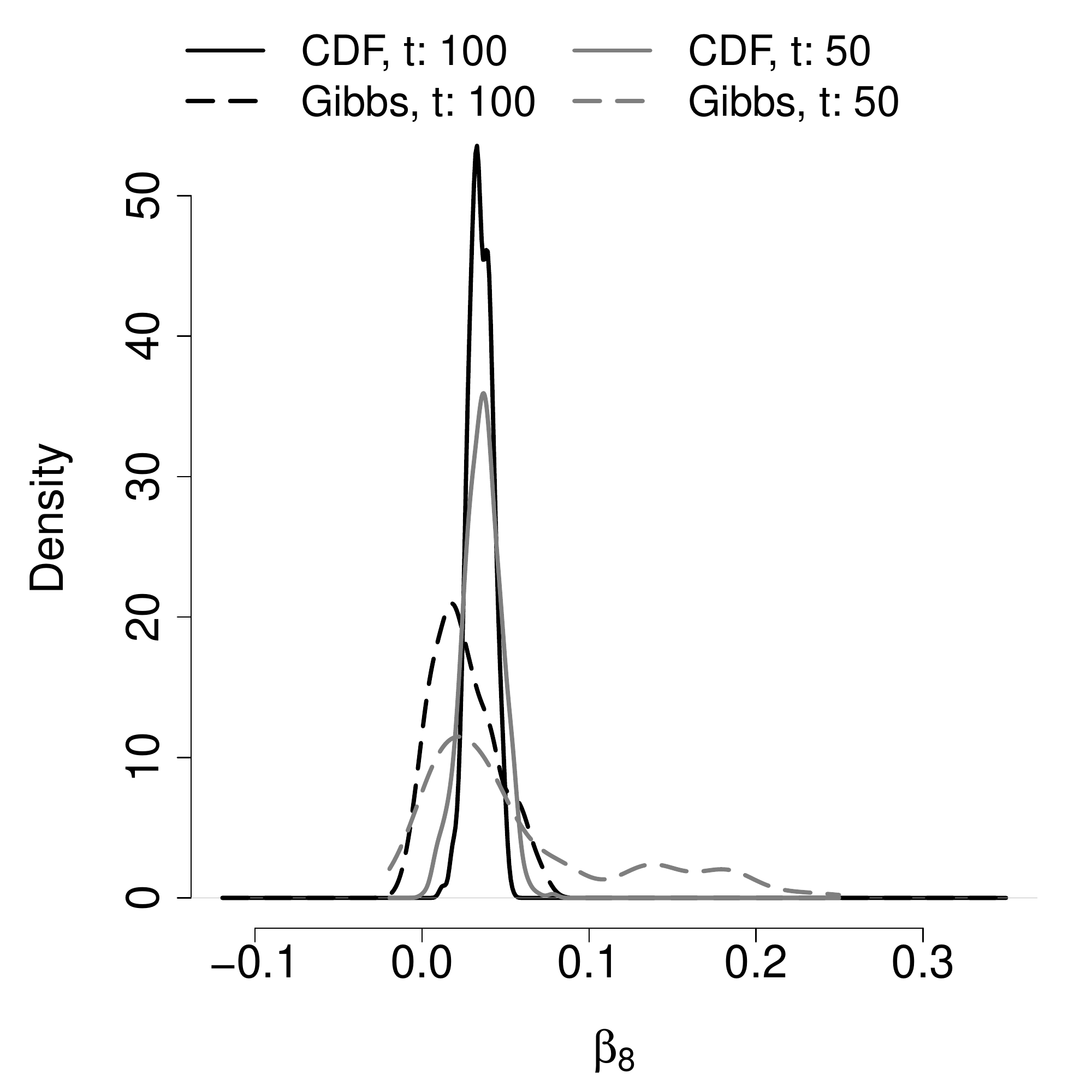}
\includegraphics[width=0.32\columnwidth]{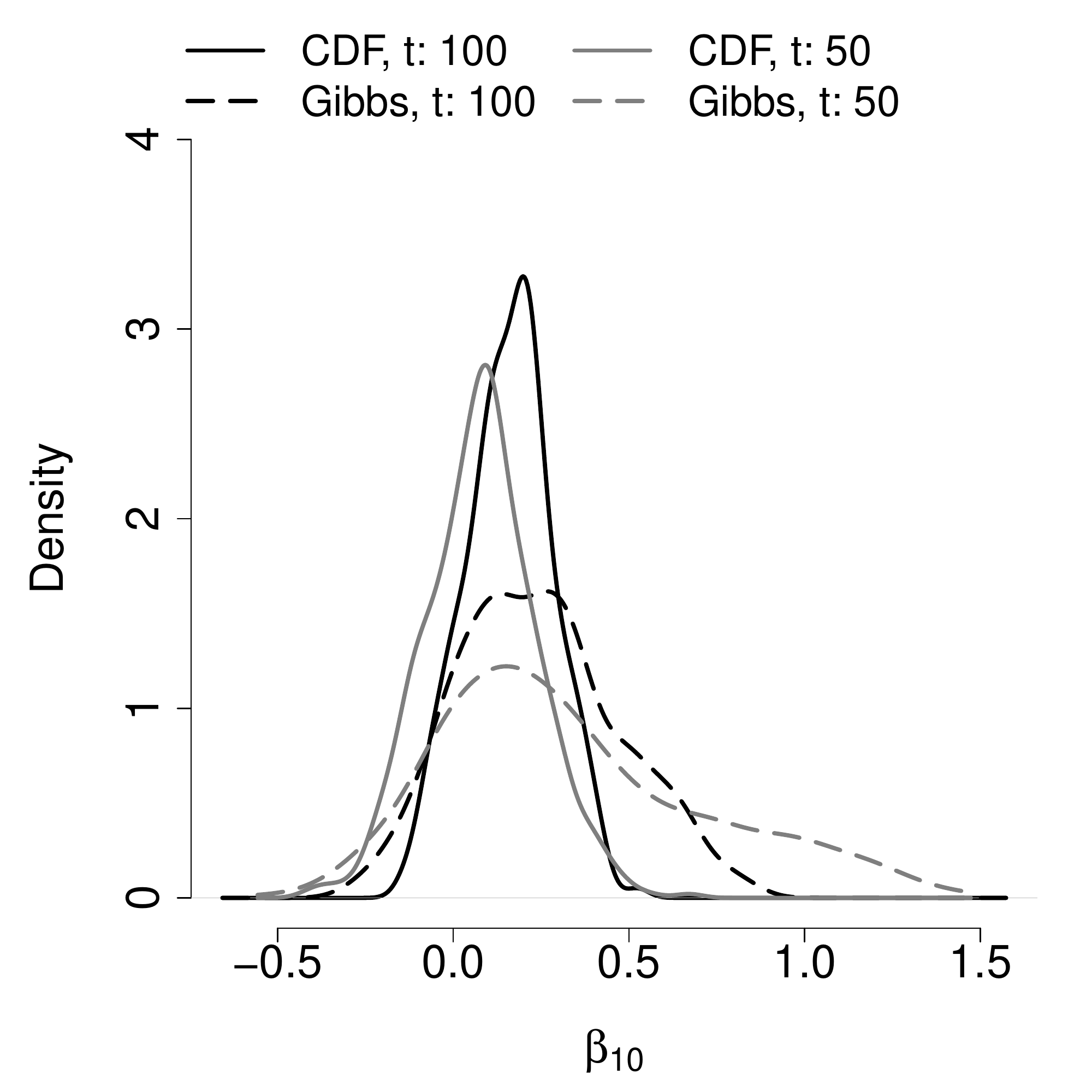}
\caption{(left to right): Kernel density estimates for posterior draws of $\beta_1, \: \beta_8, \: \beta_{10}$ using S-MCMC and the C-DF algorithm at $t = 50, 100$. Dotted and solid lines represent kernel density estimates for S-MCMC and C-DF, respectively.}
\label{fig:data_densities}
\end{figure}

\begin{figure}[!h]
\centering
\includegraphics[width=0.33\columnwidth]{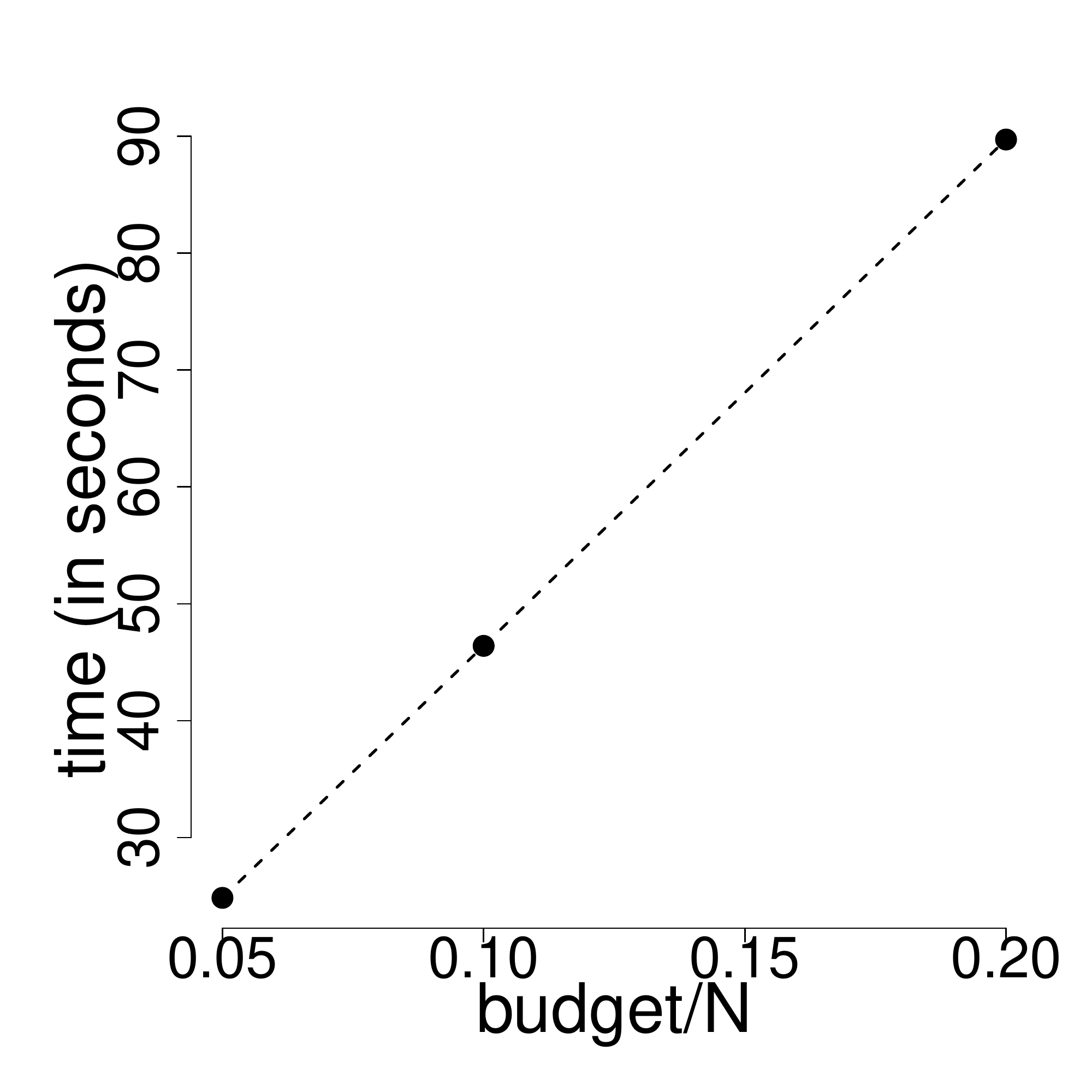} \qquad
\includegraphics[width=0.33\columnwidth]{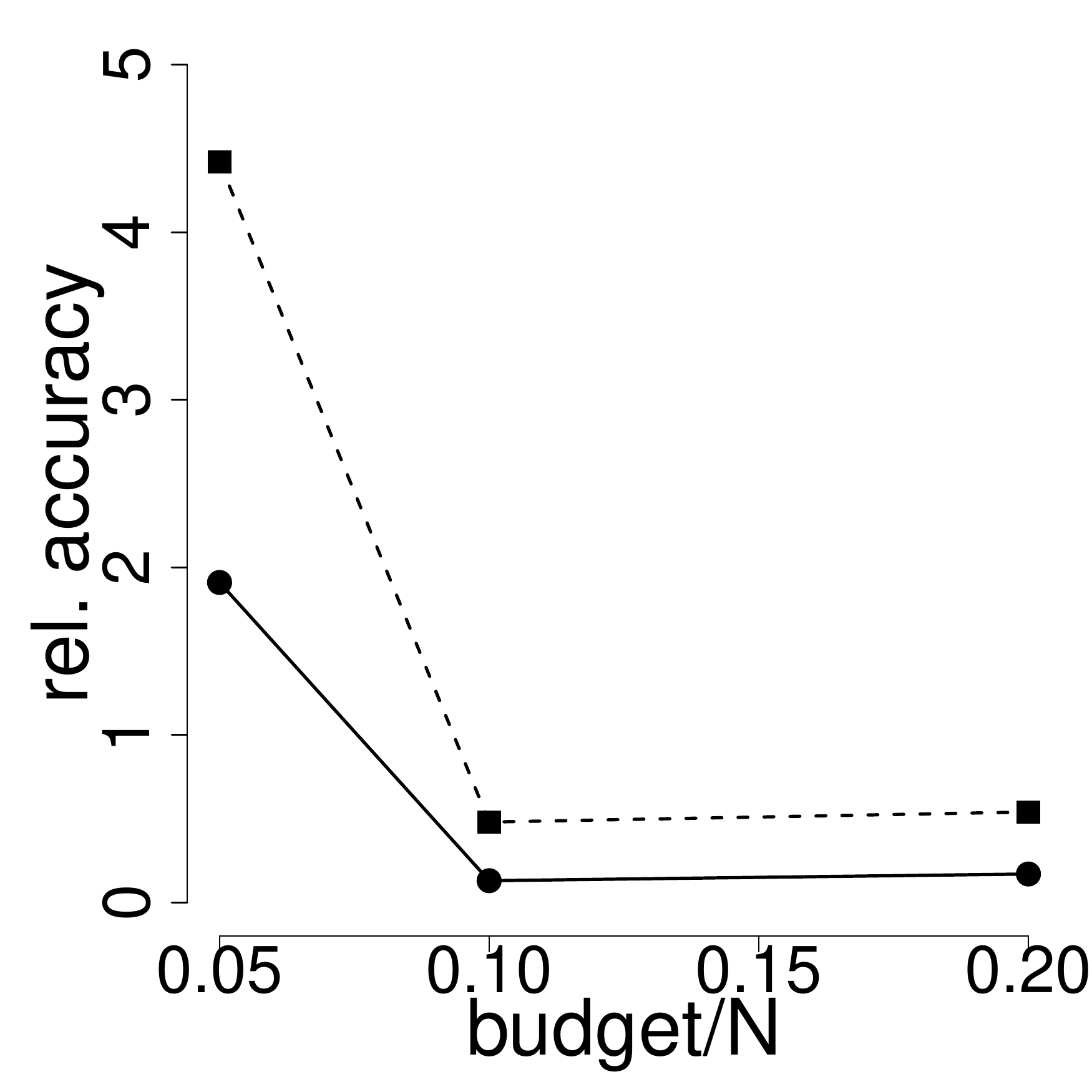}
\caption{Left: compute-time (in seconds) for C-DF with different budget sizes, $b$. Right: relative accuracy measured as $\sum_{i=1}^{p} |\widehat{\beta}_{i}^{\rm CDF}-\widehat{\beta}_i^{\rm Gibbs}| \,/ \sum_{i=1}^{p}|\widehat{\beta}_{i}^{\rm Gibbs}|$; here $\widehat{\beta}_i^{\rm CDF}$ and $\widehat{\beta}_i^{\rm Gibbs}$ are the posterior means of $\beta_i$ for C-DF and S-MCMC, respectively. Dotted and solid lines represent accuracy for $t=50$ and $t=100$, respectively. }
\label{fig:comp_accuracy}
\end{figure}

In the case of probit regression, to assess the sensitivity of C-DFs performance to the choice of budget size, we report results for three different values, $b=0.05N, 0.07N, \text{ and } 0.10N$. 
 Figure \ref{fig:data_densities} presents kernel density plots of several feature coefficients at $t=50, 100$ for C-DF and S-MCMC. As $t$ increases, it is clear that density estimates between the two methods become increasingly comparable. Figure \ref{fig:comp_accuracy} provides a measure of relative deviance between posterior means of the regression coefficients for C-DF and S-MCMC. Table \ref{Table:data} presents four $2\times 2$ tables, comparing the classification of S-MCMC with C-DF. 
Here, the $(i,j)$-th cell of each $2 \times 2$ table records $\#(y=i,\hat{y}=j) / N$, where $\hat{y}$ is the predicted response and $i, j \in \{1,2\}$. The sum of off-diagonal entries in these tables provides the overall misclassification rate.

Given the small predictor dimension for this real data, a relatively small budget size works well. In larger predictor or lower signal-to-noise ratio settings, however, the use of a larger budget size may be needed to obtain good performance\footnote{Across a range of simulation examples in  Section \ref{sec:probit}, C-DF demonstrates comparable performance to S-MCMC (batch MCMC) with a dramatic reduction in memory utilization and compute-time.}.  In fact, C-DF and S-MCMC have nearly identical classification performance, even when the budget is merely 10\% of the full data size.  It is also evident that the C-DF approximation is robust, with virtually no change in terms of estimation-error or the classification rate as $b$ is increased further; see Figure \ref{fig:comp_accuracy}. 
 Finally, compute-time for C-DF scales linearly in the budget size instead of the much large data sample size.  In particular, S-MCMC takes 1214 seconds to sample a fixed number of MCMC over the entire time horizon, while C-DF ($b = 0.10 N$) completes the same in 65 seconds. 


\begin{table}[ht]
\caption{Four tables present the proportion of predicted response and true response for C-DF with $b=0.05N, 0.10N, 0.20N$ and the batch MCMC sampler (S-MCMC).}
\begin{minipage}{.5\linewidth}
\centering
\begin{tabular}{ll | ccc}
 &  & $\hat{y}=0$ & $\hat{y}=1$ \\
\hline
\multirow{2}{*}{$b=0.05N$} & $y=0$ &  $0.69$ & $0.05$ \\ 
 & $y=1$ & $0.18$ & $0.08$ \\
 \hline
\multirow{2}{*}{$b=0.10N$} & $y=0$ &  $0.70$ & $0.05$ \\ 
 & $y=1$ & $0.16$ & $0.09$ \\
\hline
\end{tabular}
\end{minipage}%
\begin{minipage}{.5\linewidth}
\centering
\begin{tabular}{ll | ccc}
 &  & $\hat{y}=0$ & $\hat{y}=1$ \\
\hline
\multirow{2}{*}{$b=0.20N$} & $y=0$ &  $0.70$ & $0.05$ \\ 
 & $y=1$ & $0.17$ & $0.08$ \\
 \hline
\multirow{2}{*}{S-MCMC} & $y=0$ &  $0.70$ & $0.05$ \\ 
 & $y=1$ & $0.15$ & $0.10$ \\
\hline
\end{tabular}
    \end{minipage}
    \label{Table:data}
\end{table}


\section{Convergence Behavior of Approximate Samplers}\label{sec:theory_cdf}

We establish convergence behavior for a general class of approximate MCMC algorithms of which the C-DF algorithm is shown to be a special case.  We first characterize the limiting approximation error that results when a kernel having the desired target stationary distribution is approximated by another kernel in the finite sample setting. This result is general and is not limited to any specific class of approximations.  Next, we show that such kernel approximations improve under an increasing sample size to yield draws from the exact posterior asymptotically. Proofs are provided in the appendix.

\subsection{Notation and Framework}

Let $\pi_t(\cdot | \bD^{(t)})$ denote the posterior distribution having observed data $\bD^{(t)}$ through time $t$.  A sequence of probability measures $\pi_t(\cdot | \bD^{(t)})$ is defined on a corresponding sequence of measure spaces, $(\bH_t,\mathcal{H}_t), ~ t \ge 1$.   Below, we take $\bH_t=\Re^p$, with $\mathcal{H}_t=\mathcal{B}(\Re^p)$ denoting the Borel $\sigma$-algebra on $\Re^p$ for a fixed $p$-dimensional parameter space $\bTheta = (\theta_1, \cdots, \theta_p)$.  $\pi_t$ admits density $\pi_t(\bTheta)$ with respect to the Lebesgue measure $d\nu(\bTheta) = d\nu_1(\theta_1)\,d(\theta_2), \dots, d\nu_p(\theta_p)$. For a transition kernel $T_t:\Re^p\times\Re^p\rightarrow\Re^{+}$  at time $t$, we assume
\begin{enumerate}
\item $T_t(\bx,\cdot)$ is a probability measure for all $\bx\in\Re^p$
\item $T_t(\cdot,\bA)$ is a measurable function w.r.t the $\sigma$-algebra for all $\nu$-measurable sets $\bA$.
\end{enumerate}
A function $f_t:\Re^p\rightarrow\Re^{+}$ defined over $\mathcal{H}_t$ is the invariant distribution of $T_t$ if
\begin{align}\label{eq:invariant}
f_t(\bTheta') = \int T_t(\bTheta,\bTheta') f_t(\bTheta) \, d\nu(\bTheta).
\end{align}

We study the convergence of a sequence of distributions in total variation norm, namely $d_{TV}(\mu_1,\mu_2)=\sup_\bA \big|\mu_1(\bA)-\mu_2(\bA)\big|$.

\subsection{Finite sample error bound for approximate kernels}
In the finite sample setting, transition kernels and stationary distributions omit the subscript $t$.  We begin by characterizing the propagation of error induced by approximating one kernel with another (valid) kernel.
\begin{lemma}\label{lem:lin}
Let $K$ be a kernel approximated by a kernel $T$ s.t.
\begin{align*}
\sup_{\bTheta}||K(\bTheta,\cdot)-T(\bTheta,\cdot)||_{TV}\leq \rho
\end{align*}
for some constant $\rho > 0$.  Let $\mu_1$, $\mu_2$ denote the stationary distributions for $T$ and $K$, respectively, and assume $||T^{(r)}-\mu_1||_{TV}\rightarrow 0$ and $||K^{(r)}-\mu_2||_{TV}\rightarrow 0$. Then there exists an $r_0$ s.t.
\begin{align}\label{eq:linprop}
\begin{array}{ll}
\sup_{\bTheta}||K^{(r)}(\bTheta,\cdot)-T^{(r)}(\bTheta,\cdot)||_{TV} \leq r \rho, & \mbox{\rm if}~ r \le r_0 \\
\sup_{\bTheta}||K^{(r)}(\bTheta,\cdot)-T^{(r)}(\bTheta,\cdot)||_{TV}\leq \min\{\rho r, 2||\mu_1-\mu_2||_{TV}\}, & \mbox{\rm if}~r > r_0.
\end{array}
\end{align}
\end{lemma}
When the sample size is small and the data are observed sequentially or at once, the approximating (C-DF) transition kernel remains unchanged once surrogate quantities (SCSS) have been calculated over all of the observed data.  Lemma \ref{lem:lin} states that the (C-DF) approximation error increases initially before stabilizing.  If additional data are observed in time (e.g., in the streaming data setting), the C-DF kernel (as defined below) can be shown to generate draws from the exact posterior distribution asymptotically under a few additional assumptions.

\subsection{Convergence for a general approximation class}

For the sake of simplicity, assume that a $p$-dimensional model parameter $\bTheta$ is partitioned into two groups, $\bTheta_1=(\theta_{11},\cdots,\theta_{1p_1})' \in \Re^{p_1}$ and $\bTheta_2 = (\theta_{21},\cdots,\theta_{2p_2})' \in \Re^{p_2}$ where $p = p_1 + p_2$.  As described in Section \ref{sec:cdf_alg}, the C-DF algorithm relies on the existence of surrogate quantities (SCSS) that parameterize the approximating kernels to the full conditional distributions.  Within this framework and without loss of generality, we assume that $\theta_{1i} | \btheta_{1,-i}, \widehat{\bTheta}_2$, $i=1, \dots, p_1$ are conditionally conjugate distributions.

\subsubsection{The C-DF transition kernel}
Assuming two sequences of estimators $\{\widehat{\bTheta}_{1,t}\}_{t\ge1}$, $\{\widehat{\bTheta}_{2,t}\}_{t\ge1}$, the approximating C-DF kernel $T_t: \Re^{p_1}\times\Re^{p_2}\rightarrow\Re^{+}$ at time $t$ may be written in one of two forms:
\begin{enumerate}
\item Approximate (C-DF) conditional distributions $\theta_{2i}\given\btheta_{2,-i},\widehat{\bTheta}_1$, $i=1,\dots,p_2$ are conjugate: parameter updates using approximate transition kernel $T$ proceed in a Gibbs-like fashion with C-DF transition kernel $T_t$ defined as
\begin{align}\label{transker1}
\footnotesize \begin{split}
T_t(\bTheta,\bTheta') = \prod_{i=1}^{p_1} \pi_t\big(\theta_{1i}' | \widehat{\bTheta}_{2,t-1}, \theta_{1l}', l<i, \theta_{1l},l>i\big)
\times \prod_{i=1}^{p_2} \pi_t\big(\theta_{2i}' | \widehat{\bTheta}_{1,t-1}, \theta_{2l}', l<i, \theta_{2l},l>i\big)
\end{split} \end{align}
\item Some (or all) of the approximate (C-DF) conditional distributions $\theta_{2i}\given\btheta_{2,-i},\widehat{\bTheta}_1$, $i=1, \dots, p_2$ are non-conjugate. Then $\bTheta_2$ is updated using a Metropolis-Hastings step with kernel $Q(\bTheta_2, \bTheta_2' \given \widehat{\bTheta}_{1,t-1})$ with C-DF transition kernel $T_t$ defined as
\begin{align}\label{transker2}
T_t(\bTheta,\bTheta') = \prod_{i=1}^{p_1} \pi_t\big(\theta_{1i}'|\widehat{\bTheta}_{2,t-1},\theta_{1l}', l<i, \theta_{1l},l>i\big) \times Q(\bTheta_2, \bTheta_2' \given \widehat{\bTheta}_{1,t-1}).
\end{align}
\end{enumerate}

Lemma \ref{lem:stationary_dist} specifies the unique stationary distribution $f_t: \Re^p\rightarrow\Re^{+}$ of $T_t$.

\begin{lemma} \label{lem:stationary_dist}
C-DF approximate kernel $T_t$ in \eqref{transker1} and \eqref{transker2} have unique stationary distribution
$f_t(\bTheta)=\pi_t(\bTheta_1|\widehat{\bTheta}_{2,t-1})
\pi_t(\bTheta_{2}|\hat{\bTheta}_{1,t-1})$, where $\pi_t(\bTheta_1 | -)$ and $\pi_t(\bTheta_2 | -)$  are the stationary distributions for the respective parameter full conditionals.
\end{lemma}

\begin{remark}
Illustrations in Section  \ref{sec:motivating_examples} fall into the first scenario where all model parameters have conjugate full conditionals that admits surrogate quantities (SCSS).  The dynamic linear model presented in Section \ref{sec:DLM_CDF} is an example where full conditional distributions admit SCSS, although some require sampling via Metropolis-Hastings.  For all examples presented in Section \ref{sec:complex_examples}, the  C-DF algorithm  is modified to accommodate situations in which (i) the parameter space is increasing over time; or (ii) some (or all) of the conditional distributions fail to admit SCSS.  Though asymptotic guarantees on samples drawn are not established by Theorem \ref{theorem1} for such cases, C-DF proves its versatility by producing excellent inferential and predictive performance in all the examples considered (see Sections \ref{sec:motivating_examples} and \ref{sec:complex_examples}).
\end{remark}


\subsubsection{Main convergence result}
Let $\pi_0$ denote the initial distribution from which parameters are drawn.
\begin{theorem}\label{theorem1}
Assume,
(i) $\exists\alpha_t\in(0,1)$ s.t. $\forall$ $t$, $\sup\limits_{\bTheta}~d_{TV}(T_t(\bTheta,\cdot),f_t)\leq 2\alpha_t$,
(ii) $d_{TV}(f_t,f_{t-1})\rightarrow 0$, and
(iii) $d_{TV}(f_t,\pi_t)\rightarrow 0$.  In addition, let $\{n_t\}_{t\geq 1}$ be such that $\alpha_t^{n_t}<\epsilon$ for all large $t$ and for a pre-specified $\epsilon\in (0,1)$.  Then $d_{TV}(T_t^{(n_t)}\cdots T_1^{(n_1)}\pi_0,\pi_t)\rightarrow 0$.
\end{theorem}


\begin{remark}
In essence, Theorem \ref{theorem1} states that running a Markov chain with approximate kernel $T_s$ for $n_s$ iterations at each time point $s = 1, \dots, t$ will asymptotically have draws from the true joint posterior distribution $(t \rightarrow \infty)$.
\end{remark}

\begin{remark}
Condition (i) in Theorem~\ref{theorem1} is referred to as the ``universal ergodicity condition'' \citep{yang2013sequential}, wherein they show that the universal ergodicity condition is weaker than uniform ergodicity condition on the transition kernel $T$.  Condition (ii) ensures that the stationary distribution of the approximating kernel changes slowly as time proceeds.  Lemma 3.7 in \cite{yang2013sequential} shows that condition (ii) is satisfied for any regular parametric model by applying a Bernstein-Von Mises theorem. Finally, condition (iii) requires that the stationary distribution of the approximating kernel becomes `close' to the true posterior distribution at later time. Sufficient conditions under which (iii) holds are outlined in Lemma \ref{lem:TVconv}. Before stating this Lemma, we recall the definition of posterior consistency.
\end{remark}


\begin{definition}
A posterior $\Pi(\cdot|\bD^{(t)})$ is defined to be consistent at $\bTheta^0$ if, for every neighborhood $B$ of $\bTheta^0$, $\Pi(B|\bD^{t})\rightarrow 1$ under the true data generating law at $\bTheta^0$.
\end{definition}

\begin{lemma}\label{lem:TVconv}
Assume that the likelihood function $p_{\bTheta}(\cdot)$ is continuous as a function of $\bTheta$ at $\bTheta^0=(\bTheta_1^0,\bTheta_2^0)$ and $\sqrt{t}p_{\bTheta^0}(\bD^{(t)})$ in limit is bounded away from $0$ and $\infty$. Suppose $\bTheta^0$ is an interior point in the domain and prior distribution $\pi_0(\bTheta_1,\bTheta_2)$ is positive and continuous at $\bTheta^0$. Further, assume $\widehat{\bTheta}_{1,t}\rightarrow\bTheta_1^0$, $\widehat{\bTheta}_{2,t}\rightarrow\bTheta_2^0$ a.s. under the data generating law at $\bTheta_0$, and $f_t$ and $\pi_t$ are both consistent at $\bTheta^0$. Then
\begin{align*}
\int \big|\pi_t(\bTheta)-f_t(\bTheta)\big| \, d\bTheta \rightarrow 0 ~\mbox{as} ~t\rightarrow\infty
\end{align*}
almost surely under the true data generating model at $\bTheta^0$.
\end{lemma}

\begin{remark}
Using the simple fact that
\begin{align*}
d_{TV}(\pi_t,f_t)&=2\sup\limits_{\bA}|\int_{\bA}(\pi_t-f_t)|=2\int\limits_{\pi_t>f_t}(\pi_t-f_t)=\int\limits_{\pi_t>f_t}(\pi_t-f_t)+\int\limits_{f_t>\pi_t}(f_t-\pi_t)\\
&=\int|\pi_t-f_t|,
\end{align*}
it is clear that under the conditions of lemma~\ref{lem:TVconv}, $d_{TV}(\pi_t,f_t)\rightarrow 0\:\:\mbox{as}\:\:t\rightarrow\infty$.
\end{remark}

\begin{remark}
Lemma \ref{lem:TVconv} states that if the likelihood under the data generating model grows at a certain rate (satisfied under standard regularity conditions) and estimates $\widehat{\bTheta}_{1,t},\widehat{\bTheta}_{2,t}$ are consistent estimators of true parameters, the stationary distribution of C-DF approaches the stationary distribution of the Gibbs sampler, thus satisfying condition (iii) of Theorem \ref{theorem1}.  
In the appendix we argue as to how the sequence of estimators obtained by C-DF is consistent for various examples.
\end{remark}

\section{Summary and future work}

The routine collection of large volumes of complex data mandates that model fitting tools evolve quickly to keep pace with the rapidly growing dimension and complexity of data.  To date, there have been a limited number of approaches that scale Bayesian inference to data with a very large number of observations. Popular approaches that often work well in simple models include assumed density filtering (ADF), variational Bayes, expectation propagation (EP), integrated nested Laplace approximation (INLA), along with particle filtering and sequential Monte Carlo (SMC).  However, these methods face substantial difficulty, either computationally or in terms of inferential accuracy, in more complex models (e.g, when the parameter space is large) and come with no accuracy guarantee (with the exception of SMC).  

Conditional density filtering (C-DF) facilitates efficient online Bayesian inference by adapting MCMC to the online setting, with sampling based approximations to conditional posterior distributions obtained by propagating surrogate statistics as new data arrive.  These quantities are computed using sequential point estimates for model parameters along with data shards observed in time.  This eliminates the need to store or process the entire data at once which often results in large computational savings. Approximate samples using C-DF are shown in  Section \ref{sec:theory_cdf} to have the correct stationary distribution as data accrues and C-DFs versatility is demonstrated through illustrative examples in Sections \ref{sec:all_illustrating_examples} and \ref{sec:bcr}. Here, good inferential and predictive performance is accompanied with runtime, memory and sampling efficiency improvements over various state-of-the-art competitors. 

C-DF opens the door to several promising research directions to scale Bayesian inference in more complex hierarchical and nonparametric models that can capture a wide range of naturally occurring predictor-response relations. Here and elsewhere, model specification is routinely expressed in terms of a growing set of observation-specific nuisance parameters, and a more general theoretical analysis is needed to provide a convergence guarantee in such settings, and some of these extensions comprise our current research. 

\clearpage

\renewcommand{\thesubsection}{\Alph{subsection}}
\section*{Appendix}

\subsection{Proofs} \label{sec:proofs}

\noindent\emph{proof of Lemma~\ref{lem:lin}}
\begin{proof}
The proof follows by induction. First we prove an identity that is used in the proof that follows. Letting $\bA\in\mathcal{B}(\mathcal{R}^d)$,
\begin{align}\label{ident:fs}
\begin{aligned}
K^{(r)}(\bTheta,\bA)-T^{(r)}(\bTheta,\bA) &=\int \Big[K^{(r-1)}(\bTheta',\bA)-T^{(r-1)}(\bTheta',\bA)\Big]T(\bTheta,d\bTheta')\\
&~~~+\int \Big[K(\bTheta,d\bTheta')-T(\bTheta,d\bTheta')\Big]K^{(r-1)}(\bTheta',\bA).
\end{aligned}\end{align}
(\ref{ident:fs}) relates the differences between the kernels at $r$-th iteration of the Markov chain. Using $||\nu_1-\nu_2||_{TV}=\sup_{g:\mathcal{R}^d\rightarrow [0,1]}\left|\int g d\nu_1-\int g d\nu_2\right|$ and the fact the fact that r.h.s of \eqref{ident:fs} is free of $\bA$ yields
\begin{align}\label{propag}
\begin{aligned}
||K^{(r)}(\bTheta,\cdot)-T^{(r)}(\bTheta,\cdot)||_{TV} 
&\leq \int ||K^{(r-1)}(\bTheta',\cdot)-T^{(r-1)}(\bTheta',\cdot)||_{TV}T(\bTheta,d\bTheta')\\
&~~~+||K(\bTheta,\cdot)-T(\bTheta,\cdot)||_{TV}.
\end{aligned}\end{align}
Suppose (\ref{eq:linprop}) holds for $(r-1)$. Using (\ref{propag}) we find
\begin{align}\label{eq:finb}
\begin{aligned}
&\sup_{\bTheta}||K^{(r)}(\bTheta,\cdot)-T^{(r)}(\bTheta,\cdot)||_{TV} \\
&\leq \sup_{\bTheta}||K(\bTheta,\cdot)-T(\bTheta,\cdot)||_{TV} + \sup_{\bTheta}\int ||K^{(r-1)}(\bTheta',\cdot)-T^{(r-1)}(\bTheta',\cdot)||_{TV}T(\bTheta,d\bTheta')\\
&\leq \sup_{\bTheta}||K(\bTheta,\cdot)-T(\bTheta,\cdot)||_{TV}+(r-1)\rho<\rho+(r-1)\rho=\rho r.
\end{aligned}\end{align}
Also note that there exists $r_0$ s.t. for all $r>r_0$, we have $||T^{(r)}-\mu_1||_{TV}<\frac{1}{2}||\mu_1-\mu_2||_{TV}$ and $||K^{(r)}-\mu_2||_{TV}<\frac{1}{2}||\mu_1-\mu_2||_{TV}$. By the triangle inequality, for all $r>r_0$
\begin{align}\label{eq:tr}
||K^{(r)}-T^{(r)}||_{TV}\leq ||K^{(r)}-\mu_2||_{TV}+||\mu_1-\mu_2||_{TV}+||T^{(r)}-\mu_1||_{TV}<2||\mu_1-\mu_2||_{TV}.
\end{align}
Comparing (\ref{eq:finb}) and (\ref{eq:tr}) the result follows.
\end{proof}

\noindent\emph{proof of Lemma~\ref{lem:stationary_dist}}
\begin{proof}
We show that $\int T_t(\bTheta,\bTheta')f_t(\bTheta)\,d\bTheta=f_t(\bTheta')$ for case (1) where closed-form sampling from the two sets of full conditionals is assumed (i.e., when the C-DF kernel has the form of a Gibbs kernel, using appropriate approximating substitutions for the full conditionals in terms of SCSS). The proof for case (2) follows in an identical manner taking into account that the MH kernel $Q(\bTheta_2',\bTheta_2|\widehat{\bTheta}_{1,t-1})$ has $\pi_t(\bTheta_{2}|\widehat{\bTheta}_{1,t-1})$ as its stationary distribution.
\begin{align*}
&\int T_t(\bTheta,\bTheta')f_t(\bTheta)d\bTheta\\
&=\int \Big[\prod_{i=1}^{p_1}\pi_t(\Theta_{1i}'|\widehat{\bTheta}_{2,t-1},\Theta_{1l}',l<i,\Theta_{1l},l>i)\Big]
\Big[\prod_{i=1}^{p_2}\pi_t(\Theta_{2i}'|\widehat{\bTheta}_{1,t-1},\Theta_{2l}',l<i,\Theta_{2l},l>i)\Big]\\
&\qquad \pi_t(\bTheta_{1}|\widehat{\bTheta}_{2,t-1})\pi_t(\bTheta_{2}|\widehat{\bTheta}_{1,t-1}) \, d\bTheta_1 \, d\bTheta_2\\
&= \int \Big[\prod_{i=1}^{p_1}\pi_t(\Theta_{1i}'|\widehat{\bTheta}_{2,t-1},\Theta_{1l}',l<i,\Theta_{1l},l>i)\Big]\pi(\bTheta_{1}|\widehat{\bTheta}_{1,t-1}) \,d\bTheta_1\\ 
&~~~\times\int\Big[\prod_{i=1}^{p_2}\pi_t(\Theta_{2i}'|\widehat{\bTheta}_{1,t-1},\Theta_{2l}',l<i,\Theta_{2l},l>i)\Big]\pi_t(\bTheta_{2}|\widehat{\btheta}_{1,t-1}) \,d\bTheta_2\\
&=\pi_t(\bTheta_{1}'|\widehat{\bTheta}_{2,t-1})\pi(\bTheta_{2}'|\widehat{\bTheta}_{1,t-1}).
\end{align*}
The last step follows from the fact that 
\[
\Big[\prod_{i=1}^{p_1}\pi_t(\Theta_{1i}'|\widehat{\bTheta}_{1,t-1},\Theta_{1l}',l<i,\Theta_{1l},l>i)\Big] ~\mbox{and}~ \Big[\prod_{i=1}^{p_2}\pi_t(\Theta_{2i}'|\widehat{\bTheta}_{1,t-1},\Theta_{2l}',l<i,\Theta_{2l},l>i)\Big] 
\]
are the Gibbs kernels with the stationary distribution $\pi_t(\bTheta_{2}|\widehat{\bTheta}_{1,t-1})$ and
$\pi_t(\bTheta_{1}|\widehat{\bTheta}_{2,t-1})$,  respectively. 
\end{proof}

\noindent\emph{proof of Lemma~\ref{theorem1}}\\
The following proof builds on Theorem 3.6 from \cite{yang2013sequential}. Although most of the proof coincides with their result, we present the entire proof for completeness.  
\begin{proof}
Note that, for a fixed $\epsilon\in(0,1)$ $n_t$, $t\geq 1$ are s.t. $\alpha_t^{n_t}<\epsilon$. Using the fact that universal ergodicity condition implies uniform ergodicity, one obtains
\begin{align*}
d_{TV}(T_t^{n_t},\pi_t)\leq \alpha_t^{n_t}<\epsilon.
\end{align*}
Let $h=T_{t-1}^{n_{t-1}}\cdots T_1^{n_1}\pi_0$, then
\begin{align}\label{eq:ineq}
\begin{aligned}
d_{TV}(T_t^{n_t}\cdots T_1^{n_1}\pi_0,f_t)&=d_{TV}(T_t^{n_t}h,f_t)\leq d_{TV}(T_t^{n_t},f_t)\,d_{TV}(h,f_t)\\
&\leq \alpha_t^{n_t}d_{TV}(h,f_t)\leq \epsilon\left(d_{TV}(h,f_{t-1})+d_{TV}(f_t,f_{t-1})\right).
\end{aligned}
\end{align}
using the result repeatedly, one obtains
\begin{align*}
d_{TV}(T_t^{n_t}\cdots T_1^{n_1}\pi_0,f_t)\leq \sum_{l=1}^{t}\epsilon^{t+1-l}d_{TV}(f_l,f_{l-1}).
\end{align*}
R.H.S clearly converges to 0 applying condition (ii).  The proof is completed by using condition (iii) and the fact that
\begin{align*}
d_{TV}(T_t^{n_t}\cdots T_1^{n_1}\pi_0,\pi_t)\leq
d_{TV}(T_t^{n_t}\cdots T_1^{n_1}\pi_0,f_t)+d_{TV}(f_t,\pi_t).
\end{align*}
\end{proof}

\noindent\emph{proof of lemma~\ref{lem:TVconv}}
\begin{proof}
Stationary distribution $f_t$ of the C-DF transition kernel $T_t$ is the approximate posterior distribution to $\pi_t$ obtained at time $t$, and by Lemma \ref{lem:stationary_dist} is given by
\begin{align*}
f_t(\bTheta_1,\bTheta_2)&=\pi_t(\bTheta_1|\widehat{\bTheta}_{2,t})\,\pi_t(\bTheta_2|\widehat{\bTheta}_{1,t})\\
&=\frac{\prod_{l=1}^{t}p_{\bTheta_1,\widehat{\bTheta}_{2,t}}(\bD_l)\prod_{l=1}^{t}p_{\bTheta_1,\widehat{\bTheta}_{2,t}}(\bD_l)\pi_0(\widehat{\bTheta}_{1,t},\bTheta_2) \pi_0(\bTheta_1,\widehat{\bTheta}_{2,t})}{\int \prod_{l=1}^{t}p_{\bTheta_1,\widehat{\bTheta}_{2,t}}(\bD_l)\prod_{l=1}^{t}\,p_{\bTheta_1,\widehat{\bTheta}_{2,t}}(\bD_l)\pi_0(\hat{\bTheta}_{1,t},\bTheta_2)
\pi_0(\bTheta_1,\widehat{\bTheta}_{2,t})}.
\end{align*}
By assumption, $\widehat{\bTheta}_{1,t}\rightarrow\bTheta_1^0$, $\widehat{\bTheta}_{2,t}\rightarrow\bTheta_2^0$ a.s. under $\bTheta^0$, there exists $\Omega_0$ which has probability 1 under the data generating law s.t. for all $\omega\in\Omega_0$, $\widehat{\bTheta}_{1,t}(\omega)$ and $\widehat{\bTheta}_{2,t}(\omega)$ are in an  arbitrarily small neighborhood of $\bTheta_1^0$ and $\bTheta_2^0$, respectively.
Also by assumption, prior $\pi_0$ is continuous  at $\bTheta^0$. That is, given $\epsilon>0$ and $\eta>0$, there exists a neighborhood $N_{\epsilon,\eta}$ s.t. for all $\bTheta\in N_{\epsilon,\eta}$ one has
\begin{align}\label{eq:cont.}
|\pi_0(\bTheta_1,\bTheta_2)-\pi_0(\bTheta_1^0,\bTheta_2^0)|<\epsilon.
\end{align}

Using (\ref{eq:cont.}) and the consistency of $\widehat{\bTheta}_{1,t}$ and $\widehat{\bTheta}_{2,t}$ as above, one obtains for all $t>t_0$ and $\omega\in\Omega_0$
\begin{align}\label{eq:contpi}
|\pi_0(\bTheta_1,\widehat{\bTheta}_{2,t})-\pi_0(\bTheta^0)|<\epsilon, \qquad |\pi_0(\widehat{\bTheta}_{1,t},\bTheta_2)-\pi_0(\bTheta^0)|<\epsilon.
\end{align}
Similarly, continuity of $p_{\bTheta}(\cdot)$ at $\bTheta^0$ leads to the condition that for all $t>t_0$,
\begin{align}\label{eq:pcont}
|p_{\bTheta_1,\bTheta_2}(\bD_l)-p_{\bTheta_1^0,\bTheta_2^0}(\bD_l)|<\epsilon.
\end{align}
Further, consistency assumptions on $f_t$ and $\pi_t$ yield that for all $t>t_1$ and $\omega\in\Omega_1$
\begin{align*}
f_t(N_{\epsilon,\eta}|\bD^{(t)}(\omega))>1-\eta, \qquad \pi_t(N_{\epsilon,\eta}|\bD^{(t)}(\omega))>1-\eta,
\end{align*}
where $\Omega_1$ has probability 1 under the data generating law. Considering $\Omega=\Omega_0\cap\Omega_1$ and
$t_2=\max\{t_1,t_0\}$  it is evident that $\Omega$ has also probability 1 under the true data generating law and all of the above conditions hold for $t>t_2$ and $\omega\in\Omega$.
Simple algebraic manipulations yield
\begin{align}\label{eq:post}
\begin{aligned}
&\frac{f_t(\bTheta|\bD^{(t)}(\omega))}{\pi_t(\bTheta|\bD^{(t)}(\omega))} \\
&=\frac{f_t(N_{\epsilon,\eta}|\bD^{(t)}(\omega))}{\pi_t(N_{\epsilon,\eta}|\bD^{(t)}(\omega))}
\frac{\int_{N_{\epsilon,\eta}}\prod_{l=1}^{t}p_{\bTheta}(\bD_l)\pi_0(\bTheta)}{\prod_{l=1}^{t}p_{\bTheta}(\bD_l)\pi_0(\bTheta)} \\
&\times
\frac{\Big[\prod_{l=1}^{t}p_{\bTheta_1,\widehat{\bTheta}_{2,t}}(\bD_l)\prod_{l=1}^{t}p_{\bTheta_1,\widehat{\bTheta}_{2,t}}(\bD_l)\Big]
\pi_0(\widehat{\bTheta}_{1,t},\bTheta_2) \pi_0(\bTheta_1,\widehat{\bTheta}_{2,t})}{\int_{N_{\epsilon, \eta}} \Big[\prod_{l=1}^{t}p_{\bTheta_1,\widehat{\bTheta}_{2,t}}(\bD_l)\prod_{l=1}^{t}\,p_{\bTheta_1,\widehat{\bTheta}_{2,t}}(\bD_l)\Big]
\pi_0(\hat{\bTheta}_{1,t},\bTheta_2)
\pi_0(\bTheta_1,\widehat{\bTheta}_{2,t})}
\end{aligned}\end{align}
Using (\ref{eq:contpi}) we have
\begin{align*}
&(\pi_0(\bTheta^0)-\epsilon)^2\int_{N_{\epsilon,\eta}} \prod_{l=1}^{t}p_{\bTheta_1,\widehat{\bTheta}_{2,t}}(\bD_l)\,p_{\widehat{\bTheta}_{1,t},\bTheta_2}(\bD_l)\\
&\leq \int_{N_{\epsilon,\eta}} \prod_{l=1}^{t}p_{\bTheta_1,\widehat{\bTheta}_{2,t}}(\bD_l)p_{\widehat{\bTheta}_{1,t},\bTheta_2}(\bD_l) \pi_0(\widehat{\bTheta}_{1,t},\bTheta_2)
\pi_0(\bTheta_1,\widehat{\bTheta}_{2,t})\\
&\leq \big(\pi_0(\bTheta^0)+\epsilon\big)^2 \int_{N_{\epsilon,\eta}} \prod_{l=1}^{t}p_{\bTheta_1,\widehat{\bTheta}_{2,t}}(\bD_l)p_{\widehat{\bTheta}_{1,t},\bTheta_2}(\bD_l).
\end{align*}
Similarly,
\begin{align*}
(\pi_0(\bTheta^0)-\epsilon)\int_{N_{\epsilon,\eta}}\prod_{l=1}^{t}p_{\bTheta}(\bD_l)\leq
\int_{N_{\epsilon,\eta}}\Big[\prod_{l=1}^{t}p_{\bTheta}(\bD_l)\Big]\pi_0(\bTheta)\leq
(\pi_0(\bTheta^0)+\epsilon)\int_{N_{\epsilon,\eta}}\prod_{l=1}^{t}p_{\bTheta}(\bD_l).
\end{align*}
Therefore,
\begin{align*}
\begin{aligned}
&\frac{f_t(\bTheta|\bD^{(t)}(\omega))}{\pi_t(\bTheta|\bD^{(t)}(\omega))} \\
&\quad \leq
(1-\eta)^{-1} \frac{\prod_{l=1}^{t}p_{\bTheta_1,\widehat{\bTheta}_{2,t}}(\bD_l)p_{\widehat{\bTheta}_{1,t},\bTheta_2}(\bD_l)}{\int_{N_{\epsilon,\eta}} \prod_{l=1}^{t}p_{\bTheta_1,\widehat{\bTheta}_{2,t}}(\bD_l)p_{\widehat{\bTheta}_{1,t},\bTheta_2}(\bD_l)}
\frac{\int_{N_{\epsilon,\eta}}\prod_{l=1}^{t}p_{\bTheta}(\bD_l)}{\prod_{l=1}^{t}p_{\bTheta}(\bD_l)}
\frac{(\pi_0(\bTheta^0)+\epsilon)^3}{(\pi_0(\bTheta^0)-\epsilon)^3}.
\end{aligned}
\end{align*}
Using similar calculations we have
\begin{align*}
\begin{aligned}
&\frac{f_t(\bTheta|\bD^{(t)}(\omega))}{\pi_t(\bTheta|\bD^{(t)}(\omega))} \\
&\quad\geq (1-\eta)
\frac{\prod_{l=1}^{t}p_{\bTheta_1,\widehat{\bTheta}_{2,t}}(\bD_l)p_{\widehat{\bTheta}_{1,t},\bTheta_2}(\bD_l)}{\int_{N_{\epsilon,\eta}} \prod_{l=1}^{t}p_{\bTheta_1,\hat{\bTheta}_{2,t}}(\bD_l)p_{\hat{\bTheta}_{1,t},\bTheta_2}(\bD_l)}
\frac{\int_{N_{\epsilon,\eta}}\prod_{l=1}^{t}p_{\bTheta}(\bD_l)}{\prod_{l=1}^{t}p_{\bTheta}(\bD_l)}
\frac{(\pi_0(\bTheta^0)-\epsilon)^3}{(\pi_0(\bTheta^0)+\epsilon)^3}.
\end{aligned}
\end{align*}
Condition \eqref{eq:pcont} now gives us
\begin{align*}
\frac{\prod_{l=1}^{t}(p_{\bTheta^0}(\bD_l)-\epsilon)^3}{\prod_{l=1}^{t}(p_{\bTheta^0}(\bD_l)+\epsilon)^3} 
&\leq 
\frac{\prod_{l=1}^{t}p_{\bTheta_1,\hat{\bTheta}_{2,t}}(\bD_l)p_{\hat{\bTheta}_{1,t},\bTheta_2}(\bD_l)}{\int_{N_{\epsilon,\eta}} \prod_{l=1}^{t}p_{\bTheta_1,\hat{\bTheta}_{2,t}}(\bD_l)p_{\hat{\bTheta}_{1,t},\bTheta_2}(\bD_l)}
\frac{\int_{N_{\epsilon,\eta}}\prod_{l=1}^{t}p_{\bTheta}(\bD_l)}{\prod_{l=1}^{t}p_{\bTheta}(\bD_l)} \\
&\leq \frac{\prod_{l=1}^{t}(p_{\bTheta^0}(\bD_l)+\epsilon)^3}{\prod_{l=1}^{t}(p_{\bTheta^0}(\bD_l)-\epsilon)^3}.
\end{align*}
Using the condition that $\lim_{t\rightarrow \infty} \sqrt{t}p_{\bTheta^0}(\bD^{(t)})$  is bounded away from 0 and $\infty$ and choosing $\epsilon,\eta$ sufficiently small, we have
\begin{align*}
\left|\frac{f_t(\bTheta|\bD^{(t)}(\omega))}{\pi_t(\bTheta|\bD^{(t)}(\omega))}-1\right|<\kappa \quad\mbox{for all}~t>t_2~\mbox{and}~\omega\in\Omega.
\end{align*}
Finally,
\begin{align*}
\int |\pi_t(\bTheta)-f_t(\bTheta)|&\leq\int_{N_{\epsilon,\eta}} |\pi_t(\bTheta)-f_t(\bTheta)|+\int_{N_{\epsilon,\eta}^c} |\pi_t(\bTheta)-f_t(\bTheta)|\\
&\leq\int_{N_{\epsilon,\eta}} |\pi_t(\bTheta)-f_t(\bTheta)|+2\eta\\
&\leq \pi_t(N_{\epsilon,\eta})\kappa+2\eta<\kappa+2\eta.
\end{align*}
\end{proof}

\subsection{Consistent sequence of estimators}
We verify the consistency for C-DF estimators $\widehat{\bTheta}_1$ and $\widehat{\bTheta}_2$ for the examples in Section \ref{sec:motivating_examples}. Details are presented for the linear regression case, with similar steps following for the Anova and compressed regression examples. Step 3 in Algorithm \ref{alg1} proposes an estimator $\widehat{\btheta}_j$ which estimates the mean of the corresponding approximate conditional posterior $\tilde{\pi}_j(\cdot|\bTheta_{\mathcal{G}_l}^{(j)},\bC_j^{(t)})$. When the mean of this distribution is available in a closed form, one might use it to create sequence of estimators. Such mean functions are readily available for the examples in Section \ref{sec:motivating_examples}, and for ease of exposition, we focus on showing consistency for the sequence of estimators in such cases.

For linear regression example in Section \ref{sec:lm}, the sequence of C-DF estimators are given by
\begin{align*}
\widehat{\bbeta}_t &= \bigg(\sum_{l=1}^{t}\frac{\bX_l'\bX_l}{\hat{\sigma}_l^2}+\bI\bigg)^{-1}\sum_{l=1}^{t}\frac{\bX_l'\by_l}{\hat{\sigma}_l^2}\\
\hat{\sigma}_t^2 &= \frac{2b+\sum_{l=1}^{t}\by_l'\by_l-2\sum_{l=1}^{t}\widehat{\bbeta}_l'\bX_l'\by_l+\sum_{l=1}^{t}\widehat{\bbeta}_l'\bX_l'\bX_l\widehat{\bbeta}_l}
                    {2a+nt-2}.
\end{align*}
Let the true regression model be $\by_t=\bX_t\bbeta_0+\bepsilon_t$, $\epsilon_t\sim N(\bzero,\sigma^2_0\bI)$. Then, the maximum a-posteriori estimator for the vector of regression coefficients is given in closed form as 
\begin{align*}
\hat{\bbeta}_t &= \bigg(\sum_{l=1}^{t}\frac{\bX_l'\bX_l}{\hat{\sigma}_l^2}+\bI\bigg)^{-1}\sum_{l=1}^{t}\frac{\bX_l'(\bX_l\bbeta_0+\bepsilon_l)}{\hat{\sigma}_l^2}\\
&=\bbeta_0-\bigg(\sum_{l=1}^{t}\frac{\bX_l'\bX_l}{\hat{\sigma}_l^2}+\bI\bigg)^{-1}\bbeta_0 + \bigg(\sum_{l=1}^{t}\frac{\bX_l'\bX_l}{\hat{\sigma}_l^2}+\bI\bigg)^{-1}\sum_{l=1}^{t}\frac{\bX_l'\bepsilon_l}{\hat{\sigma}_l^2}.
\end{align*}

Assume (i) $\hat{\sigma}_t^2\rightarrow\sigma_0^2$ a.s. under the data generating law and (ii) that $\tfrac{1}{nt}\sum_{l=1}^{t}\bX_l'\bX_l$ has bounded eigenvalues for every $t$.  Under the above assumptions, and using the weighted SLLN (Adler \& Rosalsky, 1991), we obtain $\big(\sum_{l=1}^{t}\frac{\bX_l'\bX_l}{\hat{\sigma}_l^2}+\bI\big)^{-1}\sum_{l=1}^{t}\frac{\bX_l'\bepsilon_l}{\hat{\sigma}_l^2}\rightarrow \bzero$ a.s. and $\big(\sum_{l=1}^{t}\frac{\bX_l'\bX_l}{\hat{\sigma}_l^2}+\bI\big)^{-1}\bbeta_0\rightarrow \bzero$ a.s.  Then, $\widehat{\bbeta}_t\rightarrow\bbeta_0$ a.s. under the data generating law.  Similarly, assuming $\widehat{\bbeta}_t\rightarrow\bbeta_0$ a.s., one can show $\hat{\sigma}_t^2 \rightarrow \sigma_0^2$ a.s. Simultaneous convergence of the model parameters is argued on the basis of established results on alternate optimization \citep{byrne2013alternating}.

\clearpage

\subsection{Accuracy Plots} \label{sec:additional_figs}

\subsection{Poisson mixed effect model}\label{sec:poi_mix}

Consider the Poisson additive mixed effects model where count data $\by^t$ with predictors $\bX^t$ are modeled as
\begin{align*}
\by^t &\sim \mathrm{Poisson}\big(\exp(\bX^t \bbeta+ \bZ\bu)\big),
~\bbeta\sim N(\bzero,\sigma_{\beta}^2\bI_k) \\
\bu &\sim N\big(\bzero, \mathrm{diag}(\sigma_1^2\bI_{k_1}, \dots,
\sigma_r^2\bI_{k_r})\big), ~ \sigma_s^2\sim
IG(0.5,1/b_s), ~b_s \sim IG(1/2,1/a_s^2),
\end{align*}
with $\bI_{k_1}, \dots, \bI_{k_r}$ respectively denoting identity
matrices of order $k_1, \dots ,k_r$ and $s = 1, 2, \dots, r$.
Here, obstacles to implementing the C-DF algorithm immediately arise
as the full conditional distributions have no closed-form and do not
admit surrogate quantities to propagate.
The global approximation made by the ADF approximation becomes
increasingly unreliable and overly restrictive in higher
dimensions.  A possibly less restrictive approximation may seek to
obtain an ``optimal'' approximation to the posterior subject to ignoring
posterior dependence among different parameters.  This is given by a
variational approximation to the joint posterior over model parameters
$\bbeta, \bu, \sigma_1^2, \dots,\sigma_r^2$, namely
\begin{align*}
\pi(\bbeta, \bu, \sigma_1^2, \dots, \sigma_r^2 | \bD^t) \approx
q_1(\bbeta, \bu)
\: q_2(\sigma_1^2, \dots, \sigma_r^2)
\: q_3(b_1, \dots, b_r).
\end{align*}
Closed-form expressions for $q_1, q_2, q_3$ are derived and given by
\begin{align*}
q_1 = N(\bmu_{\bbeta,\bu},\bSigma_{\bbeta,\bu}),
\: q_2 = \prod\limits_{s=1}^{r}IG\left(\tfrac{k_s+1}{2}, \mu_{1/b_s}+
\tfrac{||\bmu_{\bu_s}||^2 + \mathrm{tr}(\bSigma_{\bu_s})}{2}\right),
\: q_3 = \prod\limits_{s=1}^{r}IG(1,\mu_{1/\sigma_s^2}+a_s^{-2}).
\end{align*}

Above, $\mu_{1/\sigma_s^2} = \int (1/\sigma_s^2)q(\sigma_s^2)$, with
$\mu_{1/b_s}$ defined similarly, and $\bmu_{\bu},\bSigma_{\bu}$
represent the mean and covariance specific to $\bu$ under
approximating density $q_1$. The approximate posterior is completely specified in terms of
$\bmu_{\bbeta,\bu}, \: \bSigma_{\bbeta,\bu}, \text{ and } \mu_{1/b_s}, \:
\mu_{1/\sigma_s^2}$ for $s = 1, \dots, r$, hence the C-DF algorithm in
this setup is applied directly on these parameters.  In particular,
one only needs point estimates for these parameters to fully specify
the approximate posterior, hence sampling steps (see Algorithm
\ref{alg1}) are replaced by fixed-point iteration.  With a partition
$\bTheta_{\mathcal{G}_1} = \{\bmu_{\bbeta,\bu},
\bSigma_{\bbeta,\bu}\}$, $\bTheta_{\mathcal{G}_2} = \{ \mu_{1/b_s}, \:
\mu_{1/\sigma_s^2}, \:s = 1, \dots, r\}$ for the
parameters of the variational distribution, the C-DF algorithm
proceeds as follows:
\begin{enumerate}[(1)]
\item Observe data shard $(\by^t,\bX^t)$ at time $t$.  If $t=1$, initialize all the parameters using draws from respective priors.  Otherwise set $\bmu_{\bbeta,\bu}^t \leftarrow\bmu_{\bbeta,\bu}^{(t-1)}, \: \bSigma_{\bbeta,\bu}^t \leftarrow\bSigma_{\bbeta,\bu}^{(t-1)}, \: \mu_{1/\sigma_s^2}^t \leftarrow\mu_{1/\sigma_s^2}^{(t-1)}, \: \mu_{1/b_s}^t \leftarrow \mu_{1/b_s}^{(t-1)}$;
\item The following steps are repeated a maximum of $N_\mathrm{fx}$ iterations (or until convergence):
\begin{enumerate}[(a)]
\item Update $\bD \leftarrow \mathrm{blockdiag}(\sigma_{\beta}^{-2}\bI_k, \hat{\mu}_{1/\sigma_1^2}\bI_{k_1}^t, \dots, \hat{\mu}_{1/\sigma_r^2}^t\bI_{k_r})$;
\item Update $w_{\bbeta,\bu} \leftarrow \exp\big(\bc^{t'}\bmu_{\bbeta,\bu} + \frac{1}{2}\bc^{t'}\bSigma_{\bbeta,\bu}\bc^t\big),$ with $\bc^t=[\bX^t,\bZ]$;
\item Set $H_1=\bC_{1,1}^{(t-1)} + \bc^t\by^t$, $H_2 = \bC_{1,2}^{(t-1)} + \bc^t w_{\bbeta,\bu}$, and $H_3 = \bC_{1,3}^{(t-1)} + w_{\bbeta,\bu}\bc^t\bc^{t'}$;
\item Update $\bmu_{\bbeta,\bu} \leftarrow \bmu_{\bbeta,\bu}^{(t-1)} + \bSigma_{\bbeta,\bu} \big(H_1 - H_2 - \bD\bmu_{\bbeta,\bu}\big)$, and
 $\bSigma_{\bbeta,\bu} \leftarrow \big(H_3 + \bD\big)^{-1}$.
 \end{enumerate}
 \item Set $\bmu_{\bbeta,\bu}^t, \: \bSigma_{\bbeta,\bu}^t$ at the final values after $N_\mathrm{fx}$ iterations;
 \item Also repeat the following step a maximum of $N_\mathrm{fx}$ iterations (or until convergence):
 \begin{enumerate}[(a)]
\item Update $\mu_{1/b_s} \leftarrow (\mu_{1/\sigma_s^2}+a_s^{-2})^{-1}$, $\mu_{1/\sigma_s^2} \leftarrow (k_s+1) / \big(2\mu_{1/b_s} + ||\bmu_{\bu}^t||^2 + \mathrm{tr}(\bSigma_{\bu}^t)\big), ~ s = 1, \dots, r$.
\end{enumerate}
\item Set $\mu_{1/\sigma_s^2}^t, \: \mu_{1/b_s}^t$ at the final values
  after $N_\mathrm{fx}$ iterations;
\item Update surrogate quantities $\bC_{1,1}^t \leftarrow
  \bC_{1,1}^{(t-1)} + \bc^t \by^t$; $\bC_{1,2}^t \leftarrow
  \bC_{1,2}^{(t-1)} + \bc^t w_{\bbeta,\bu}^t$; $\bC_{1,3}^t \leftarrow
  \bC_{1,3}^{(t-1)} + w_{\bbeta,\bu}^t\bc^t\bc^{t'}$.
\end{enumerate}

With the arrival of new data, $N_\mathrm{fx}$ fixed-point iterations update parameters estimates for the approximating distributions.  Consistent with the definition of SCSS, an update to the parameters for $\bTheta_{\mathcal{G}_1}$ under $q_1$ may involve estimates from the previous time-point, in addition to estimates for parameters $\bTheta_{\mathcal{G}_2}$ and vice versa.  This  online sampling scheme has excellent empirical performance, both in terms of parametric inference and prediction (see \cite{luts2013variational}).  This establishes the broad applicability of propagating surrogate quantities (SCSS) using the C-DF algorithm in a non-conjugate setting using variational approximation methods. 

\bibliographystyle{chicago}
\bibliography{bcdf_jmlr_arxiv.bbl}

\end{document}

%% file: defs.tex
\newtheorem{theorem}{Theorem}[section]
\newtheorem{lemma}[theorem]{Lemma}

\newenvironment{proof}[1][Proof]{\begin{trivlist}
\item[\hskip \labelsep {\bfseries #1}]}{\end{trivlist}}
\newenvironment{definition}[1][Definition:]{\begin{trivlist}
\item[\hskip \labelsep {\bfseries #1}]}{\end{trivlist}}

\newenvironment{remark}[1][Remark]{\begin{trivlist}
\item[\hskip \labelsep {\bfseries #1}]}{\end{trivlist}}

\newcommand{\qed}{\nobreak \ifvmode \relax \else
      \ifdim\lastskip<1.5em \hskip-\lastskip
      \hskip1.5em plus0em minus0.5em \fi \nobreak
      \vrule height0.75em width0.5em depth0.25em\fi}

\newcommand{\bbeta}{ {\boldsymbol \beta} }

\newcommand{\bgamma}{ {\boldsymbol \gamma} }

\newcommand{\bepsilon}{ {\boldsymbol \epsilon} }

\newcommand{\bPhi}{ {\boldsymbol \Phi} }

\newcommand{\btau}{ {\boldsymbol \tau}}

\newcommand{\blambda}{ {\boldsymbol \lambda} }

\newcommand{\bmu}{ {\boldsymbol \mu} }

\newcommand{\bSigma}{ {\boldsymbol \Sigma} }
\newcommand{\btheta}{ {\boldsymbol \theta} }
\newcommand{\bTheta}{ {\boldsymbol \Theta} }
\newcommand{\bzeta}{ {\boldsymbol \zeta} }
\newcommand{\bPsi}{ {\boldsymbol \Psi} }

\DeclareMathOperator{\sgn}{\mathop{\mathrm{sign}}}

\DeclareMathOperator{\diag}{diag}

\newcommand{\bzero}{ {\boldsymbol 0} }
\newcommand{\bones}{ {\boldsymbol 1} }

\newcommand{\bA}{ {\boldsymbol A} }

\newcommand{\bc}{ {\boldsymbol c} }
\newcommand{\bC}{ {\boldsymbol C} }

\newcommand{\bD}{ {\boldsymbol D} }

\newcommand{\bF}{ {\boldsymbol F} }

\newcommand{\bH}{ {\boldsymbol H} }

\newcommand{\bI}{ {\boldsymbol I} }

\newcommand{\bK}{ {\boldsymbol K} }

\newcommand{\bL}{ {\boldsymbol L} }

\newcommand{\bR}{ {\boldsymbol R} }

\newcommand{\bS}{ {\boldsymbol S} }

\newcommand{\bu}{ {\boldsymbol u} }

\newcommand{\bW}{ {\boldsymbol W} }
\newcommand{\bx}{ {\boldsymbol x} }
\newcommand{\bX}{ {\boldsymbol X} }
\newcommand{\by}{ {\boldsymbol y} }

\newcommand{\bz}{ {\boldsymbol z} }
\newcommand{\bZ}{ {\boldsymbol Z} }

\newcommand{\given}{\,|\,}

\newcommand{\mt}[1]{\tilde{#1}}